%% file: icml_submission.tex
\documentclass{article}

\usepackage{microtype}
\usepackage{graphicx}
\usepackage{subfigure}
\usepackage{booktabs} 

\usepackage{hyperref}

\usepackage[accepted]{icml2021}

\usepackage{comment}

\usepackage{smile}
\usepackage{extarrows}
\usepackage[OT1]{fontenc}
\usepackage{algorithm}
\usepackage{algorithmic}
\RequirePackage{times}
\RequirePackage{natbib}
\usepackage[usenames,dvipsnames]{xcolor}
\hypersetup{
  pdftex,
  pdffitwindow=true,
  pdfstartview={FitH},
  pdfnewwindow=true,
  colorlinks,
  linktocpage=true,
  linkcolor=Red,
  urlcolor=Red,
  citecolor=blue
}
\usepackage[capitalize]{cleveref}

\newcommand{\mP}{\mathbb{P}}

\usepackage[textsize=tiny]{todonotes}

\def\red#1{\textcolor{red}{#1}}
\def\vec{\text{vec}}

\icmltitlerunning{Bootstrapping Fitted Q-Evaluation for Off-Policy Inference}

\begin{document}

\twocolumn[
\icmltitle{Bootstrapping Fitted Q-Evaluation for Off-Policy Inference}



\icmlsetsymbol{equal}{*}

\begin{icmlauthorlist}
\icmlauthor{Botao Hao}{to}
\icmlauthor{Xiang Ji}{goo}
\icmlauthor{Yaqi Duan}{goo}
\icmlauthor{Hao Lu}{goo}
\icmlauthor{Csaba Szepesv\'ari}{to,ua}
\icmlauthor{Mengdi Wang}{to,goo}
\end{icmlauthorlist}

\icmlaffiliation{to}{Deepmind}
\icmlaffiliation{goo}{Princeton University}
\icmlaffiliation{ua}{University of Alberta}

\icmlcorrespondingauthor{Botao Hao}{haobotao000@gmail.com}
\icmlcorrespondingauthor{Mengdi Wang}{mengdiw@princeton.edu}

\icmlkeywords{Machine Learning, ICML}

\vskip 0.3in
]



\printAffiliationsAndNotice{}  

\begin{abstract}
Bootstrapping provides a flexible and effective approach for assessing the quality of batch reinforcement learning, yet its theoretical properties are poorly understood. In this paper, we study the use of bootstrapping in off-policy evaluation (OPE), and in particular, we focus on the fitted Q-evaluation (FQE) that is known to be minimax-optimal in the tabular and linear-model cases. We propose a bootstrapping FQE method for inferring the distribution of the policy evaluation error and show that this method is asymptotically efficient and distributionally consistent for off-policy statistical inference. To overcome the computation limit of bootstrapping, we further adapt a subsampling procedure that improves the runtime by an order of magnitude. We numerically evaluate the bootrapping method in classical RL environments for confidence interval estimation, estimating the variance of off-policy evaluator, and estimating the correlation between multiple off-policy evaluators.

\end{abstract}

\section{Introduction}

Off-policy evaluation (OPE) often serves as the starting point of batch reinforcement learning (RL).
The objective of OPE is to estimate the
value of a target policy based on batch episodes of state-transition trajectories that were generated using a different and possibly unknown behavior
policy. In this paper, we investigate statistical inference for OPE. In particular, we analyze the popular fitted Q-evaluation (FQE) method, which is a basic model-free approach that fits unknown value function from data using function approximation and backward dynamic programming \citep{fonteneau2013batch, munos2008finite, le2019batch}. In practice, FQE has demonstrated robust and satisfying performances on many classical RL tasks under different metrics \citep{voloshin2019empirical}. A more recent study by \citet{paine2020hyperparameter} demonstrated surprising scalability and effectiveness of FQE with deep neural nets in a range of complex continuous-state RL tasks. On the theoretical side, FQE was proved to be a minimax-optimal policy evaluator in the tabular and linear-model cases \cite{yin2020asymptotically,duan2020minimax}.

The aforementioned research mostly focuses on point estimation for OPE. In practical batch RL applications, a point estimate is far from enough. Statistical inference for OPE is of great interests. For instance, one often hopes to construct tight confidence interval around policy value, estimate the variance of off-policy evaluator, or evaluate multiple policies using the same data and estimate their correlations. Bootstrapping \citep{efron1982jackknife}, is a conceptually simple and generalizable approach to infer the error distribution based on batch data. Therefore, in this work, we study the use of bootstrapping for \emph{off-policy inference}. We will provide theoretical justifications as well as numerical experiments. 

Our main results are summarized below:
\begin{itemize}
    \item First we analyze the asymptotic  distribution of FQE with linear function approximation and show that the policy evaluation error asymptotically follows a normal distribution (Theorem \ref{thm:asy_normality_OPE}). The asymptotic variance matches the Cramér–Rao lower bound for OPE (Theorem \ref{thm:efficiency}) and implies that this estimator is asymptotically efficient. 
    \item We propose a bootstrapping FQE method for estimating the distribution of off-policy evaluation error. We prove that bootstrapping FQE is asymptotically consistent in estimating the distribution of the original FQE (Theorem \ref{thm:bootstrap_consistency}) and establish the consistency of bootstrap confidence interval as well as bootstrap variance estimation. Further, we propose a subsampled bootstrap procedure to improve the computational efficiency of bootstrapping FQE. 
    \item We highlight the necessity of {\it bootstrapping by episodes}, rather than by individual sample transition as considered in previous works; see \citet{ kostrikov2020statistical}. 
    The reason is that  bootstrapping dependent data in general fails to characterize the right error distribution (Remark 2.1 in \citet{singh1981asymptotic}). We illustrate this  phenomenon via experiments (see Figure \ref{fig:distribution}).
        All our theoretical analysis applies to episodic dependent data, and we do not require the i.i.d. sample transition assumption commonly made in OPE literatures \citep{jiang2020minimax,kostrikov2020statistical, dai2020coindice}. 
   
    \item Finally, we evaluate subsampled bootstrapping FQE in a range of classical RL tasks, including a discrete tabular domain, a continuous control domain and a simulated healthcare example. We test variants of bootstrapping FQE with tabular representation, linear function approximation, and neural networks. We carefully examine the effectiveness and tightness of bootstrap confidence intervals, as well as the accuracy of bootstrapping for estimating the variance and correlation for OPE.
\end{itemize}

\textbf{Related Work.} Point estimation of OPE receives considerable attentions in recent years. Popular approaches include direct methods \citep{lagoudakis2003least, ernst2005tree, munos2008finite, le2019batch}, double-robust / importance sampling  \citep{precup2000eligibility,jiang2016doubly, thomas2016data}, marginalized importance sampling \citep{hallak2017consistent, liu2018breaking, xie2019towards, nachum2019dualdice, uehara2019minimax, zhang2020gendice, zhang2020gradientdice}. On the theoretical side, \citet{uehara2019minimax, yin2020asymptotically} established asymptotic optimality and efficiency for OPE in the tabular setting and \citet{kallus2020double} provided a complete study of semiparametric efficiency in a more general setting.  \citet{duan2020minimax, hao2020sparse} showed that FQE with linear/sparse lienar function approximation is minimax optimal and \citet{wang2020statistical} studied the fundamental hardness of OPE with linear function approximation.
 
Confidence interval estimation of OPE is also important in many high-stake applications. \citet{thomas2015high} proposed a high-confidence OPE based on importance sampling and empirical Bernstein inequality. \citet{kuzborskij2020confident} proposed a tighter confidence interval for contextual bandits based on empirical Efron-Stein inequality. However, importance sampling suffers from the curse of horizon \citep{liu2018breaking} and concentration-based confidence intervals are typically overly-conservative since they only exploit tail information \citep{hao2019bootstrapping}. Another line of recent works formulated the estimation of confidence intervals into an optimization problem \citep{feng2020accountable,feng2021nonasymptotic, dai2020coindice}. 
These works are specific to confidence interval construction for OPE, and they do not provide distributional consistency guarantee. Thus, they don't easily generalize to other statistical inference tasks. 

In statistics community, \citet{liao2019off} studied OPE in an infinite-horizon undiscounted MDP and derived the asymptotic distribution of empirical Bellman residual minimization estimator. Their asymptotic variance had a tabular representation and thus didn't show the effect of function approximation. \citet{shi2020statistical} considered asymptotic confidence interval for policy value but under different model assumption that assumes Q-function is smooth.

Several existing work has investigated the use of bootstrapping in OPE. \citet{thomas2015high, hanna2017bootstrapping} constructed confidence intervals by bootstrapping importance sampling estimator or learned models but didn't come with any consistency guarantee. The most related work is \citet{kostrikov2020statistical} that provided the first asymptotic consistency of bootstrap confidence interval for OPE. Our analysis improves their work in the following aspects. First, we study FQE with linear function approximation while \citet{kostrikov2020statistical} only considered the tabular case. Second, we provide distributional consistency of bootstrapping FQE which is stronger than the consistency of confidence interval in \citet{kostrikov2020statistical}.

\section{Preliminary}

Consider an episodic Markov decision process (MDP) that is defined by a tuple $M = (\cS, \cA, P, r, H)$. Here, $\cS$ is the state space, $\cA$ is the action space, $P(s'|s,a)$ is the probability of reaching state $s'$ when taking action $a$ in state $s$,  $r:\cS\times \cA\to[0, 1]$ is the reward function, and $H$ is the length of horizon. A policy $\pi: \cS \to \cP(\cA)$ maps states to a distribution over actions. The state-action value function (Q-function) is defined as, for $h=1,\ldots,H$, 
\begin{equation*}
\label{Q_def} Q_h^{\pi}(s,a) = \mathbb{E}^{\pi} \Bigg[ \sum_{h'=h}^H \! r(s_{h'}, a_{h'}) \, \Bigg| \, s_h = s, a_h = a \Bigg], \end{equation*}
where $a_{h'} \sim  \pi(\cdot \, | \, s_{h'}), s_{h'+1}  \sim P(\cdot \, | \, s_{h'}, a_{h'})$ and $\mathbb{E}^{\pi} $ denotes expectation over the sample path generated under policy $\pi$. The Q-function satisfies the Bellman equation for policy $\pi$: 
$$
Q_{h-1}^{\pi}(s,a) = r(s,a)+\mathbb E\Big[V_{h}^{\pi}(s')\big|s,a\Big],
$$
where $s'\sim P(\cdot|s,a)$ and $V_{h}^{\pi}:\cS\to\mathbb R$ is the value function defined as $V_{h}^{\pi}(s) = \int_{a}Q_h^{\pi}(s,a)\pi(a|s){\rm d}a.$ 

Let $[n]=\{1,\ldots, n\}$. For a positive semidefinite matrix $X$, we denote $\lambda_{\min}(X)$ as the minimum eigenvalue of $X$. Denote $I_d\in\mathbb R^{d\times d}$ as a diagonal matrix with 1 as all the diagonal entry and 0 anywhere else.

\paragraph{Off-policy evaluation.} Suppose that the batch data $\cD = \{\cD_1,\ldots, \cD_K\}$ consists of $K$ independent episodes collected using an \emph{unknown behavior policy} $\bar{\pi}$. Each episode, denoted as $\cD_k = \{ (s_h^k,a_h^k, r_h^k)\}_{h\in[H]}$, is a trajectory of $H$ state-transition tuples. It is easy to generalize our analysis to multiple unknown behavior policies since our algorithms do not require the knowledge of the behavior policy. Let $N=KH$ be the total number of sample transitions; and we sometimes write $\cD = \{(s_n,a_n,r_n)\}_{n\in[N]}$ for simplicity. The goal of OPE is to estimate the expected cumulative return (i.e., value) of a \emph{target policy} $\pi$ from a a fixed initial distribution $\xi_1$, based on the dataset $\cD$. The value is defined as
\begin{equation*} 
\label{vpi} v_{\pi} = \mathbb{E}^{\pi} \Bigg[ \sum_{h=1}^H r(s_h,a_h) \, \Bigg| \, s_1 \sim \xi_1 \Bigg].
\end{equation*}

\paragraph{Fitted Q-evaluation.} Fitted Q-evaluation (FQE) is an instance of the fitted Q-iteration method, dated back to \citet{fonteneau2013batch, le2019batch}.  Let $\cF$ be a given function class, for examples a linear function class or a neural network class. Set $\hat{Q}_{H+1}^{\pi}=0$. For $h=H,\ldots,1$, we recursively estimate $Q_h^{\pi}$ by regression and function approximation:
\begin{equation*}\label{eqn:supervised}
     \hat{Q}^{\pi}_h =\argmin_{f\in\cF}\Big\{\frac{1}{N}\sum_{n=1}^{N} \Big(f(s_n, a_n)-y_n\Big)^2+ \lambda \rho(f)\Big\},
\end{equation*}
where $y_n = r_n+\int_{a}\hat{Q}^{\pi}_{h+1}(s_{n+1}, a)\pi(a|s_{n+1}){\rm d}a$ and $\rho(f)$ is a proper regularizer. The value estimate is 
\begin{equation}\label{eqn:FQE}
    \hat{v}_{\pi} = \mathbb E_{s\sim\xi_1, a\sim\pi(\cdot|s)}\Big[\hat{Q}^{\pi}_1(s, a)\Big],
\end{equation}
which can be directly computed based on $\hat{Q}^{\pi}_1$.
See the full description of FQE in Appendix \ref{sec:alg_FQE}.

\paragraph{Off-policy inference.} 
Let $\hat{v}_{\pi}$ be an off-policy estimator of the target policy value $v_{\pi}$. In addition to the point estimator, we are primarily interested in the distribution of the off-policy evaluation error $\hat{v}_{\pi}-v_{\pi}$.
We aim to infer the error distribution of $\hat{v}_{\pi}-v_{\pi}$ in order to conduct statistical inference. Suppose $F$ is an estimated distribution of $\hat{v}_{\pi}-v_{\pi}$. Then we can use $F$ for a range of downstream off-policy inference tasks, for examples:
\begin{itemize}
    \item \emph{Moment estimation.} With $F$, we can estimate the $p$-th moment of $\hat{v}_{\pi}-v_{\pi}$ by $\int x^p d F(x)$. Two important examples are bias estimation and variance estimation.
    \item \emph{Confidence interval construction.} Define the quantile function of $F$ as
    $\cG(p) = \inf \{x\in\mathbb R, p\leq F(x)\}.
    $
    Specify a confidence level $0<\delta\leq 1$. With $F$, we can construct the $1-\delta$ confidence interval as $[\hat{v}_{\pi}-\cG(1-\delta/2), \hat{v}_{\pi}-\cG(\delta/2)].$
  If $F$ is close to the true distribution of $\hat{v}_{\pi}-v_{\pi}$, the above one would be the nearly tightest confidence interval for $v_{\pi}$ based on $\hat{v}_{\pi}$. 
    \item \emph{Evaluating multiple policies and estimating their correlation.} Suppose there are two target policies $\pi_1,\pi_2$ to evaluate and the corresponding off-policy estimators are $\hat{v}_{\pi_1}, \hat{v}_{\pi_2}$. Let $F_{12}$ be the estimated joint distribution of $\hat{v}_{\pi_1}-v_{\pi_1}$ and $\hat{v}_{\pi_2}-v_{\pi_2}$. The Pearson correlation coefficient between the two estimators is 
    \begin{equation*}
        \rho(\hat{v}_{\pi_1}, \hat{v}_{\pi_2}) = \frac{\text{Cov}(\hat{v}_{\pi_1}, \hat{v}_{\pi_2})}{\sqrt{\text{Var}(\hat{v}_{\pi_1})\text{Var}(\hat{v}_{\pi_2})}}.
    \end{equation*}
   Both the covariance and variance can be estimated from $F_{12}$, so we can further estimate the correlation between off-policy evaluators.
\end{itemize}

\begin{remark}[Practical scenarios of estimating correlations]
Correlation is a basic statistical metric for comparing two estimators, and we used it as an example to illustrate that bootstrapping can be used for estimating a variety of statistics not limited to confidence intervals. In medical applications, we may have multiple target treatment policies to compare against, where a correlation estimate together with confidence intervals would make physicians better informed to make a fairer comparison. 
\end{remark}

\section{Bootstrapping Fitted Q-Evaluation (FQE)}

As shown in \citet{le2019batch,voloshin2019empirical, duan2020minimax, paine2020hyperparameter}, FQE not only demonstrates strong empirical performances, but also enjoys provably optimal theoretical guarantees. Thus it is natural to conduct bootstrapping on top of FQE for off-policy inference.

Recall the original dataset $\cD$ consists of $K$ episodes. We propose to bootstrap FQE {\it by episodes}: Draw sample episodes $\cD_1^*,\ldots,\cD_K^*$ independently with replacement from $\cD$.
This is the standard Efron's nonparametric bootstrap \citep{efron1982jackknife}. Then we run FQE on the new bootstrapped set $\cD^*=\{\cD_1^*,\ldots, \cD_K^*\}$ as in Eq.~\eqref{eqn:FQE} and let the output $\hat{v}_{\pi}^*$ as the bootstrapping FQE estimator. By repeating the above process, we may obtain multiple samples of  $\hat{v}_{\pi}^*$, and may use these samples to further conduct off-policy inference (see Section \ref{sec:inference} for details).

\subsection{Bootstrap by episodes {\it vs.}   boostrap by sample transitions}

Practitioners may wonder what is the right way to bootstrap a data set. This question is quite well understood in supervised learning when the data points are independent and identically distributed; there the best way to bootstrap is to resample data points directly. However, in episodic RL, although episodes may be generated independently from one another, sample transitions $(s_n, a_n, r_n)$ in the same episode are highly dependent. Therefore, we choose to bootstrap the batch dataset \emph{by episodes}, rather than \emph{by sample transitions} which was commonly done according to previous literatures \citep{ kostrikov2020statistical}. 
 \begin{figure}[t]
 \centering
 \includegraphics[width=0.35\linewidth]{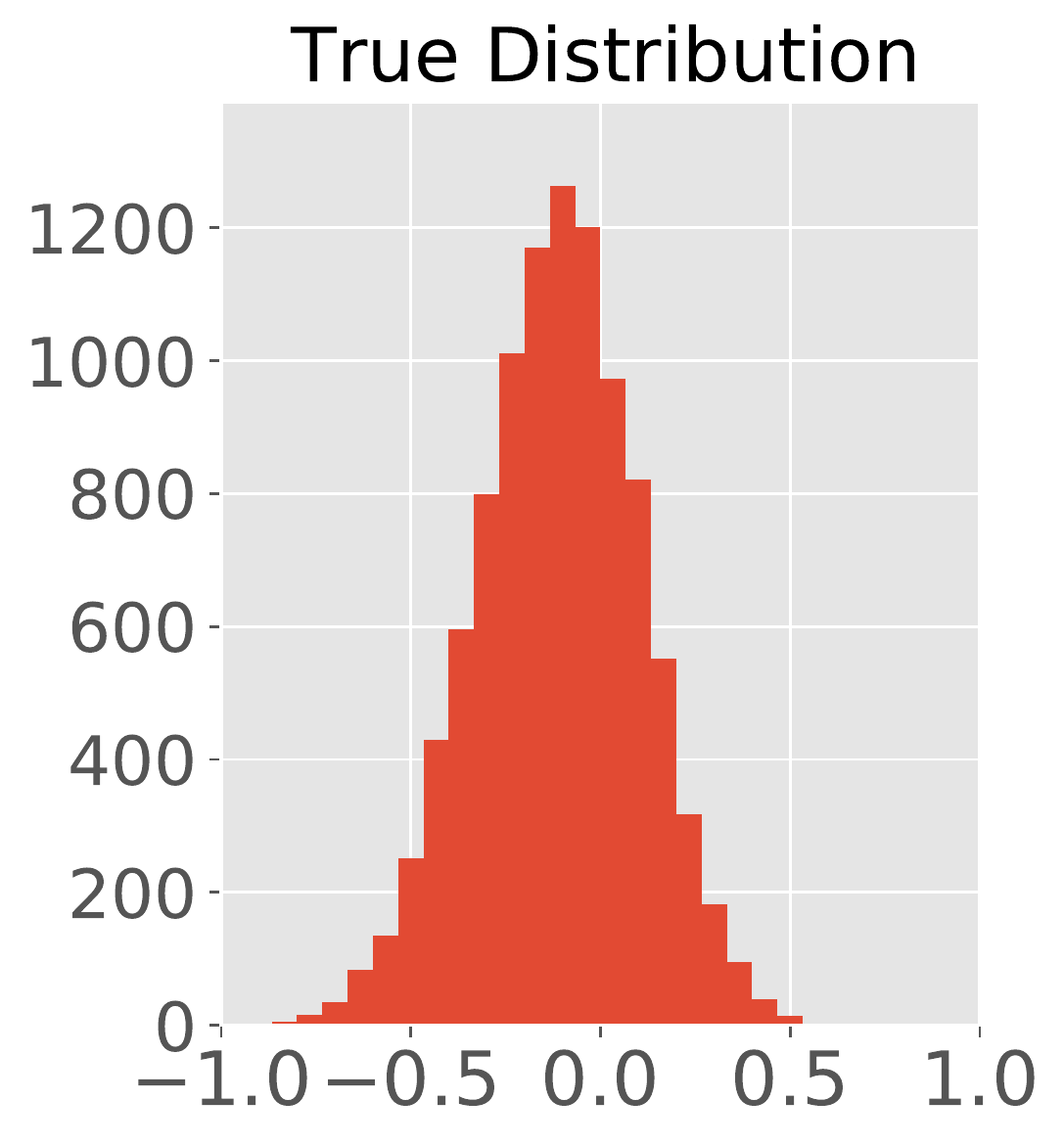}
\includegraphics[width=0.35\linewidth]{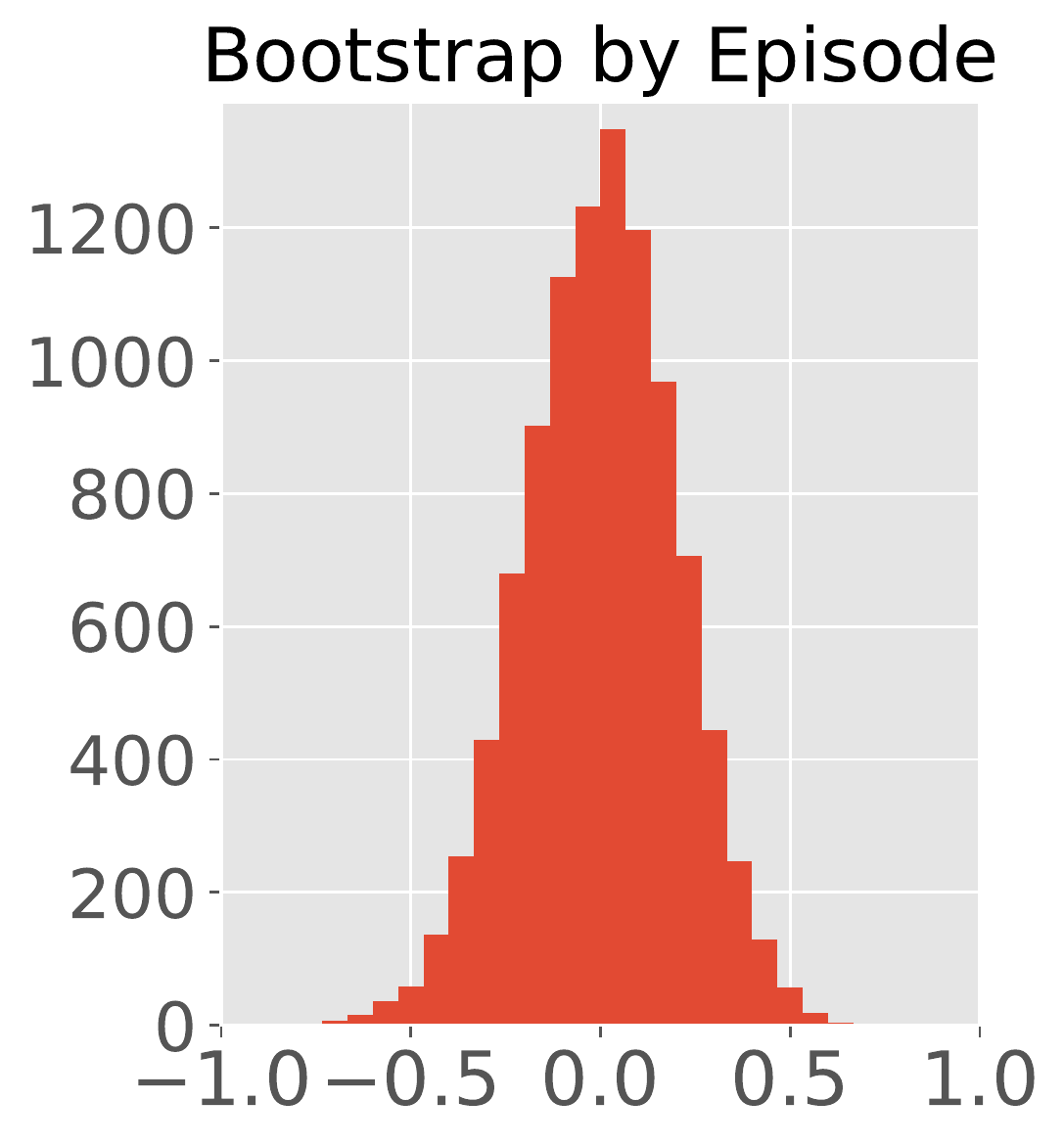}
\includegraphics[width=0.35\linewidth]{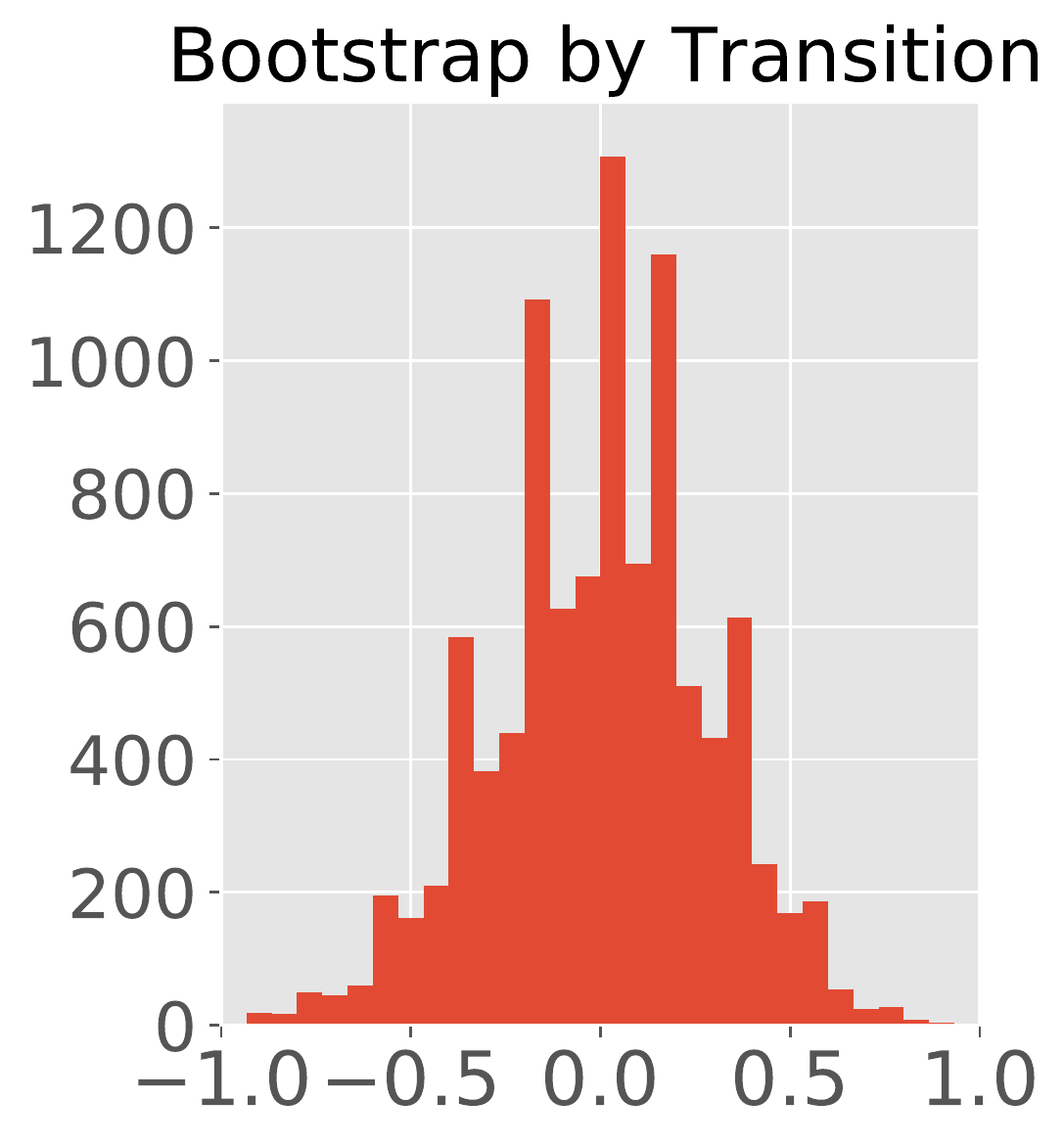}
\caption{\textbf{Bootstrap by episodes vs.  by sample transitions.} The first panel is the true FQE error distribution by Monte Carlo approximation. The second panel is the bootstrap distribution by episode while the third one is by sample transitions. Both behavior and target policies are the optimal policy. The number of Monte Carlo and bootstrap samples is 10000.}
\label{fig:distribution}
\end{figure}
We argue that bootstrapping by sample transitions may fail to correctly characterize the target error distribution of OPE. This is due to the in-episode dependence. To illustrate this phenomenon, we conduct numerical experiments using a toy Cliff Walking environment. We compare the true distribution of FQE error obtained by Monte Carlo sampling with error distributions obtained using bootstrapping FQE. Figure \ref{fig:distribution} clearly shows that the bootstrap distribution of $\hat{v}_{\pi}^*-\hat{v}_{\pi}$ (by episodes) closely approximates the true error distribution of $\hat{v}_{\pi}-v_{\pi}$, while the bootstrap distribution by sample transition is highly irregular and incorrect. This validates our belief that it is necessary to bootstrap by episodes and handle dependent data carefully for OPE.

\section{Asymptotic Distribution and Optimality of FQE}
Before analyzing the use of bootstrap, we first study the asymptotic properties of FQE estimators. 
For the sake of theoretical abstraction, we focus our analysis on the FQE with linear function approximation, because it is the most basic and universal function approximation.
We will show that the FQE error is asympotically normal and its asymptotic variance exactly matches the Cramér–Rao lower bound. 
All the proofs are deferred to Appendix \ref{proof:asy_normality_OPE} and \ref{sec:proof_efficiency}.

\textbf{Notations.} Given a feature map $\phi:\cS\times \cA\to \mathbb R^d$, we let $\cF$ be a linear function class spanned by $\phi$. Without loss of generality, we assume $\|\phi(s,a)\|_{\infty}\leq 1$ for any $(s,a)\in\cS\times \cA$. 
Define the Bellman operator for policy $\pi$ as $\cP^{\pi}:\mathbb R^{\cS\times \cA}\to\mathbb R^{\cS\times \cA}$ such that for any $f:\cS\times \cA\to\mathbb R, \cP^{\pi}f(s,a) = \mathbb E_{s'\sim P(\cdot|s,a), a'\sim \pi(\cdot|s')}[f(s',a')].$
Denote the expected covariance matrix induced by the feature $\phi$ as $\Sigma = \mathbb{E} [ \frac{1}{H} \sum_{h=1}^{H} \phi(s_{h}^1, a_{h}^1) \phi(s_{h}^1,a_{h}^1)^{\top} ],$
where $\mathbb{E}$ is the expectation over population distribution generated by the behavior policy.

\subsection{Asymptotic normality}
 
 We need a representation condition about the function class $\cF$, which will ensure sample-efficient policy evaluation via FQE.
\begin{condition}[\bf (Policy completeness)]\label{assum:completeness}
For any $f\in\cF$, we assume $\mathcal{P}^{\pi}f\in \cF$, and $r\in\cF.$
\end{condition}

Policy completeness requires the function class $\cF$ can well capture the Bellman operator. It is crucial for the estimation consistency of FQE \citep{le2019batch,duan2020minimax} and  implies the realizability condition  $Q_h^{\pi}\in\cF$ for $h\in[H]$. Recently, \citet{wang2020statistical} established a lower bound showing that the condition $Q_h^{\pi}\in\cF$ alone is not enough for sample-efficient OPE. Thus we need the policy completeness condition in order to leverage the generalizability of linear function class.

Next we present our first main result. The theorem presents the asymptotic normality of FQE with linear function approximation. For any $h_1\in[H], h_2\in[H]$, define the cross-time-covariance matrix as
\begin{equation*}
    \Omega_{h_1,h_2}= \mathbb E\Bigg[\frac{1}{H}\sum_{h'=1}^H\phi(s_{h'}^{1}, a_{h'}^{1})\phi(s_{h'}^{1}, a_{h'}^{1})^{\top}\varepsilon_{h_1,h'}^{1}\varepsilon_{h_2,h'}^{1}\Bigg],
\end{equation*}
where  $\varepsilon_{h_1,h'}^{1} =Q_{h_1}^{\pi}(s_{h'}^{1}, a_{h'}^{1}) - (r_{h'}^{1}+V_{h_1+1}^{\pi}(s_{h'+    1}^{1}))$.
\begin{theorem}[\bf Asymptotic normality of FQE]\label{thm:asy_normality_OPE}
Suppose $\lambda_{\min}(\Sigma)>0$ and Condition \ref{assum:completeness} holds. The FQE with linear function approximation is $\sqrt{N}$-consistent and asymptotically normal:
\begin{equation}\label{eqn:distri_FQE}
   \sqrt{N}\left(\widehat{v}_{\pi} - v_{\pi}\right)\overset{d}{\to} \cN(0, \sigma^2), \ \text{as} \ N\to\infty,
\end{equation}
where $\overset{d}{\to}$ denotes converging in distribution. The asymptotic variance $\sigma^2$ is given by
\begin{equation}\label{def:asy_variance}
\begin{split}
     \sigma^2 =& \sum_{h=1}^{H}(\nu_h^{\pi})^{\top}\Sigma^{-1}\Omega_{h,h}\Sigma^{-1}\nu_h^{\pi} \\
     &+ 2\sum_{h_1<h_2}(\nu_{h_1}^{\pi})^{\top}\Sigma^{-1}\Omega_{h_1,h_2}\Sigma^{-1}\nu_h^{\pi},
\end{split}
\end{equation} 
where $\nu_h^{\pi} = \mathbb E^{\pi}[\phi(s_h, a_h)|s_1\sim\xi_1]$.
\end{theorem}

The proof is based on a decomposition of the FQE error $\sqrt{N}(\widehat{v}_{\pi}-v_{\pi})$ into the sum of a primary term, which is a sum of the martingale differences, and two small-order terms that are asymptotically negligible. For the primary term, we utilize classical martingale central limit theorem \citep{mcleish1974dependent} to prove its asymptotic normality.

\begin{remark}
The second term on the right-hand side of Eq.~\eqref{def:asy_variance} (cross-product term) characterizes the dependency between two different fitted-Q steps. When considering a tabular time-inhomogeneous MDP that was used in \citet{yin2020asymptotically}, this cross-product term disappears and the asymptotic variance becomes
\begin{equation*}
    \sum_{h=1}^H\mathbb E\Bigg[\frac{\mu_h^{\pi}(s_h^1,a_h^1)^2}{\bar{\mu}_h(s_h^1,a_h^1)^2}(\varepsilon^1_{h,h})^2\Bigg],
\end{equation*}
where $\bar{\mu}_h$ is the marginal distribution of $(s_{h}^1, a_h^1)$ and $\mu_h^{\pi}$ is the marginal distribution of $(s_h,a_h)$ under policy $\pi$. This matches the asymptotic variance term in Remark 3.2 of \citet{yin2020asymptotically}.

\end{remark}

Next, we give a corollary about the joint asymptotic error distribution when evaluating multiple policies.
Denote $\Pi = \{\pi_1, \ldots, \pi_L\}$ as a set of target policies to evaluate and denote $\hat{v}_{\pi_k}$ as the FQE estimator of the policy $\pi_k$. For each $\pi_k\in\Pi$, let $\varepsilon_{h_1,h'}^{1,k} =Q_{h_1}^{\pi_k}(s_{h'}^{1}, a_{h'}^{1}) - (r_{h'}^{1}+V_{h_1+1}^{\pi_k}(s_{h'+    1}^{1}))$. For any $h_1\in[H], h_2\in[H]$, denote 
\begin{equation*}
    \Omega_{h_1,h_2}^{j,k}= \mathbb E\Bigg[\frac{1}{H}\sum_{h'=1}^H\phi(s_{h'}^{1}, a_{h'}^{1})\phi(s_{h'}^{1}, a_{h'}^{1})^{\top}\varepsilon_{h_1,h'}^{1,j}\varepsilon_{h_2,h'}^{1,k}\Bigg].
\end{equation*}
\begin{corollary}[\bf Multiple policies]
Suppose the conditions in Theorem \ref{thm:asy_normality_OPE} hold. 
The set of FQE estimators converge in distribution to a multivariate Gaussian distribution:
$$
\begin{gathered}
\begin{pmatrix} 
\sqrt{N}\left(\widehat{v}_{\pi_1} - v_{\pi_1}\right) \\
\vdots\\
\sqrt{N}\left(\widehat{v}_{\pi_L} - v_{\pi_L}\right) \end{pmatrix}
\overset{d}{\to} \cN(0, \Gamma),
\end{gathered}
$$
where the covariance matrix $\Gamma = (\sigma_{jk}^2)_{j,k=1}^L\in\mathbb R^{L\times L}$ with 
\begin{equation*}\label{def:asy_variance_multiple}
\begin{split}
     \sigma_{jk}^2=& \sum_{h=1}^{H}(\nu_h^{\pi_j})^{\top}\Sigma^{-1}\Omega_{h,h}^{j,k}\Sigma^{-1}\nu_h^{\pi_k} \\
     &+ 2\sum_{h_1<h_2}(\nu_{h_1}^{\pi_j})^{\top}\Sigma^{-1}\Omega_{h_1,h_2}^{j,k}\Sigma^{-1}\nu_h^{\pi_k}.
\end{split}
\end{equation*}
\end{corollary}

\subsection{Asymptotic efficiency}

An asymptotic efficient estimator has the minimal variance among all the unbiased estimator or its variance matches the Cramér–Rao bound asymptotically.

\begin{theorem}[\bf Linear Cramér–Rao lower bound]\label{thm:efficiency} Under Condition~\ref{assum:completeness} with linear function class, the variance of any unbiased OPE estimator is lower bounded by $\sigma^2$ defined in Eq.~\eqref{def:asy_variance}.
\end{theorem}

The above theorem implies FQE with linear function approximation is asymptotic efficient. 
 \citet{jiang2016doubly} derived the first Cramér–Rao lower bound for \emph{the tabular MDP} that depends the size of state and action spaces. Our lower bound is stronger in the sense that it only depends on the feature dimension $d$. \citet{kallus2020double} studied more general semiparametric efficiency bound but can not be directly applied to our case since they do not consider the policy completeness assumption.

\section{Distributional
Consistency of Bootstrapping FQE}

In this section, we show that the bootstrapping FQE method is distributionally consistent. More precisely, we prove that, the bootstrap distribution of $\sqrt{N}(\hat{v}_{\pi}^* - \hat{v}_{\pi})$, conditioned on data $\cD$, asymptotically imitates the true 
error distribution 
$\sqrt{N}(\hat{v}_{\pi} - v_{\pi})$. Consequently, we may use the method to construct confidence regions with  asymptotically correct and tight coverage. All the proofs are deferred to Appendix \ref{sec:bootstrap_consistency} and \ref{sec:proof_CI_consistency}. 

Suppose that the batch dataset $\cD$ is generated from a probability space $(\cX, \cA, \mathbb P_{\cD})$, and the bootstrap weight $W^*$ is from an independent probability space $(\cW, \Omega, \mathbb P_{W})$. Their joint probability measure is $\mathbb P_{\cD W^*} = \mathbb P_{\cD}\times \mathbb P_{W^*}$. Let $\mathbb P_{W^*|\cD}$ denote the conditional distribution once the dataset $\cD$ is given.
\begin{theorem}[\bf Distributional consistency]\label{thm:bootstrap_consistency}
Suppose the same assumptions in Theorem \ref{thm:asy_normality_OPE} hold. Conditioned on $\cD$, we have 
\begin{equation}\label{eqn:bootstrap_distribution}
  \sqrt{N}\big(\hat{v}_{\pi}^*- \hat{v}_{\pi}\big)\overset{d}{\to} \cN(0, \sigma^2), \ \text{as} \ N\to\infty,
\end{equation}
where $\sigma^2$ is defined in Eq.~\eqref{def:asy_variance}. Consequently, it implies
\begin{equation*}
\begin{split}
    \sup_{\alpha\in(0, 1)}\Big|\mathbb P_{W^*|\cD}\Big(&\sqrt{N}\big(\hat{v}_{\pi}^*- \hat{v}_{\pi}\big)\leq \alpha\Big) \\
    &- \mathbb P_{\cD}\Big(\sqrt{N}(\hat{v}_{\pi} - v_{\pi})\leq \alpha\Big)\Big|\to 0.
\end{split}
\end{equation*}
\end{theorem}

Note that the convergence in distribution result applies to the sequence of probability measures $\mathbb P_{W^*|\cD}$ where datasize grows to infinity.
The proof of Theorem \ref{thm:bootstrap_consistency} uses techniques that are different from classical analysis of supervised learning. 
This is because FQE is a fixed-point iteration type algorithm and it has no objective function to minimize directly. This poses some difficulties to apply conventional bootstrap analysis. Thus, our proof utilizes the equivalence between FQE and a model-based plug-in estimator described in Appendix \ref{sec:model-based}, together with the Mallows metric \citep{bickel1981some, freedman1981bootstrapping} and the multivariate delta theorem. 

Theorem \ref{thm:bootstrap_consistency} sets the theoretical foundation for using bootstrapping for off-policy inference.
Eq.~\eqref{eqn:distri_FQE} and Eq.~\eqref{eqn:bootstrap_distribution} together show that the bootstrap error distribution converges to the same limit as the target error distribution of FQE, which are both asymptotically efficient and match the Cramér–Rao lower bound. 

By using the distributional consistency of bootstrapping FQE, we may further construct consistent confidence intervals. Denote the lower $\delta$th quantile of bootstrap error distribution $q_{\delta}^{\pi} = \inf\{t:\mathbb P_{W^*|\cD}(\hat{v}_{\pi}^*-\hat{v}_{\pi}\leq t)\geq \delta\}$. Then we construct the $1-\delta$ confidence interval of the policy value by:
$\text{CI}(\delta) = [\hat{v}_{\pi}-q_{1-\delta/2}^{\pi}, \hat{v}_{\pi}-q_{\delta/2}^{\pi}]. 
$

We next establish that the coverage probability
of the percentile bootstrap confidence interval for $v_{\pi}$ converges to the nominal level as a consequence of Theorem \ref{thm:bootstrap_consistency} and the consistency of bootstrap moment estimation. 
\begin{corollary}[\bf Consistency of the coverage probability]\label{cor:CI_consistency}
Under the assumptions in Theorem \ref{thm:bootstrap_consistency}, we have as $N\to\infty$, 
$\mathbb P_{\cD W^*}(v_{\pi}\in\text{CI}(\delta))\to 1-\delta.
$
\end{corollary}

\begin{remark}
\citet{kostrikov2020statistical} proved the consistency of bootstrap confidence interval in the tabular case. In contrast, our result is more general. We establish the distributional consistency for OPE with function approximation.  
\end{remark}

\begin{corollary}[\bf Consistency of the moment estimation]\label{cor:moment}
Suppose the assumptions in Theorem \ref{thm:bootstrap_consistency} holds and $\limsup_{N\to\infty} \mathbb E_{W^*|\cD}[(\sqrt{N}(\hat{v}_{\pi}^*-\hat{v}_{\pi}))^q]<\infty$ for some $q>2$. Then we have for any $1\leq r<q$,
\begin{equation*}
    \mathbb E_{W^*|\cD}\Big[\big(\sqrt{N}(\hat{v}_{\pi}^*-\hat{v}_{\pi})\big)^r\Big]\to \int t^r {\rm d}\mu(t),
\end{equation*}
where $\mu(\cdot)$ is the distribution of $\cN(0,\sigma^2)$.
\end{corollary}

The consistency of bootstrap variance estimate is immediately implied by setting $r=2$.

\input{subsample_bootstrap}

\input{experiments}

\input{discussion}

\section*{Acknowledgements}
Csaba Szepesv\'ari gratefully  acknowledges  funding  from 
the Canada CIFAR AI Chairs Program, Amii and NSERC. Mengdi Wang gratefully acknowledges funding from the U.S. National Science Foundation (NSF) grant CMMI1653435, Air Force Office of Scientific Research (AFOSR) grant FA9550-19-1-020, and C3.ai DTI. We thank Ruiqi Zhang for pointing out several typos. 
\bibliography{ref}
\bibliographystyle{icml2021}
\clearpage
\onecolumn
\appendix
\input{appendix}

\end{document}

%% file: subsample_bootstrap.tex
\section{Subsampled Bootstrapping FQE}

Computing bootstrap-based quantities can be prohibitively demanding as the data size grows. 
Inspired by recent developments from statistics community \citep{kleiner2014scalable, sengupta2016subsampled}, we adapt a simple subsampled bootstrap procedure for FQE to accelerate the computation. 

\subsection{Subsampled bootstrap} 

Let the original dataset be $\cD = \{\cD_1,\ldots, \cD_K\}$. For any dataset $\tilde{\cD}$, we denote by $\hat{v}_{\pi}(\tilde{\cD})$ the FQE estimator based on dataset $\tilde{\cD}$ and $B$ as the number of bootstrap samples. The subsampled bootstrap includes the following three steps. For each $b\in[B]$, we first construct a random subset $\cD_{K,s}^{(b)}$ of $s$ episodes where each sample episode is drawn independently \emph{without replacement} from dataset $\cD$. Typically $s=K^{\gamma}$ for some $0<\gamma\leq 1$.
   Then we generate a resample set $\cD_{K,s}^{(b)*}$ of $K$ episodes where each sample episode is drawn independently \emph{with replacement} from $\cD_{K,s}^{(b)}$. Note that when $s=K$, $\cD_{K,s}^{(b)}$ is always equal to $\cD$ such that the subsampled bootstrap reduces to vanilla bootstrap.
  In the end, we compute $\varepsilon^{(b)} = \hat{v}_{\pi}(\cD_{K,s}^{(b)*})-\hat{v}_{\pi}(\cD^{(b)}_{K,s})$. 
Algorithm \ref{alg:BFQE} gives the full description.

\begin{remark}[\bf Computational benefit]
In Algorithm \ref{alg:BFQE}, although each run of FQE is still over a dataset of $K$ episodes, only $s$ of them are distinct. As a result, the runtime of running FQE on a bootstrapped set can be substantially reduced. 
With linear function approximation, one run of FQE requires solving $H$ least square problems. Thus the total runtime complexity of the subsampled bootstrapping FQE  is $O(B(K^{2\gamma}H^3d+Hd^3))$, where $0<\gamma<1$ controls the subsample size. When $\gamma$ is small, we achieve significant speedup by an order of magnitude improvements.
\end{remark}

\begin{small}
\begin{algorithm}[t]
\caption{Subsampled Bootstrapping FQE}
\begin{algorithmic}[1]\label{alg:BFQE}
\INPUT{Dataset $\cD=\{\cD_1,\ldots,\cD_K\}$, target policy $\pi$, 
confidence level $\delta$, subset size $s$, number of bootstrap samples $B$.}
\STATE Compute FQE estimator $\hat{v}_{\pi}(\cD)$ (Algorithm \ref{alg:FQE}).
\FOR{$b=1,\ldots, B$}
\STATE Build a random subset $\cD^{(b)}_{K,s}$. 
\STATE Feed $\cD^{(b)}_{K,s}$ to FQE and compute $\hat{v}_{\pi}(\cD^{(b)}_{K,s})$.
\STATE Generate a resample set  $\cD_{K,s}^{(b)*}$.
\STATE Compute  $\hat{v}_{\pi}(\cD_{K,s}^{(b)*})$.
\STATE Compute $\varepsilon^{(b)} = \hat{v}_{\pi}(\cD_{K,s}^{(b)*})-\hat{v}_{\pi}(\cD^{(b)}_{K,s})$.
\ENDFOR
\OUTPUT  $\{\varepsilon^{(1)},\ldots, \varepsilon^{(B)}\}$.
\end{algorithmic}
\end{algorithm}

\end{small}

\subsection{Off-policy inference via bootstrapping FQE}\label{sec:inference}
We describe how to conduct off-policy inference based on the output of Algorithm \ref{alg:BFQE}.
\begin{itemize}
    \item {\bf Bootstrap variance estimation.} To estimate the variance of FQE estimators, we calculate the bootstrap sample variance as 
 $$
        \widehat{\text{Var}}(\hat{v}_{\pi}(\cD)) = \frac{1}{B-1}\sum_{b=1}^B(\varepsilon^{(b)}-\bar{\varepsilon})^2,
$$
    where $\bar{\varepsilon} = \frac{1}{B}\sum_{b=1}^B\varepsilon^{(b)}$.
    \item {\bf Bootstrap confidence interval.}  Compute the $\delta/2$ and $1-\delta/2$ quantile of the empirical distribution $\{\varepsilon^{(1)},\ldots, \varepsilon^{(B)}\}$, denoted as $\hat{q}^{\pi}_{\delta/2},\hat{q}^{\pi}_{1-\delta/2}$ respectively. The percentile bootstrap confidence interval is $[\hat{v}_{\pi}(\cD) - \hat{q}^{\pi}_{1-\delta/2}, \hat{v}_{\pi}(\cD)-\hat{q}^{\pi}_{\delta/2}]$. 
    \item {\bf Bootstrap correlation estimation.} For any of two target policies $\pi_{1}$ and $\pi_{2}$, we want to estimate the Pearson correlation coefficient between their FQE estimators. The bootstrap sample correlation can be computed as $ \hat{\rho}(\hat{v}_{\pi_1}(\cD), \hat{v}_{\pi_2}(\cD)) =$
    \begin{equation*}
        \frac{\sum_{b=1}^B(\varepsilon_1^{(b)}-\bar{\varepsilon}_1)(\varepsilon_{2}^{(b)}-\bar{\varepsilon}_{2})}{\sqrt{\sum_{b=1}^B(\varepsilon_{1}^{(b)}-\bar{\varepsilon}_1)^2}\sqrt{\sum_{b=1}^B(\varepsilon_{2}^{(b)}-\bar{\varepsilon}_{2})^2}}.
    \end{equation*}

\end{itemize}

%% file: experiments.tex
\section{Experiments}
In this section, we numerically evaluate the proposed bootstrapping FQE method in several RL environments. 
For constructing confidence intervals, we fix the confidence level at $\delta = 0.1$. For estimating variance and correlations, we average the results over 200 trials. More details about the experiment are given in Appendix \ref{sec:experi_appendix}. 

\subsection{Experiment with tabular discrete environment}
We first consider the Cliff Walking environment \citep{sutton2018reinforcement}, with artificially added randomness to create stochastic transitions (see Appendix \ref{sec:experi_appendix} for details). The target policy is chosen to be a near-optimal policy, trained using Q-learning. Consider three choices of the behavior policy: the same as the target policy (on-policy), 0.1 $\epsilon$-greedy policy and soft-max policy with temperature $1.0$ based on the learned optimal Q-function. The results for soft-max policy and correlation estimation are deferred to Appendix \ref{sec:experi_appendix}.

We test three different methods. The first two methods are subsampled bootstraping FQE with subsample sizes $s=K$ (the vanilla bootstrap) and $s=K^{0.5}$ (the computational-efficient version), where $B=100$. The third method is the high-confidence off-policy evaluation (HCOPE) \citep{thomas2015high}, which we use as a baseline for comparison. HCOPE is a method for constructing off-policy confident interval for tabular MDP, and it is based concentration inequalities and has  provable coverage guarantee. We also compare these methods with the oracle confidence interval (which is the true distribution's quantile obtained by Monte Carlo simulation).

\begin{figure}[!t]
 \centering
 \includegraphics[width=0.49\linewidth]{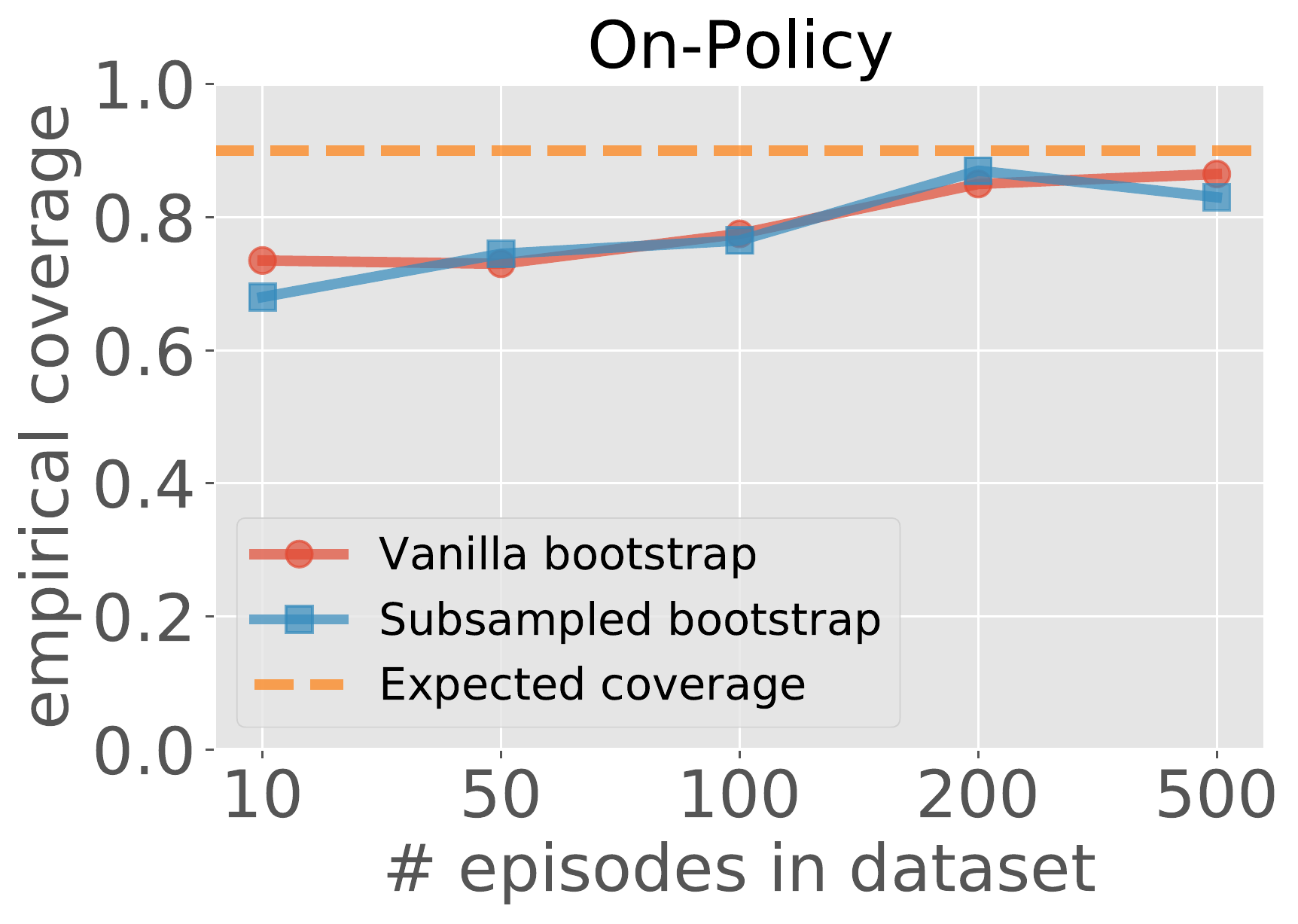}
  \includegraphics[width=0.49\linewidth]{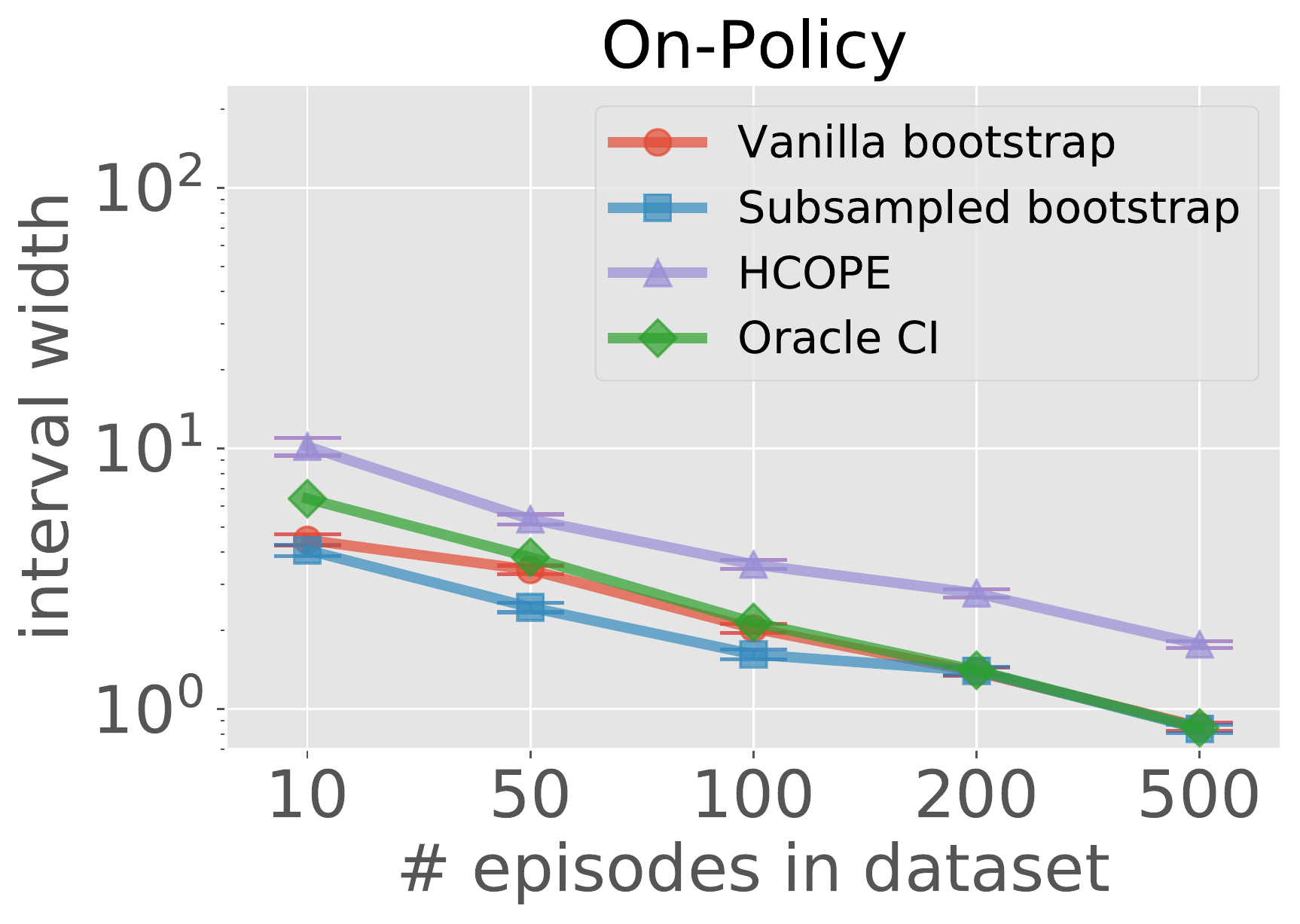}
   \includegraphics[width=0.49\linewidth]{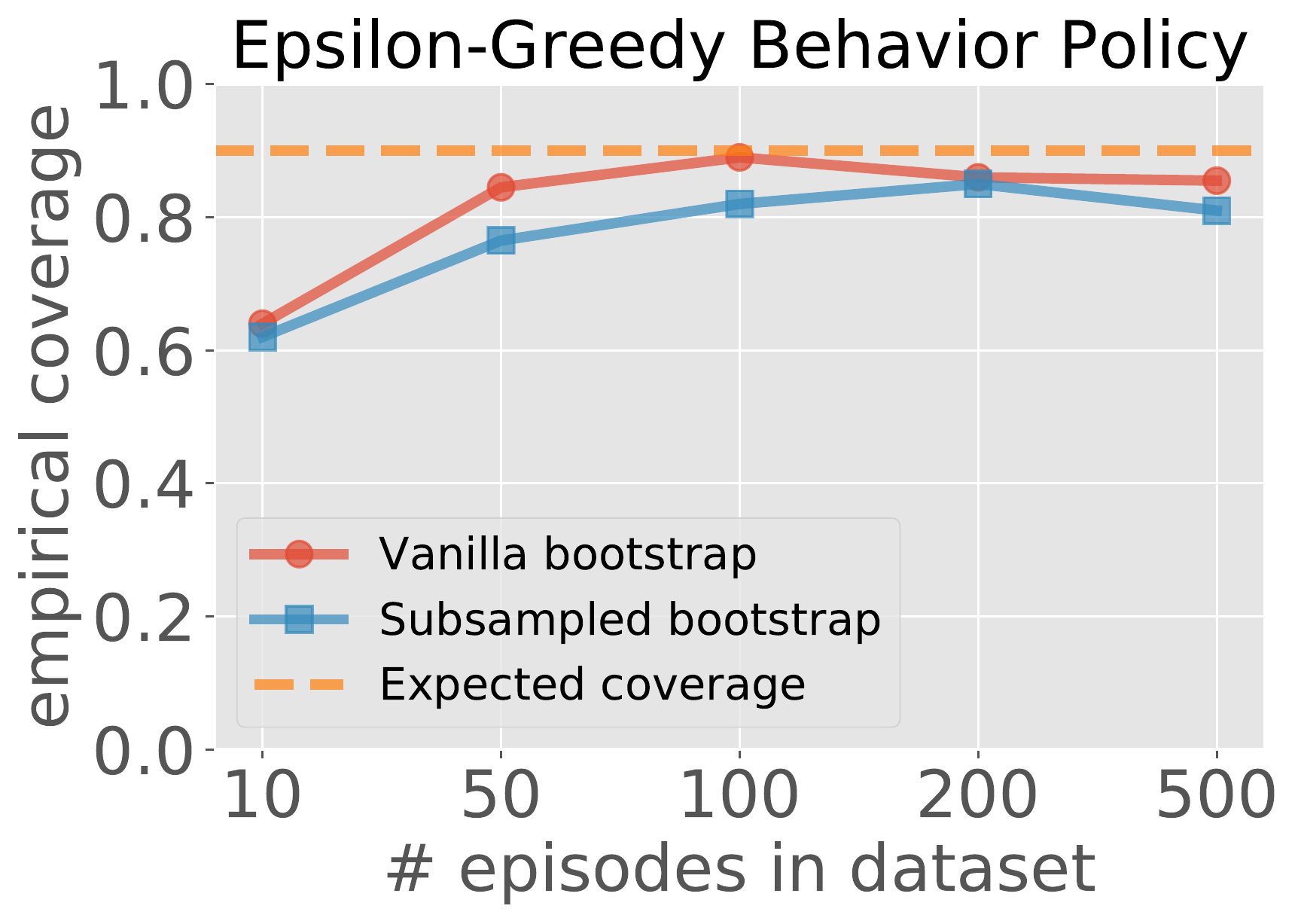}
  \includegraphics[width=0.49\linewidth]{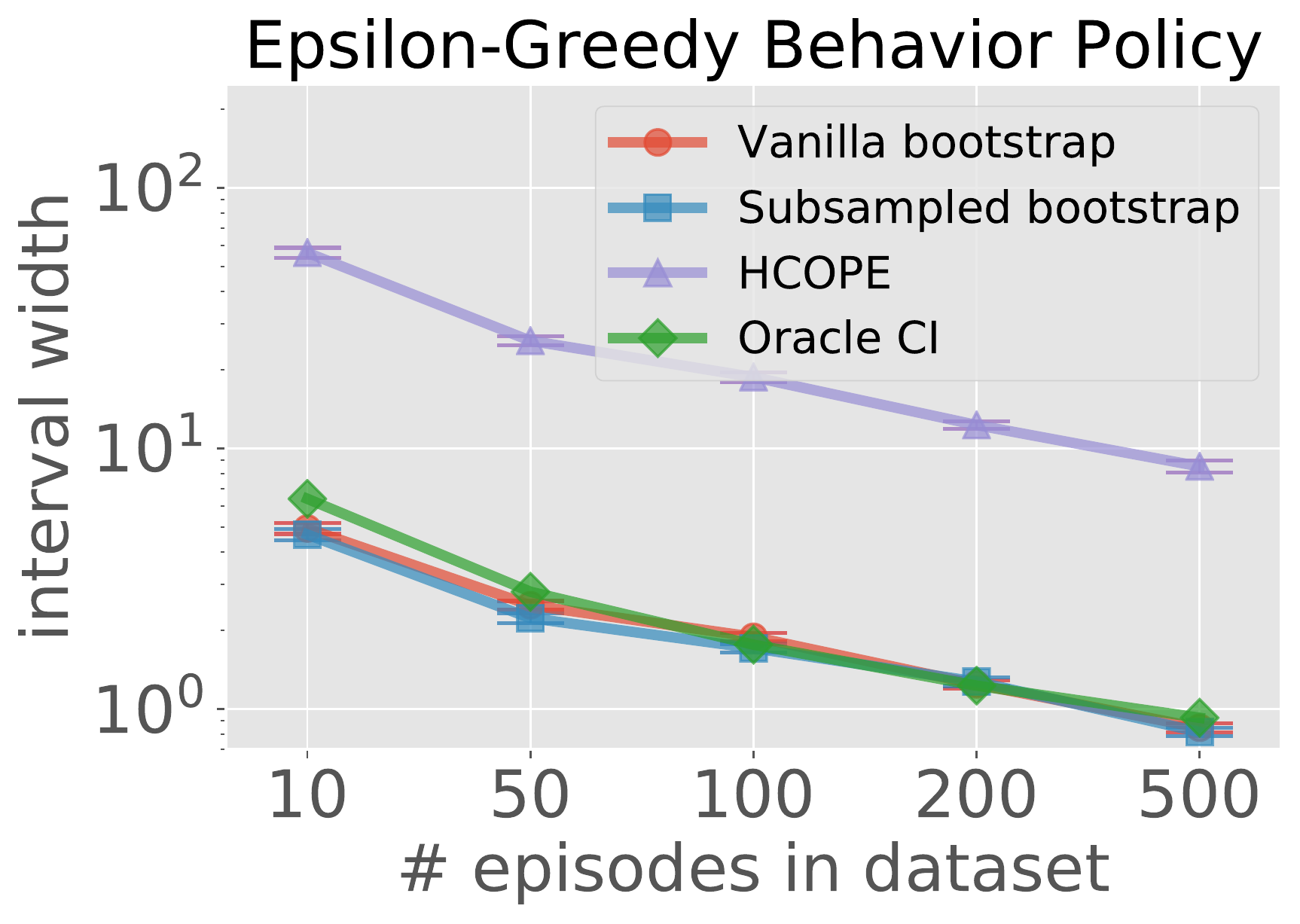}
\caption{\textbf{Off-policy CI for Cliff Walking.} Left: Empirical coverage probability of CI; Right: CI width under different behavior policies. Boostrapping-FQE confidence interval method demonstrates better and tighter coverage of the groundtruth. It closely resembles the oracle confidence interval which comes from the true error distribution.}
\label{fig:tabular_CI}
\end{figure}

\begin{figure}[!t]
 \centering
  \includegraphics[width=0.49\linewidth]{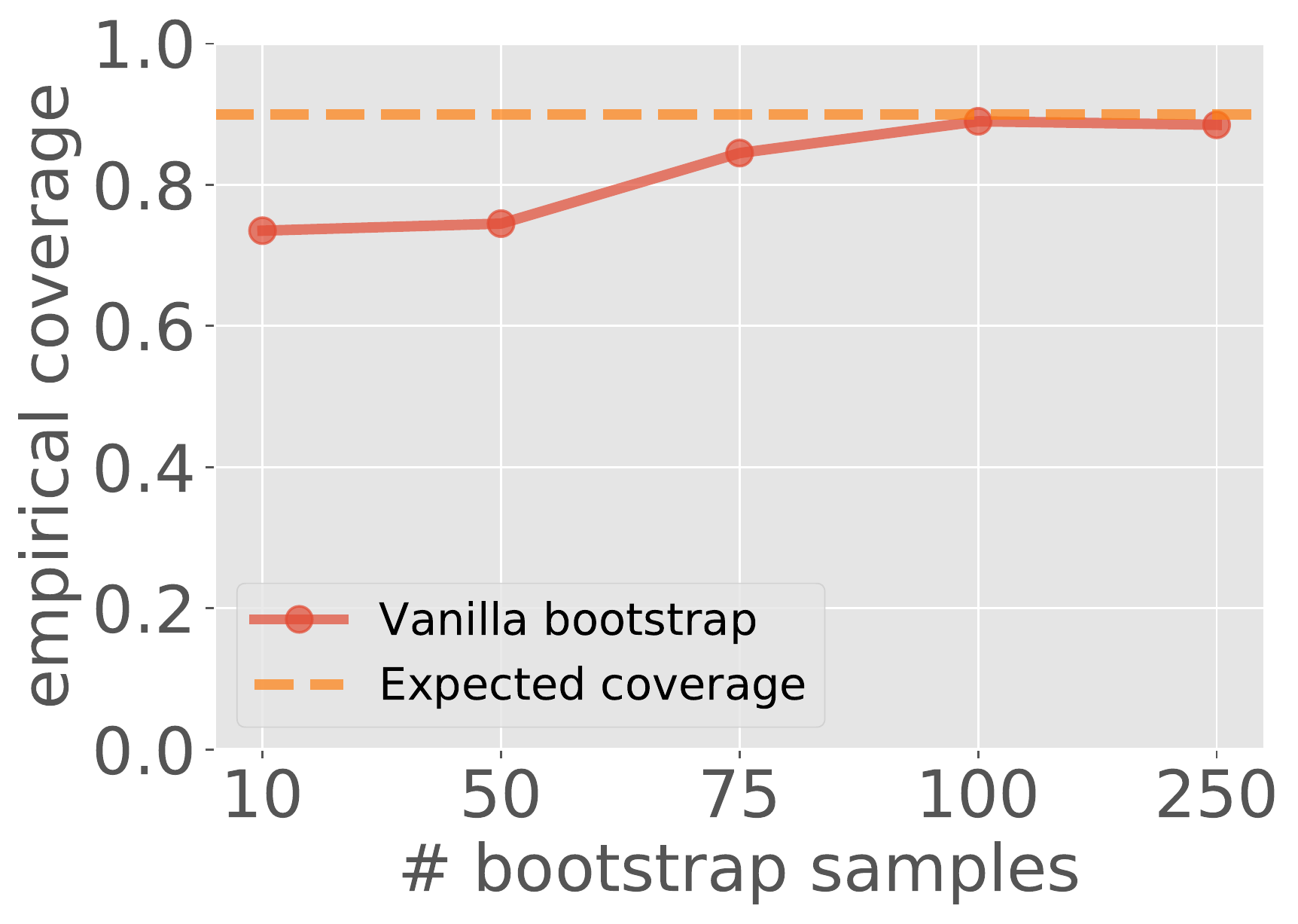}
 \includegraphics[width=0.49\linewidth]{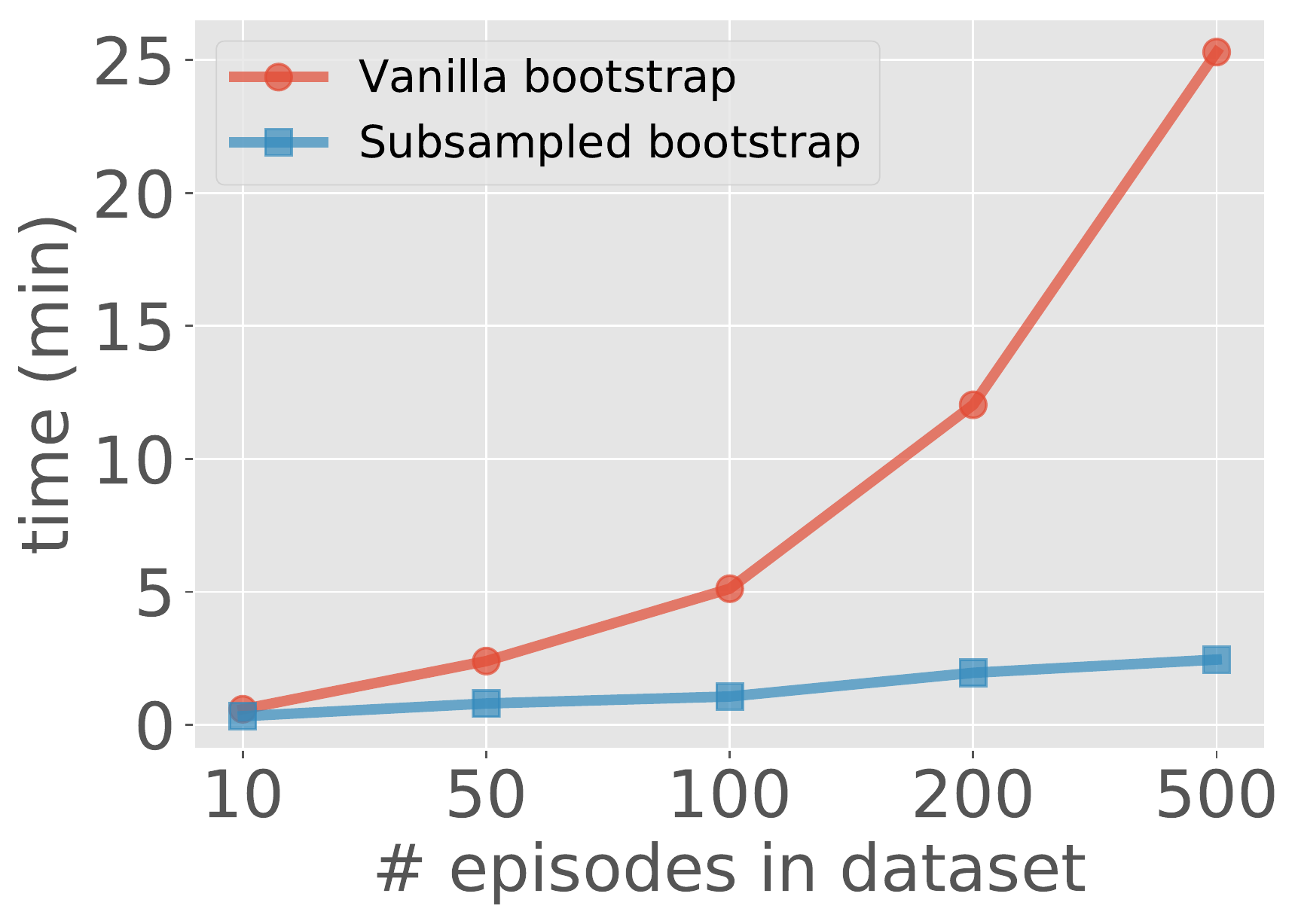}
\caption{\textbf{Sample and time efficiency of bootstrapping FQE.} Left: Empirical coverage of bootstrapping-FQE CI, as $\#$bootstrap samples increases. Right: Runtime of bootstrapping FQE, as datasize increases (with subsample size $s= K^{0.5}$).}
\label{fig:comp_time}
	\vspace{-0.2in}	
\end{figure}

\textbf{Coverage and tightness of off-policy confidence interval (CI).}
We study the empirical coverage probability and interval width with different number of episodes.
Figure \ref{fig:tabular_CI} shows the result under different behavior policies. In the left panel of Figure \ref{fig:comp_time}, we report the effect of the number of bootstrap samples on empirical coverage probability ($\epsilon$-greedy behavior policy, $K=100$).
 It is clear that the empirical coverage of our confidence interval based on bootstrapping FQE becomes increasingly close to the expected coverage ($=1-\delta$) as the number of episodes increases. The width of bootstrapping-FQE confidence interval is significantly tighter than that of the HCOPE and very close to the oracle one. It is worth noting that, even in the on-policy case, our bootstrap-based confidence interval still has a clear advantage over the concentration-based confidence interval. The advantage of our method comes from that it fully exploits the distribution information. However, bootstrap confidence interval tends to be under-estimate when the number of episodes is extremely small $(K=10)$. Thus we suggest the practitioner to use bootstrap methods when the sample size is moderately large $(K>50)$.
 
 Further, the subsampled bootstrapping FQE demonstrates a competitive performance as well as significantly reduced computation time. The saving in computation time becomes increasingly substantial as the data gets big; see the right panel of Figure \ref{fig:comp_time}. 

\begin{figure}[!t]
 \centering
 \includegraphics[width=0.45\linewidth]{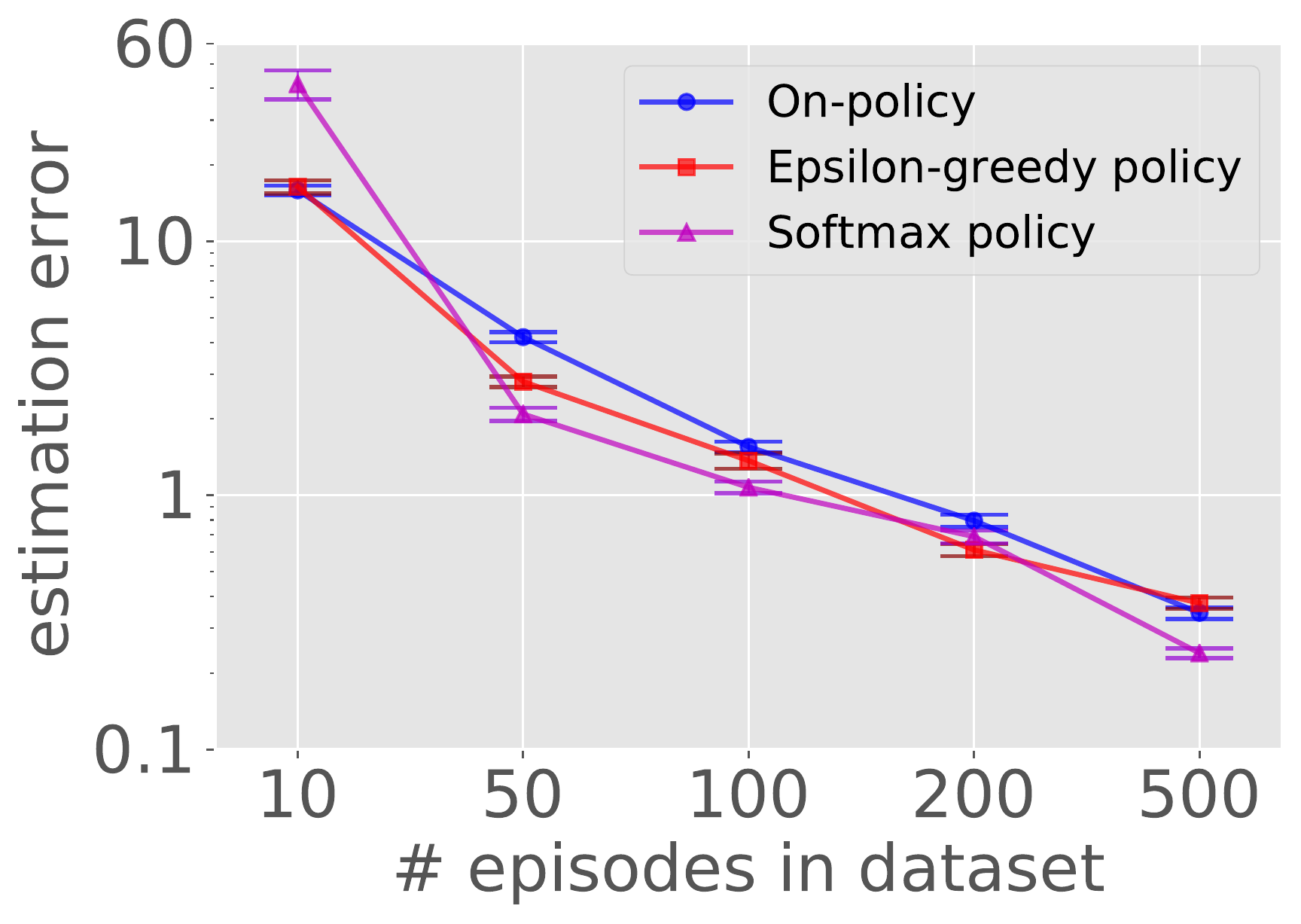}
 \includegraphics[width=0.45\linewidth]{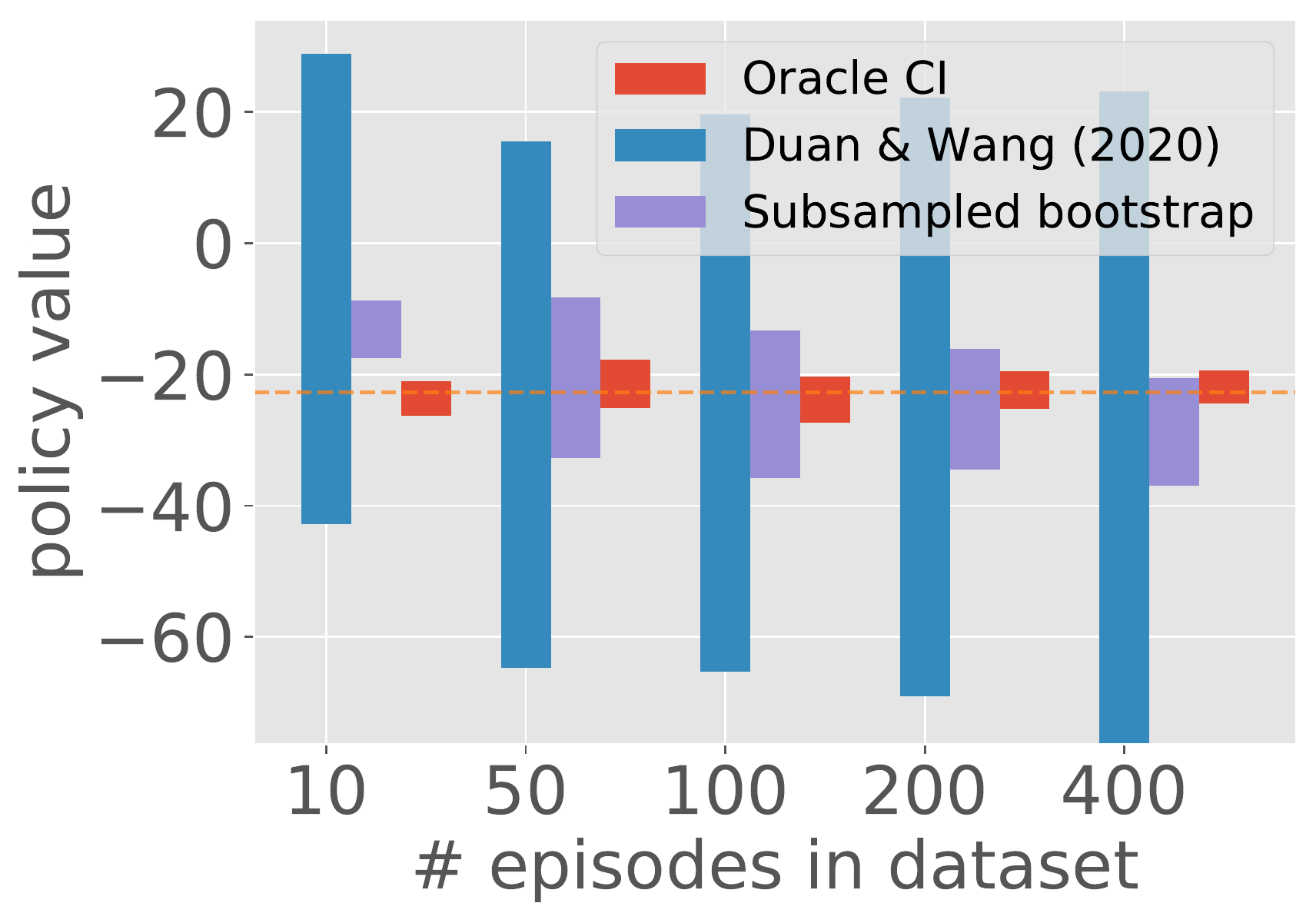}
\caption{\textbf{Bootstrapping for variance estimation and with function approximation} Left: Error of variance estimates, as data size increases. Right: Confidence interval constructed using bootstrapping FQE with linear function approximation.}

	\vspace{-0.2in}		
\label{fig:varince_est}

\end{figure}
\textbf{Bootstrapping FQE for variance estimation.}
We study the performance of variance estimation using subsampled bootstrapping FQE under three different behavior policies. We vary the number of episodes and the true $\text{Var}(\hat{v}_{\pi}(\cD))$ is computed through Monte Carlo method. We report the estimation error of $\hat{\text{Var}}(\hat{v}_{\pi}(\cD)) - \text{Var}(\hat{v}_{\pi}(\cD))$ across 200 trials in the left panel of Figure \ref{fig:varince_est}.

\subsection{Experiment with Mountain Car using linear function approximation}

Next we test the methods on the classical Mountain Car environment \citep{moore1990efficient} with linear function approximation. We artificially added a Gaussian random force to the car's dynamics to create stochastic transitions. For the linear function approximation, we choose 400 radial basis functions (RBF) as the feature map. The target policy is chosen as the optimal policy trained by Q-learning; and the behavior policy is chosen to be the $0.1$ $\epsilon$-greedy policy based on the learned optimal Q-function. 

For comparison, we compute an empirical Bernstein-inequality-based confidence interval \citep{duan2020minimax}, which to our best knowledge is the only provable CI based on FQE with function approximation (see Appendix \ref{sec:experi_appendix} for its detailed form). We also compute the oracle CI using Monte Carlo simulation. Figure \ref{fig:varince_est} right give all the results. According to the results, our method demonstrates good coverage of the groundtruth and is much tighter than the concentration-based CI, even both of them use linear function approximation. 

\subsection{Experiment with septic management using neural nets for function approximation}

Lastly, we consider a real-world healthcare problem for treating sepsis in the intensive care unit (ICU). We use the septic management simulator by \citet{oberst2019counterfactual} for our study. It simulates a patient's vital signs, e.g. the heart rate, blood pressure, oxygen concentration, and glucose levels, with three treatment actions (antibiotics, vasopressors, and mechanical ventilation) to choosen from at each time step. The reward is +1 when a patient is discharged and $-1$ if the patient reaches a life critical state.

We apply the bootstrapping FQE using neural network function approximator with three fully connected layers, where the first layer uses 256 units and a Relu activation function, the second layer uses 32 units and a Selu activation function, and the last layer uses Softsign. The network takes as input the state-action pair (a $11$-dim vector) and outputs a Q-value estimate. Let the behavior policy be the 0.15 $\epsilon$-greedy policy.

\begin{figure}[!t]
 \centering
  \includegraphics[width=0.45\linewidth]{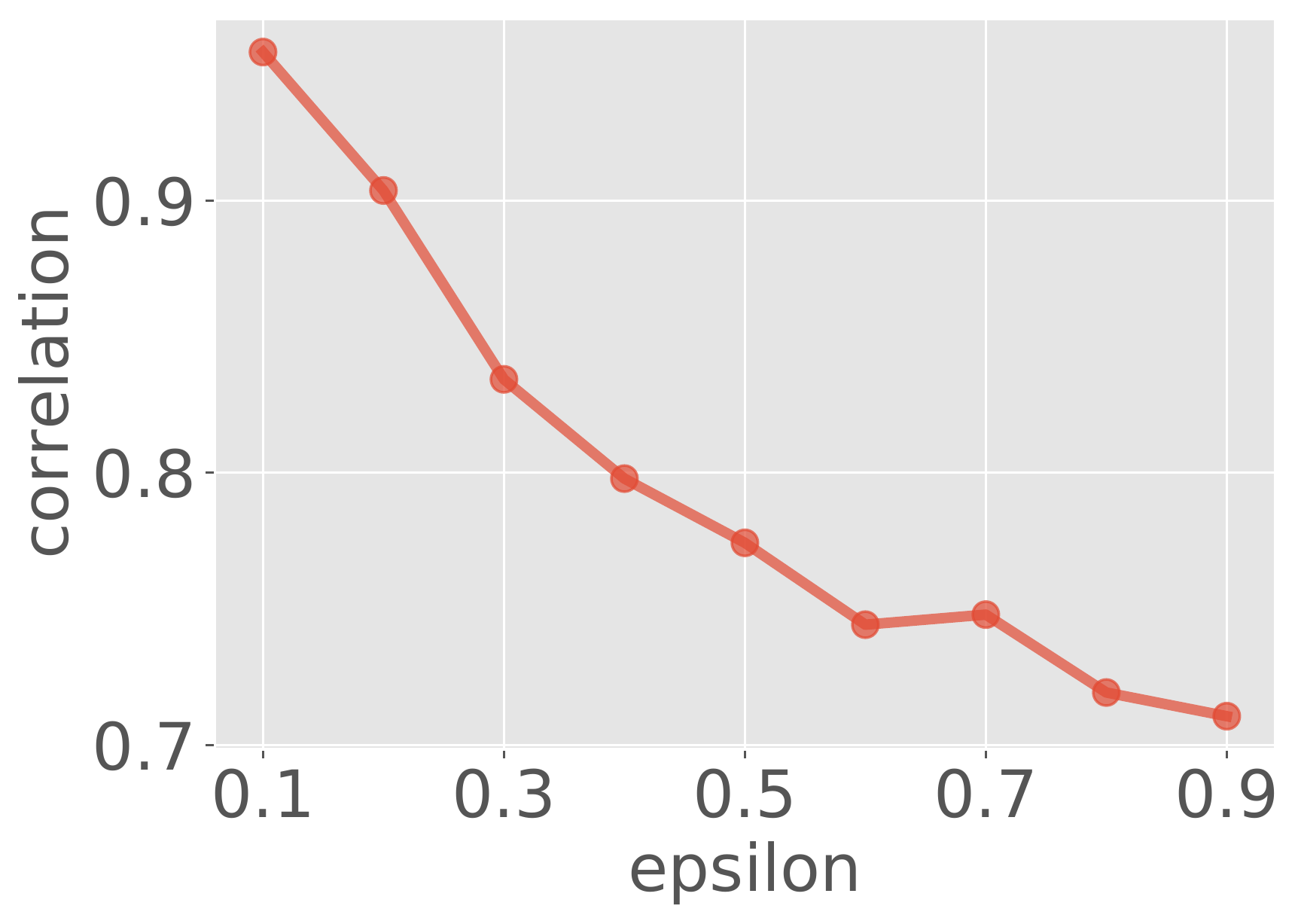}
 \includegraphics[width=0.45\linewidth]{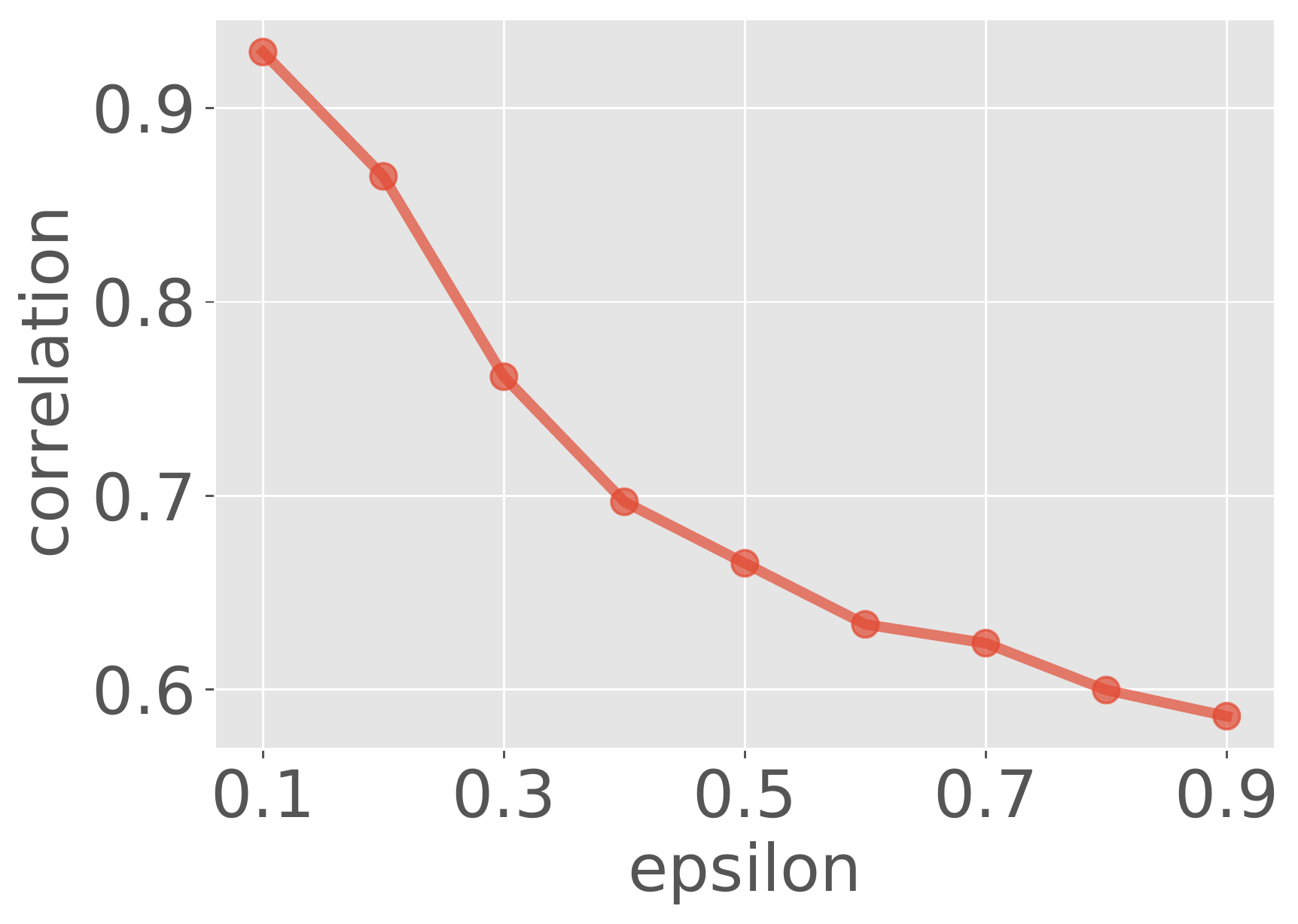}

\caption{\textbf{Bootstrapping FQE with neural nets for estimating the correlation between two FQE estimators.} The left panel is using 300 episodes, while the right panel is using 500 episodes.}
\label{fig:collection}
\end{figure}

\begin{figure}[!t]
 \centering
   \includegraphics[width=0.32\linewidth]{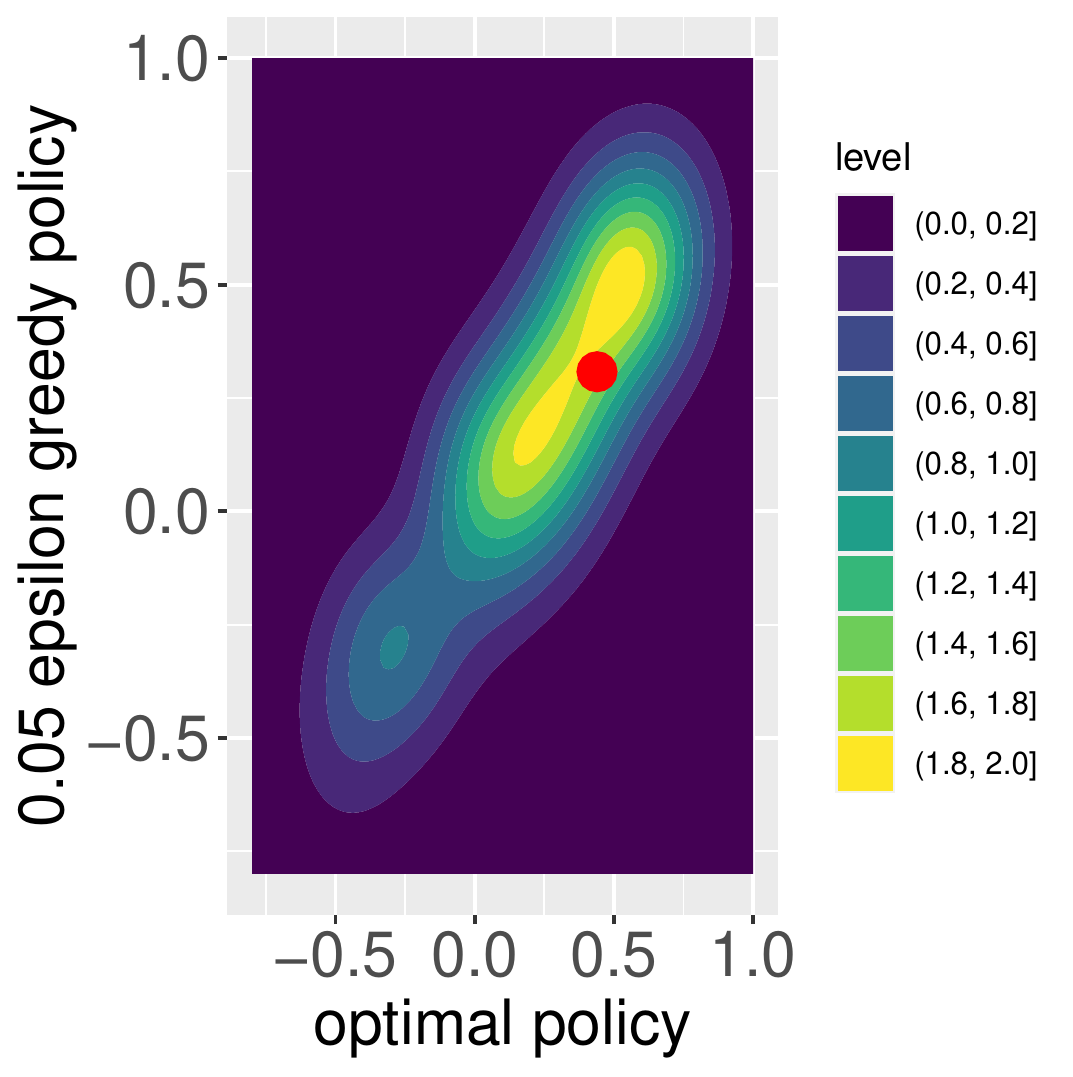}
   \includegraphics[width=0.32\linewidth]{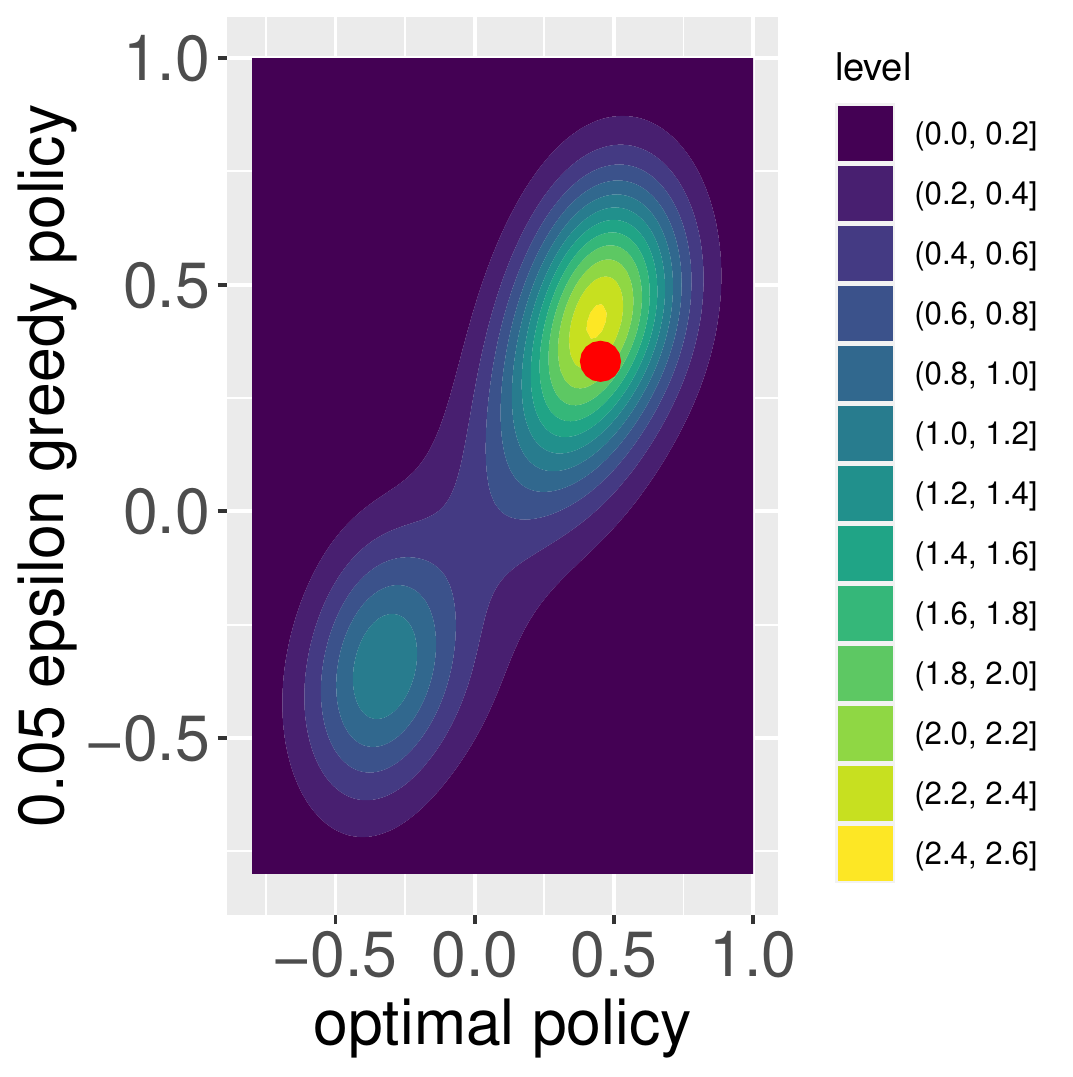}
   \includegraphics[width=0.32\linewidth]{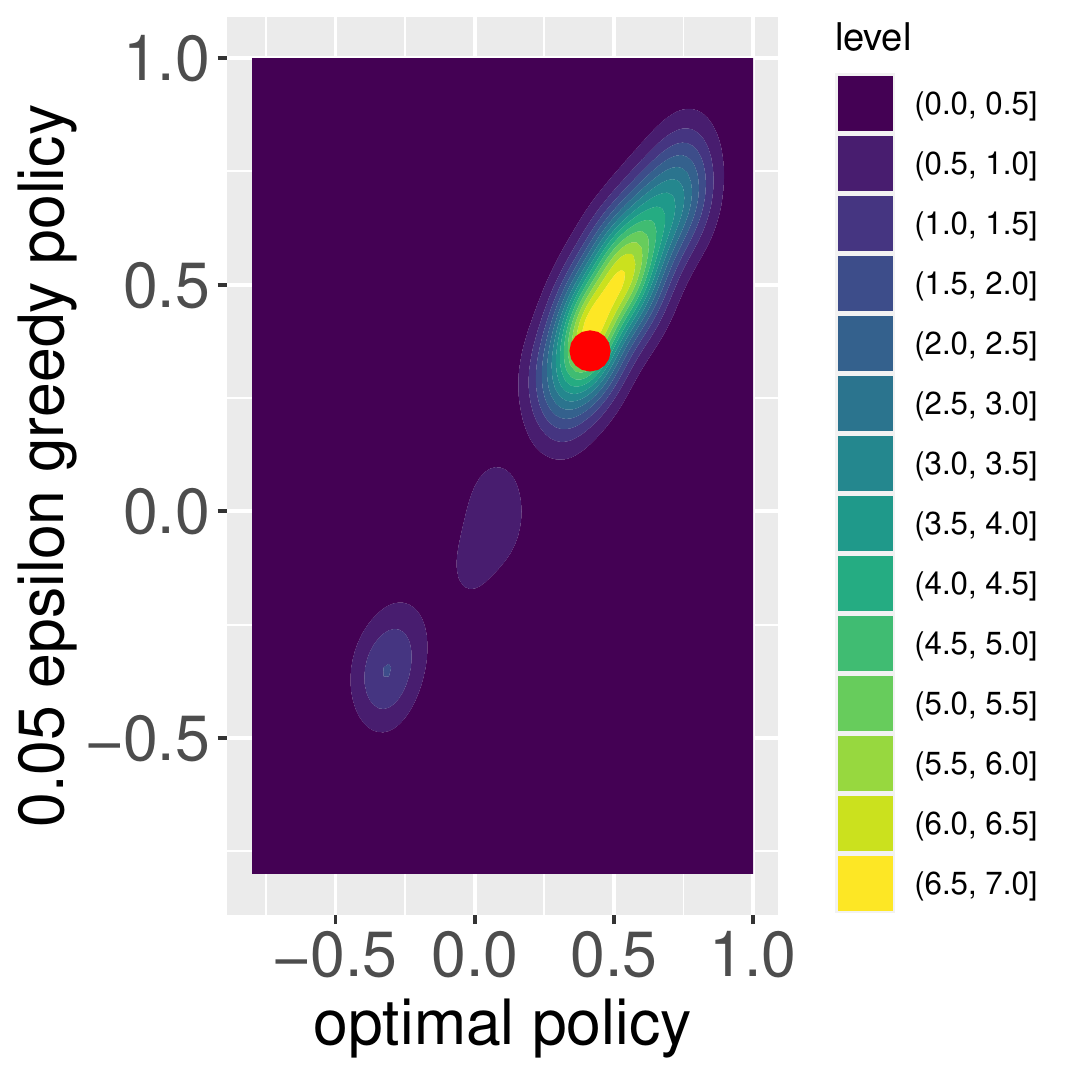}
\caption{\textbf{Estimated confidence region for evaluating two policies using bootstrapping FQE with neural networks.} Two target policies are optimal policy and 0.15 $\epsilon$-greedy policy. Red point are true values of those two target policies. From left to right, the sample sizes are $K=100, 300,500$.}
\label{fig:CR}
	\vspace{-0.2in}	
\end{figure}

We evaluate two policies based on the same set of data. This is very common in healthcare problem since we may have multiples treatments by the doctor. One target policy is fixed to be the optimal policy while we vary the other one with different $\epsilon$-greedy noise. We expect the correlation decreases as the difference between two target policies increases. Figure \ref{fig:collection} is well aligned with our expectation. In Figure \ref{fig:CR}, we plot the confidence region of two target policies obtained by bootstrapping FQE using neural networks.  According to Figures \ref{fig:collection} and \ref{fig:CR}, the bootstrapping FQE method can effectively construct confidence regions and correlation estimates, even when using neural networks for function approximation. These results suggest that the proposed bootstrapping FQE method reliably achieves off-policy inference, with more general function approximators.


%% file: discussion.tex
\section{Conclusion}

This paper studies bootstrapping FQE for statistical off-policy inference and establishes its asymptotic distributional consistency as a theoretical benchmark. Our experiments suggest that bootstrapping FQE is effective and efficient in a range of tasks, from tabular problems to continuous problems, with linear and neural network approximation. 

%% file: appendix.tex
\newpage
\appendix

\section{Proofs of Main Theorems}
\subsection{Full Algorithm of General FQE}\label{sec:alg_FQE}
\begin{small}
\begin{algorithm}
\caption{Fitted Q-Evaluation \citep{le2019batch}}
\begin{algorithmic}[1]\label{alg:FQE}
\INPUT{Dateset $\cD=\{\cD_1,\ldots,\cD_K\}$, target policy $\pi$, function class $\cF$, initial state distribution $\xi_0$.}
\STATE Initialize $\hat{Q}_{H+1}^{\pi} = 0$.
\FOR{$h = H, H-1,\ldots 1$}
\STATE Compute regression targets for any $k\in [K], h'\in[H]$:
$$
y_{h,h'}^{k}=r_{h'}^{k}+\int_{a}\hat{Q}^{\pi}_{h+1}(s_{h'+1}^{k}, a)\pi(a|s_{h'+1}^{k}){\rm d}a.
$$ 
\STATE Build training set $\{(s_{h'}^{k}, a_{h'}^{k}), y_{h,h'}^{k}\}_{k\in[K],h'\in[H]}$.
\STATE Solve a supervised learning problem:
\begin{equation*}\label{eqn:supervised}
\hat{Q}^{\pi}_h = \argmin_{f\in\cF}\left\{\frac{1}{K}\sum_{k=1}^K\frac{1}{H}\sum_{h'=1}^{H} \left(f(s_{h'}^{k}, a_{h'}^{k})-y_{h,h'}^{k}\right)^2+ \lambda \rho(f)\right\},
\end{equation*}
where $\rho(f)$ is a proper regularizer.
\ENDFOR
\OUTPUT  $\hat{v}^{\pi}= \int_{s}\int_a \hat{Q}^{\pi}_1(s, a)\xi_1(s)\pi(a|s){\rm d}s{\rm d}a$.
\end{algorithmic}
\end{algorithm}
\end{small}
We restate the full algorithm of FQE in Algorithm \ref{alg:FQE}. Here we simply assume the initial state distribution $\xi_1$ is known. In practice, we always have the access to sample from $\xi_1$ and thus we can approximate it by Monte Carlo sampling. 
\subsection{Equivalence between FQE and model-based plug-in estimator}\label{sec:model-based}

We show that the FQE in Algorithm \ref{alg:FQE} with linear function class $\cF$ is equivalent to a plug-in estimator. This equivalence is helpful to derive the asymptotic normality of FQE and bootstrapping FQE. Define
\begin{equation}\label{eqn:mean_embedding}
    \hat{M}_{\pi} = \hat{\Sigma}^{-1}\sum_{n=1}^N\phi(s_n, a_n)\phi^{\pi}(s_{n+1})^{\top}, \hat{R} = \hat{\Sigma}^{-1}\sum_{n=1}^Nr_n\phi(s_n,a_n), \hat{\Sigma} = \sum_{n=1}^N\phi(s_n,a_n)\phi(s_n,a_n)^{\top}+ \lambda I_{d},
\end{equation}
where $\phi^{\pi}(s) = \int_{a}\phi(s,a)\pi(a|s){\rm d}a$, $s_{N+1}$ is the terminal state and $\lambda$ is the regularization parameter. 
Choosing $\rho(f) = \lambda I$, the FQE is equivalent to, for $h=H,\ldots,1$,  $\hat{Q}_h(s,a) = \phi(s,a)^{\top}\hat{w}_h^{\pi}$ with
\begin{equation*}
    \begin{split}
        \hat{w}_h^{\pi} 
        &= \hat{\Sigma}^{-1}\sum_{n=1}^N\phi(s_n,a_n)\Big(r_n+\int_{a}\hat{Q}^{\pi}_{h+1}(s_{n+1}, a)\pi(a|s_{n+1}){\rm d}a\Big)\\
        & = \hat{\Sigma}^{-1}\sum_{n=1}^N\phi(s_n,a_n)\Big(r_n+\int_{a}\phi(s_{n+1}, a)^{\top}\hat{w}_{h+1}^{\pi}\pi(a|s_{n+1}){\rm d}a\Big)\\
        & = \hat{\Sigma}^{-1}\sum_{n=1}^N\phi(s_n,a_n)r_n + \hat{\Sigma}^{-1}\sum_{n=1}^N\phi(s_n,a_n)\phi^{\pi}(s_{n+1})^{\top} \hat{w}_{h+1}\\
        & = \hat{R} + \hat{M}_{\pi}\hat{w}_{h+1}^{\pi}.
    \end{split}
\end{equation*}
This gives us a recursive form of $\hat{w}_h^{\pi}$. Denoting $\hat{w}_{H+1}^{\pi} = 0$ and $\nu_1^{\pi} = \mathbb E_{s\sim \xi_1, a\sim \pi(\cdot|s)}[\phi(s, a)]$, the FQE estimator can be written into
\begin{equation}\label{eqn:plug-in}
\begin{split}
     \hat{v}_{\pi} &= \int_s\int_a \hat{Q}_1(s,a)\xi_1\pi(a|s) {\rm d}a{\rm d}s=(\nu_1^{\pi})^{\top}\hat{w}_1^{\pi} = (\nu_1^{\pi})^{\top}\sum_{h=0}^{H-1}(\hat{M}_{\pi})^h\hat{R}.
\end{split}
\end{equation}

On the other hand, from Condition \ref{assum:completeness}, there exists some $w_r, w_h^{\pi} \in \mathbb{R}^{d}$ such that $Q_h^{\pi}(\cdot, \cdot) = \phi(\cdot, \cdot)^{\top} w_h^{\pi}$ for each $h\in[H]$ and $r(\cdot, \cdot) = \phi(\cdot, \cdot)^{\top} w_r$ and there exists $M_{\pi}\in\mathbb R^{d\times d}$ such that $\phi(s,a)^{\top}M_{\pi} = \mathbb E[\phi^{\pi}(s')^{\top}|s,a]$. From Bellman equation and Condition \ref{assum:completeness},
\begin{equation}\label{eqn:Q_bellman}
\begin{split}
     Q_h^{\pi}(s,a) &= r(s,a) + \mathbb E\Big[\int_aQ_{h+1}^{\pi}(s', a)\pi(a|s'){\rm d}a|s,a\Big]\\
     &=\phi(s,a)^{\top}w_r + \phi(s,a)^{\top}\mathbb E[\phi^{\pi}(s')^{\top}|s,a]w_{h+1}^{\pi} = \phi(s,a)^{\top}\Big(w_r + M_{\pi}w_{h+1}^{\pi}\Big)\\
     &=\phi(s,a)^{\top}\sum_{h=0}^{H-h}(M_{\pi})^hw_r.
\end{split}
\end{equation}
Therefore, the true scalar value function can be written as 
\begin{equation*}
     v_{\pi} = \mathbb E_{s\sim\xi_1, a\sim \pi(\cdot|s)}\Big[Q_1^{\pi}(s,a)\Big]= (\nu_1^{\pi})^{\top}\sum_{h=0}^{H-1}(M_{\pi})^hw_r,
\end{equation*}
which implies Eq.~\eqref{eqn:plug-in} is a plug-in estimator.

\subsection{Proof of Theorem \ref{thm:asy_normality_OPE}: Asymptotic normality of FQE}\label{proof:asy_normality_OPE}
Recall $\nu_h^{\pi} = \mathbb E^{\pi}[\phi(x_h, a_h)|x_1\sim\xi_1]$ and denote $(\widehat{\nu}_h^{\pi})^{\top} = ( \nu_1^{\pi} )^{\top} \big(\widehat{M}_{\pi}\big)^{h-1}$. We follow Lemma B.3 in \citet{duan2020minimax} to decompose the error term into following three parts:
\begin{equation*}  \begin{aligned} 
	\sqrt{N}(v_{\pi} - \hat{v}_{\pi}) = E_1 + E_2 + E_3,
	\end{aligned} 
	\end{equation*}
	where
\begin{equation*}
    \begin{split}
        	& E_1 = \frac{1}{\sqrt{N}} \sum_{n=1}^N  \sum_{h=1}^H ( \nu_h^{\pi} )^{\top} \Sigma^{-1}\phi(s_n,a_n) \Big( Q_h^{\pi}(s_n,a_n) - \big( r_n + V_{h+1}^{\pi}(s_{n+1}) \big) \Big), \\
	& E_2 = \sum_{h=1}^{H} \Big( N ( \widehat{\nu}_h^{\pi} )^{\top} \widehat{\Sigma}^{-1} -  ( \nu_{h}^{\pi} )^{\top} \Sigma^{-1} \Big) \Big(  \frac{1}{\sqrt{N}} \sum_{n=1}^N  \phi(s_n,a_n) \Big( Q_h^{\pi}(s_n, a_n) - \big( r_n + V_{h+1}^{\pi}(s_{n+1}) \big) \Big)  \Big),  \\
	& E_3 = \lambda \frac{1}{\sqrt{N}} \sum_{h=0}^{H} ( \widehat{\nu}_h^{\pi} )^{\top} \widehat{\Sigma}^{-1} w_h^{\pi}. 
    \end{split}
\end{equation*}
To prove the asymptotic normality of $\sqrt{N}(v_{\pi} - \hat{v}_{\pi})$, we will first prove the asymptotic normality of $E_1$ and then show both $E_1$ and $E_2$ are asymptotically negligible. 

For $ n = 1,2,\ldots,N$, we denote
\begin{equation*}
    \begin{split}
       e_n = \frac{1}{\sqrt{N}}\sum_{h=1}^{H} (\nu_h^{\pi})^{\top} \Sigma^{-1} \phi(s_n,a_n) \Big(Q_h^{\pi}(s_n,a_n) - \big( r_{n+1} + V_{h+1}^{\pi}(s_{n+1}) \big)\Big).
    \end{split}
\end{equation*}
Then $E_1 = \sum_{n=1}^N e_n$. Define a filtration $\big\{ \mathcal{F}_n \big\}_{n=1,\ldots,N}$ with $\mathcal{F}_n$ generated by $(s_1, a_1, s_2), \ldots, (s_{n-1}, a_{n-1}, s_{n})$ and $(s_n,a_n)$.
From the definition of value function, it is easy to see
$\mathbb{E}\big[ e_n \, \big| \, \mathcal{F}_n \big] = 0$ that
implies that $\{e_n\}_{n\in[N]}$ is a martingale difference sequence. To show the asymptotic normality, we use the following martingale central limit theorem for triangular arrays. 
\begin{theorem}[Martingale CLT, Corollary 2.8 in \cite{mcleish1974dependent}]\label{thm:MCLT}
Let $\{X_{mn}; n=1,\ldots, k_m\}$ be a martingale difference array (row-wise) on the probability triple $(\Omega, \cF, P)$. Suppose $X_{mn}$ satisfy the following two conditions:
\begin{equation*}
    \begin{split}
        \max_{1\leq n\leq k_m}|X_{mn}|\overset{p}{\to} 0, \text{and} \ \sum_{n=1}^{k_m} X_{mn}^2\overset{p}{\to} \sigma^2,
    \end{split}
\end{equation*}
for $k_m\to \infty$. Then 
$\sum_{n=1}^{k_m} X_{mn} \overset{d}{\to} \cN(0, \sigma^2).
$
\end{theorem}
Recall that the variance $\sigma^2$ is defined as 
\begin{equation}\label{eqn:variance}
      \sigma^2 = \sum_{h=1}^{H}(\nu_h^{\pi})^{\top}\Sigma^{-1}\Omega_{h,h}\Sigma^{-1}\nu_h^{\pi} + 2\sum_{h_1<h_2}(\nu_{h_1}^{\pi})^{\top}\Sigma^{-1}\Omega_{h_1,h_2}\Sigma^{-1}\nu_{h_2}^{\pi},
\end{equation}
and for any $h_1\in[H], h_2\in[H]$, 
\begin{equation*}
    \Omega_{h_1,h_2}= \mathbb E\Big[\frac{1}{H}\sum_{h'=1}^H\phi(s_{h'}^{1}, a_{h'}^{1})\phi(s_{h'}^{1}, a_{h'}^{1})^{\top}\varepsilon_{h_1,h'}^{1}\varepsilon_{h_2,h'}^{1}\Big],
\end{equation*}
where  $\varepsilon_{h_1,h'}^{1} =Q_{h_1}^{\pi}(s_{h'}^{1}, a_{h'}^{1}) - (r_{h'}^{1}+V_{h_1+1}^{\pi}(s_{h'+1}^{1}))$. To apply Theorem \ref{thm:MCLT}, we let $k_m = N$, $X_{mn} = e_n$ and we need to verify the following two conditions:
\begin{equation}\label{eqn:clt_con1}
     \max_{1\leq n\leq N}\Big|   \sum_{h=1}^H ( \nu_h^{\pi} )^{\top} \Sigma^{-1} \left( \frac{1}{\sqrt{N}} \phi(s_n,a_n)  \Big(Q_h^{\pi}(s_n,a_n) - \big( r_{n+1} + V_{h+1}^{\pi}(s_{n+1}) \big)\Big) \right)\Big|\overset{p}{\to} 0, \ \text{as} \ N\to\infty,
\end{equation}
and
\begin{equation}\label{eqn:clt_con2}
    \sum_{n=1}^N\Big(\frac{1}{\sqrt{N}}\sum_{h=1}^{H} (\nu_h^{\pi})^{\top} \Sigma^{-1} \phi(s_n,a_n) \Big(Q_h^{\pi}(s_n,a_n) - \big( r_{n+1} + V_{h+1}^{\pi}(s_{n+1}) \big)\Big)\Big)^2 \overset{p}{\to} \sigma^2, \ \text{as} \ N\to\infty.
\end{equation}

\paragraph{Verify Condition \ref{eqn:clt_con1}:} Since $r \in [0,1]$, we have $r_n+V_{h+1}^{\pi}(s_{n+1}) \in [0,H-h]$. For any $n\in[N]$, we have 
\begin{equation*}
    \begin{split}
          &\left|\sum_{h=1}^H ( \nu_h^{\pi} )^{\top} \Sigma^{-1} \Bigg( \frac{1}{\sqrt{N}} \phi(s_n,a_n)  \Big(Q_h^{\pi}(s_n,a_n) - \big( r_{n+1} + V_{h+1}^{\pi}(s_{n+1}) \big)\Big) \Bigg)\right| \\
          &\leq \frac{1}{\sqrt{N}}\sum_{h=1}^{H} \Big| (\nu_h^{\pi})^{\top} \Sigma^{-1} \phi(s_n,a_n) \Big|\Big| Q_h^{\pi}(s_n, a_n) - \big( r_n+V_{h+1}^{\pi}(s_{n+1}) \big) \Big| 
          \\ 
          &\leq \frac{1}{\sqrt{N}} \sum_{h=1}^H (H-h+1) \Big| (\nu_h^{\pi})^{\top} \Sigma^{-1} \phi(s_n,a_n) \Big|.
    \end{split}
\end{equation*}
Note that $(\nu_h^{\pi})^{\top} \Sigma^{-1} \phi(s_n,a_n)$ is independent of $N$.
Then for fixed $d, H$, Condition \ref{eqn:clt_con1} is satisfied when $N\to\infty$.

\paragraph{Verify Condition \ref{eqn:clt_con2}:} Recall the definition of $\sigma^2$ in Eq.~\eqref{eqn:variance} and let $\sigma^2 = \sigma_1^2 + \sigma_2^2$ for
\begin{equation*}
    \begin{split}
         &\sigma_1^2 = \sum_{h=1}^{H}(\nu_h^{\pi})^{\top}\Sigma^{-1}\Omega_{h,h}\Sigma^{-1}\nu_h^{\pi}, \\
         &\sigma_2^2 = 2\sum_{h_1<h_2}(\nu_{h_1}^{\pi})^{\top}\Sigma^{-1}\Omega_{h_1,h_2}\Sigma^{-1}\nu_{h_2}^{\pi}.
    \end{split}
\end{equation*}
Using the following decomposition,
\begin{equation*}
\begin{split}
     &\sum_{n=1}^N\Big(\frac{1}{\sqrt{N}}\sum_{h=1}^{H} (\nu_h^{\pi})^{\top} \Sigma^{-1} \phi(s_n,a_n) \Big(Q_h^{\pi}(s_n,a_n) - \big( r_{n+1} + V_{h+1}^{\pi}(s_{n+1}) \big)\Big)\Big)^2\\ 
     =&\sum_{n=1}^N\frac{1}{N}\sum_{h=1}^H(\nu_h^{\pi})^{\top}\Sigma^{-1}\phi(s_n,a_n)\phi(s_n,a_n)^{\top}\Sigma^{-1}\nu_h^{\pi}\Big(Q_h^{\pi}(s_n,a_n)-(r_n+V_{h+1}^{\pi}(s_{n+1}))\Big)^2\\   &+\sum_{n=1}^N\frac{1}{N}2\sum_{h_1<h_2}(\nu_{h_1}^{\pi})^{\top}\Sigma^{-1}\phi(s_n,a_n)(\nu_{h_2}^{\pi})^{\top}\Sigma^{-1}\phi(s_n,a_n) \\
     &\cdot \Big(Q_{h_1}^{\pi}(s_n,a_n) - \big( r_{n+1} + V_{{h_1}+1}^{\pi}(s_{n+1}) \big)\Big) \Big(Q_{h_2}^{\pi}(s_n,a_n) - \big( r_{n+1} + V_{h_2+1}^{\pi}(s_{n+1}) \big)\Big).
\end{split}
\end{equation*}
We denote the first term as $I_1$, the second term as $I_2$ and separately bound $I_1-\sigma_1^2$ and $I_2-\sigma_2^2$ as follows:
\begin{itemize}
    \item  We rewrite $I_1$ in terms of episodes as 
\begin{equation*}
    I_1 = \sum_{h=1}^H(\nu_h^{\pi})^{\top}\Sigma^{-1/2}\Big(\frac{1}{K}\sum_{k=1}^K\frac{1}{H}\sum_{h=1}^H\Sigma^{-1/2}\phi(s_{h'}^k, a_{h'}^k)\phi(s_{h'}^k, a_{h'}^k)^{\top}(\varepsilon_{hh'}^k)^2\Sigma^{-1/2}\Big)\Sigma^{-1/2}\nu_h^{\pi}.
\end{equation*}
Moreover, denote 
\begin{equation*}
    Z_h = \Sigma^{-1/2}\mathbb E\Big[\frac{1}{H}\sum_{h'=1}^H\phi(s_{h'}^1, a_{h'}^1)\phi(s_{h'}^1, a_{h'}^1)^{\top}(\varepsilon_{hh'}^1)^2\Big]\Sigma^{-1/2}\in\mathbb R^{d\times d}.
\end{equation*}
Then we have 
\begin{equation*}
\begin{split}
    |I_1-\sigma_1^2| &= \sum_{h=1}^H(\nu_h^{\pi})^{\top}\Sigma^{-1/2}\Big(\frac{1}{K}\sum_{k=1}^K\frac{1}{H}\sum_{h'=1}^H\Sigma^{-1/2}\phi(s_{h'}^k, a_{h'}^k)\phi(s_{h'}^k, a_{h'}^k)^{\top}(\varepsilon_{hh'}^k)^2\Sigma^{-1/2}-Z_h\Big)\Sigma^{-1/2}\nu_h^{\pi}\\
    &\leq \sum_{h=1}^H\Big\|(\nu_h^{\pi})^{\top}\Sigma^{-1/2}\Big\|_2^2\Big\|\frac{1}{K}\sum_{k=1}^K\Big(\frac{1}{H}\sum_{h'=1}^H\Sigma^{-1/2}\phi(s_{h'}^k, a_{h'}^k)\phi(s_{h'}^k, a_{h'}^k)^{\top}(\varepsilon_{hh'}^k)^2\Sigma^{-1/2}-Z_h\Big)\Big\|_2,
\end{split}
\end{equation*}
where the last inequality is from Cauchy–Schwarz inequality. From Lemma \ref{lemma:Term1}, we reach $I_1\overset{p}{\to} \sigma_1^2$ as $K\to\infty$.
\item We rewrite $I_2$ as 
\begin{equation*}
    I_2 = 2\sum_{h_1<h_2}(\nu_{h_1}^{\pi})^{\top}\Sigma^{-1/2}\Big(\frac{1}{K}\sum_{k=1}^K\frac{1}{H}\sum_{h'=1}^H\Sigma^{-1/2}\phi(s_{h'}^k, a_{h'}^k)\phi(s_{h'}^k, a_{h'}^k)^{\top}\varepsilon_{h_1h'}^k\varepsilon_{h_2h'}^k\Sigma^{-1/2}\Big)\Sigma^{-1/2}\nu_{h_2}^{\pi},
\end{equation*}
and denote 
\begin{equation*}
     Z_{h_1h_2} = \Sigma^{-1/2}\mathbb E\Big[\frac{1}{H}\sum_{h'=1}^H\phi(s_{h'}^1, a_{h'}^1)\phi(s_{h'}^1, a_{h'}^1)^{\top}\varepsilon_{h_1h'}^1\varepsilon_{h_2h'}^1\Big]\Sigma^{-1/2}\in\mathbb R^{d\times d}.
\end{equation*}
Then we have 
\begin{equation*}
\begin{split}
    &|I_2-\sigma_2^2| = 2\sum_{h_1<h_2}(\nu_{h_1}^{\pi})^{\top}\Sigma^{-1/2}\Big(\frac{1}{K}\sum_{k=1}^K\frac{1}{H}\sum_{h'=1}^H\Sigma^{-1/2}\phi(s_{h'}^k, a_{h'}^k)\phi(s_{h'}^k, a_{h'}^k)^{\top}\varepsilon_{h_1h'}^k\varepsilon_{h_2h'}^k\Sigma^{-1/2}-Z_{h_1h_2}\Big)\Sigma^{-1/2}\nu_{h_2}^{\pi}\\
    &\leq 2\sum_{h_1<h_2}\Big\|(\nu_{h_1}^{\pi})^{\top}\Sigma^{-1/2}\Big\|_2\Big\|(\nu_{h_2}^{\pi})^{\top}\Sigma^{-1/2}\Big\|_2\Big\|\frac{1}{K}\sum_{k=1}^K\Big(\frac{1}{H}\sum_{h'=1}^H\Sigma^{-1/2}\phi(s_{h'}^k, a_{h'}^k)\phi(s_{h'}^k, a_{h'}^k)^{\top}\varepsilon_{h_1h'}^k\varepsilon_{h_2h'}^k\Sigma^{-1/2}-Z_{h_1h_2}\Big)\Big\|_2.
\end{split}
\end{equation*}
From Lemma \ref{lemma:Term1}, we reach $I_2\overset{p}{\to} \sigma_2^2$ as $K\to\infty$.
\end{itemize}
Putting the above two steps together, we have verified Condition \ref{eqn:clt_con2}. Then applying Theorem \ref{thm:MCLT} we obtain that $E_1\overset{d}{\to} \cN(0,\sigma^2)$.

On the other hand, according to Lemmas B.6, B.10 in \cite{duan2020minimax},
\begin{equation*}
\begin{split}
     &|E_2| \leq 15 \sqrt{(\nu_0^{\pi})^{\top} (\Sigma^{\pi})^{-1}  \nu_0^{\pi}} \cdot \big\| (\Sigma^{\pi})^{1/2} \Sigma^{-1/2} \big\|_2 \cdot \sqrt{C_1\kappa_1}(2+\kappa_2) \cdot \frac{\ln(8dH/\delta)dH^{3.5}}{\sqrt{N}}\\
     &|E_3| \leq \sqrt{( \nu_0^{\pi} )^{\top} (\Sigma^{\pi})^{-1} \nu_0^{\pi}} \cdot \big\| (\Sigma^{\pi})^{1/2} \Sigma^{-1/2} \big\|_2 \cdot \frac{5\ln(8dH/\delta) C_1dH^2}{\sqrt{N}},
\end{split}
\end{equation*}
with probability at least $1-\delta$ and $\kappa_1,\kappa_2$ are some problem-dependent constants that do not depend on $N$. When $N\to\infty$, both $|E_2|, |E_3|$ converge in probability to 0. By Slutsky's theorem, we have proven the asymptotic normality of $\sqrt{N}(v_{\pi} - \hat{v}_{\pi})$.

\hfill $\blacksquare$\\

\subsection{Proof of Theorem \ref{thm:efficiency}: Efficiency bound}\label{sec:proof_efficiency}

\input{lb.tex}
\subsection{Proof of Theorem \ref{thm:bootstrap_consistency}: Distributional  consistency of bootstrapping FQE}\label{sec:bootstrap_consistency}
 In order to simplify the derivation, we assume $\lambda = 0$ and the empirical covariance matrix $\sum_{n=1}^N\phi(s_n,a_n)\phi(s_n,a_n)^{\top}$ is invertible in this section since the effect of $\lambda$ is asymptotically negligible.
For a matrix $A\in\mathbb R^{m\times n}$, suppose the vec operator stacks the column of a matrix such that $\vec(A)\in\mathbb R^{mn\times 1}$. We use the equivalence form of FQE in Eq.~\eqref{eqn:plug-in} such that 
\begin{equation*}
    \hat{v}_{\pi} = (\nu_1^{\pi})^{\top}\sum_{h=0}^{H-1}(\hat{M}_{\pi})^h\hat{R}.
\end{equation*}
$\hat{M}_{\pi}$ can be viewed as the solution of the following multivariate linear regression:
\begin{equation*}
    \phi^{\pi}(s_{n+1})^{\top} = \phi(s_n,a_n)^{\top}M_{\pi} + \eta_n,
\end{equation*}
where $\eta_n =  \phi^{\pi}(s_{n+1})^{\top} - \phi(s_n,a_n)^{\top}M_{\pi}$. We first derive the asymptotic distribution of $\sqrt{N}\vec(\hat{M}_{\pi} - M_{\pi})$ that follows:
\begin{equation}\label{eqn:multi_linear}
\begin{split}
     \sqrt{N}\vec(\hat{M}_{\pi}-M_{\pi}) &= \vec\Big(\sqrt{N}\hat{\Sigma}^{-1}\sum_{n=1}^N\phi(s_n,a_n)\Big(\phi^{\pi}(s_n')^{\top}-\phi(s_n,a_n)^{\top}M_{\pi}\Big)\Big)\\
     &= (N\hat{\Sigma}^{-1}\otimes I_d)\frac{1}{\sqrt{K}}\sum_{k=1}^K\vec\Big(\frac{1}{\sqrt{H}}\sum_{h=1}^H\phi(s_h^k,a_h^k)\Big(\phi^{\pi}(s_h^{k'})^{\top}-\phi(s_h^k,a_h^k)^{\top}M_{\pi}\Big)\Big),
\end{split}
\end{equation}
where $\otimes$ is kronecker product. Define $\xi_h^k = \phi^{\pi}(s_{h+1}^k)^{\top}-\phi(s_h^k, a_h^k)^{\top}M_{\pi}$.
From the definition of $M_{\pi}$, it is easy to see
\begin{equation*}
    \mathbb E[\phi^{\pi}(s_{h+1}^{k})^{\top}|s_h^k, a_h^k]=\int_{s}\mathbb P(s|s_h^k, a_h^k)\int_a \pi(a|s)\phi(s,a)^{\top}{\rm d}a{\rm d}s = \phi(s_h^k, a_h^k)^{\top}M_{\pi}.
\end{equation*}
Again with martingale central limit theorem and independence between each episode, we have as $K\to\infty$,
\begin{equation}\label{eqn:CLT}
    \frac{1}{\sqrt{K}}\sum_{k=1}^K\vec\Big(\frac{1}{\sqrt{H}}\sum_{h=1}^H\phi(s_h^k,a_h^k)\xi_h^k\Big)\overset{d}{\to} N(0, \Delta),
\end{equation}
where $\Delta\in\mathbb R^{d^2\times d^2}$ is the covariance matrix defined as: for $j,k\in[d^2]$
\begin{equation}\label{eqn:covariance_matrix}
    \Delta_{jk} = \mathbb E\Big[\vec\Big(\frac{1}{\sqrt{H}}\sum_{h=1}^H\phi(s_h^k,a_h^k)\xi_h^k\Big)_j\vec\Big(\frac{1}{\sqrt{H}}\sum_{h=1}^H\phi(s_h^k,a_h^k)\xi_h^k\Big)_k\Big].
\end{equation}
   
Next we start to derive the conditional bootstrap asymptotic distribution. For notation simplicity, denote $\phi_{hk} = \phi(s_h^k, a_h^k)$ and $y_{hk} = \phi^{\pi}(s_{h+1}^{k})^{\top}$. We rewrite the dataset combined with feature map $\phi(\cdot,\cdot)$ such that $\cD_k = \{\phi_{hk}, y_{hk}, r_{hk}\}_{h=1}^H$. Recall that we bootstrap $\cD$ by episodes such that  each episode is sampled with replacement to form the starred data $\cD_k^* = \{\phi_{hk}^{*}, y_{hk}^{*}, r_{hk}^*\}_{h=1}^H$ for $k\in[K]$. More specifically, 
\begin{equation*}
   \phi_{hk}^{*} = \sum_{k=1}^KW_k^*\phi_{hk}, y_{hk}^{*} =  \sum_{k=1}^KW_k^*y_{hk}, r_{hk}^* = \sum_{k=1}^KW_k^*r_{hk},
\end{equation*}
where $W^* = (W_1^*, \ldots, W_K^*)$ is the bootstrap weight. For example, $W^*$ could be a multinomial random vector with parameters
$(K; K^{-1},\ldots,K^{-1})$ that forms the standard nonparametric bootstrap. Note that for different $h\in[H]$, they have the same bootstrap weight and given the original samples $\cD_1,\ldots, \cD_K$, the resampled vectors are independent.
Define the corresponding starred quantity $\hat{M}_{\pi}^*,\hat{R}^*$ as
\begin{equation*}
    \hat{M}_{\pi}^* = \hat{\Sigma}^{*-1}\sum_{k=1}^K\sum_{h=1}^K\phi_{hk}^*y_{hk}^*, \ \hat{R}^* = \hat{\Sigma}^{*-1}\sum_{k=1}^K\sum_{h=1}^Hr_{hk}^*\phi_{hk}^*,
\end{equation*}
where 
\begin{equation*}
    \hat{\Sigma}^* =\sum_{k=1}^K\sum_{h=1}^H\phi_{hk}^*\phi_{hk}^{*\top}.
\end{equation*}
We will derive the asymptotic distribution of $\sqrt{N}(\vec(\hat{M}_{\pi}^*-\hat{M}_{\pi}))$ by using the following decomposition:
\begin{equation*}
\begin{split}
     \sqrt{N}\vec(\hat{M}^*_{\pi}-\hat{M}_{\pi}) &= \sqrt{N} \vec\Big(\hat{\Sigma}^{*-1}\sum_{k=1}^K\sum_{h=1}^H\phi_{hk}^*(y_{hk}^*-\phi_{hk}^{*\top}\hat{M}_{\pi})\Big)\\
     & =(I_d\otimes N\hat{\Sigma}^{*-1})\vec\Big(\frac{1}{\sqrt{K}}\sum_{k=1}^K\frac{1}{\sqrt{H}}\sum_{h=1}^H\phi_{hk}^*(y_{hk}^*-\phi_{hk}^{*\top}\hat{M}_{\pi})\Big).
\end{split}
\end{equation*}
We denote 
$$
Z =\frac{1}{\sqrt{K}}\sum_{k=1}^K\frac{1}{\sqrt{H}}\sum_{h=1}^H\phi_{hk}(y_{hk}-\phi_{hk}M_{\pi}), Z^* =\frac{1}{\sqrt{K}}\sum_{k=1}^K\frac{1}{\sqrt{H}}\sum_{h=1}^H\phi_{hk}^*(y_{hk}^*-\phi_{hk}^*\hat{M}_{\pi}).
$$  
Both $Z$ and $Z^*$ are the sum of independent $d\times d$ random matrices. We prove the bootstrap consistency using the Mallows metric as a central tool. The Mallows metric, relative to the Euclidean norm $\|\cdot\|$, for two probability measures $\mu,\nu$ in $\mathbb R^d$ is defined as
\begin{equation*}
    \Lambda_{l}(\mu,\nu) = \inf_{U\sim\mu, V\sim\nu}\mathbb E^{1/l}(\|U-V\|^l),
\end{equation*}
where $U$ and $V$ are two random vectors that $U$ has law $\mu$ and $V$ has law $\nu$. For random variables $U, V$, we sometimes write $\Lambda_l(U,V)$ as the $\Lambda_l$-distance between the laws of $U$ and $V$. We refer \citet{bickel1981some, freedman1981bootstrapping} for more details about the properties of Mallows metric. Suppose the common distribution of original $K$ episodes $\{\cD_1,\ldots, \cD_K$ is $\mu$ and their empirical distribution is $\mu_K$. Both $\mu$ and $\mu_K$ are probability in $\mathbb R^{2Hd+H}$. From Lemma \ref{lemma:BF_8_4}, we know that $\Lambda_4(\mu_K,\mu)\to 0$ a.e. as $K\to\infty$.

\begin{itemize}
    \item \textbf{Step 1.} We prove $\widehat{\Sigma}^{*}/N$ converges in conditional probability to $\Sigma$.
From the bootstrap design, $\frac{1}{H}\sum_{h=1}^H\phi_{kh}^*\phi_{kh}^{*\top}$ is independent of $\frac{1}{H}\sum_{h=1}^H\phi_{k'h}^*\phi_{k'h}^{*\top}$ for any $k\neq k'$. According to Lemma \ref{lemma:BF8_6}, we have 
\begin{equation*}
\begin{split}
     \Lambda_1\Big(\sum_{k=1}^K\frac{1}{H}\sum_{h=1}^H\phi_{kh}^*\phi_{kh}^{*\top}, \sum_{k=1}^K\frac{1}{H}\sum_{h=1}^H\phi_{kh}\phi_{kh}^{\top}\Big)
     &\leq\sum_{k=1}^K \Lambda_1\Big(\frac{1}{H}\sum_{h=1}^H\phi_{kh}^*\phi_{kh}^{*\top}, \frac{1}{H}\sum_{h=1}^H\phi_{kh}\phi_{kh}^{\top}\Big)\\
     &= K \Lambda_1\Big(\frac{1}{H}\sum_{h=1}^H\phi_{kh}^*\phi_{kh}^{*\top}, \frac{1}{H}\sum_{h=1}^H\phi_{kh}\phi_{kh}^{\top}\Big).
\end{split}
\end{equation*}
Both sides of the above inequality are random variables such that the distance is computed between the conditional distribution of the starred quantity and the unconditional distribution of the unstarred quantity. Define a mapping $f:\mathbb R^{Hd}\to\mathbb R^{d\times d}$ such that for any $x_1,\ldots, x_{H}\in\mathbb R^d$,
\begin{equation*}
    f(x_1, \ldots, x_H) = \frac{1}{H}\sum_{h=1}^Hx_hx_h^{\top}.
\end{equation*}
From Lemma \ref{lemma:BF_8_5} with $f$, we have as $K$ goes to infinity
\begin{equation*}
    \Lambda_1\Big(\frac{1}{H}\sum_{h=1}^H\phi_{kh}^*\phi_{kh}^{*\top}, \frac{1}{H}\sum_{h=1}^H\phi_{kh}\phi_{kh}^{\top}\Big)\to 0\,.
\end{equation*}
This implies the conditional law of $\frac{1}{H}\sum_{h=1}^H\phi_{kh}^*\phi_{kh}^{*\top}$ is close to the unconditional law of $\frac{1}{H}\sum_{h=1}^H\phi_{kh}\phi_{kh}^{\top}$. By the law of large numbers:
\begin{equation}\label{eqn:empirical_cov}
    \frac{1}{K}\sum_{k=1}^K\frac{1}{H}\sum_{h=1}^H\phi_{kh}\phi_{kh}^{\top}\overset{p}{\to}\Sigma.
\end{equation}
This further implies the conditional on $\cD$, we have $\hat{\Sigma}^*/N\overset{p}{\to}\Sigma$.

\item \textbf{Step 2.} We prove $Z^* $ conditionally converges to a multivariate Gaussian distribution. From Lemma \ref{lemma:BF8_7},
\begin{equation*}
    \Lambda_2(\vec(Z^*), \vec(Z))^2\leq \Lambda_2\Big(\vec(\frac{1}{\sqrt{H}}\sum_{h=1}^H\phi_{hk}^*(y_{hk}^*-\phi_{hk}^*\hat{M}_{\pi})), \vec(\frac{1}{\sqrt{H}}\sum_{h=1}^H\phi_{hk}(y_{hk}-\phi_{hk}M_{\pi}))\Big)^2.
\end{equation*}
Using Lemma \ref{lemma:converge}, we have the right side converges to 0, a.e. as $K\to\infty$.
This means the conditional law of $\vec(Z^*)$ is close to the unconditional law of $\vec(Z)$, and the latter essentially converges to a multivariate Gaussian distribution with zero mean and covariance matrix $\Delta$ from Eq.~\eqref{eqn:CLT}. 
\end{itemize}
By Slutsky's theorem, we have conditional on $\cD$,
\begin{equation}\label{eqn:dis_M}
     \sqrt{N}\vec(\hat{M}^*_{\pi}-\hat{M}_{\pi})\overset{d}{\to} N\Big(0, (I_d\otimes\Sigma^{-1})\Delta(I_d\otimes\Sigma^{-1})\Big),
\end{equation}
where $\Delta$ is defined in Eq.~\eqref{eqn:covariance_matrix}.

According to the equivalence between FQE and plug-in estimator in Section \ref{sec:model-based},
\begin{equation*}
    \hat{v}_{\pi}^* = (\nu_1^{\pi})^{\top}\sum_{h=0}^{H-1}(\hat{M}_{\pi}^*)^h\hat{R}^*,  \hat{v}_{\pi} = (\nu_1^{\pi})^{\top}\sum_{h=0}^{H-1}(\hat{M}_{\pi})^h\hat{R}.
\end{equation*}
Define a function $g:\mathbb R^{d\times d}\to \mathbb R$ as
\begin{equation*}
    g(M):= (\nu_1^{\pi})^{\top}\sum_{h=0}^{H-1}(M)^h w_r.
\end{equation*} 
By the high-order matrix derivative \citep{petersen2008matrix}, we have 
\begin{equation*}
    \frac{\partial}{\partial M} (\nu_1^{\pi})^{\top}(M)^h w_r = \sum_{r=1}^{h-1}(M^r)^{\top}\nu_1^{\pi}w_r^{\top}(M^{h-1-r})^{\top}\in\mathbb R^{d\times d}.
\end{equation*}
This implies the gradient of $g$ at $\vec(M_{\pi})$
\begin{equation*}
\begin{split}
     \nabla g(\vec(M_{\pi})) &= \vec\Big(\sum_{h=0}^{H-1}\sum_{r=1}^{h-1}(M_{\pi}^r)^{\top}\nu_1^{\pi}w_r^{\top}(M_{\pi}^{h-1-r})^{\top}\Big)\\
     & = \vec\Big(\sum_{h=1}^{H}\nu_h^{\pi}w_r^{\top}\sum_{h'=1}^{H-h}(M_{\pi}^{h'-1})^{\top}\Big)\in\mathbb R^{d^2\times 1}.
\end{split}
\end{equation*}
Applying multivariate delta theorem (Theorem \ref{thm:delta}) for Eq.~\eqref{eqn:dis_M}, we have conditional on $\cD$
\begin{equation*}
    \sqrt{N}\Big(g(\hat{M}_{\pi}^*)- g(\hat{M}_{\pi})\Big) \overset{d}{\to} \cN\Big(0, \nabla^{\top} g(\vec(\hat{M}_{\pi}))(I_d\otimes\Sigma^{-1})\Delta(I_d\otimes\Sigma^{-1})\nabla g(\vec(\hat{M}_{\pi}))\Big),
\end{equation*}
where $\Delta$ is defined in Eq.~\eqref{eqn:covariance_matrix}. From Eq.~\eqref{eqn:empirical_cov}, we have $\hat{\Sigma}/N\overset{p}{\to} \Sigma$. Using Slutsky's theorem and Eqs.~\eqref{eqn:multi_linear}-\eqref{eqn:CLT}, we have
\begin{equation*}
    \sqrt{N}\Big(\vec(\hat{M}_{\pi}-M_{\pi})\Big)\overset{d}{\to}\cN(0, (I_d\otimes\Sigma^{-1})\Delta(I_d\otimes\Sigma^{-1})).
\end{equation*}
This further implies $\hat{M}_{\pi}\overset{p}{\to}M_{\pi}$. By continuous mapping theorem,
\begin{equation*}
    \sqrt{N}\Big(g(\hat{M}_{\pi}^*)- g(\hat{M}_{\pi})\Big) \overset{d}{\to} \cN\Big(0, \nabla^{\top} g(\vec(M_{\pi}))(I_d\otimes\Sigma^{-1})\Delta(I_d\otimes\Sigma^{-1})\nabla g(\vec(M_{\pi}))\Big).
\end{equation*}

Now we simplify the variance term as follows:
\begin{equation*}
    \begin{split}
      &\nabla^{\top} g(\vec(M_{\pi}))(I_d\otimes\Sigma^{-1})\Delta(I_d\otimes\Sigma^{-1})\nabla g(\vec(M_{\pi}))\\
     =&\sum_{h=1}^H(\nu_h^{\pi})^{\top}\Sigma^{-1}w_r^{\top}\sum_{h'=1}^{H-h} (M_{\pi})^{h'-1} \mathbb E\Big[\frac{1}{H}\sum_{h=1}^H\xi_{h}^{\top}\phi(s_{h}^1, a_{h}^1)\phi(s_{h}^1, a_{h}^1)^{\top}\xi_{h}\Big]\sum_{h=1}^H\sum_{h'=1}^{H-h} (M_{\pi})^{h'-1}w_r\Sigma^{-1}(\nu_h^{\pi})^{\top}\\
     =& \sum_{h=1}^H(\nu_h^{\pi})^{\top}\Sigma^{-1}w_r^{\top}\sum_{h'=1}^{H-h} (M_{\pi})^{h'-1} \mathbb E\Big[\frac{1}{H}\sum_{h=1}^H\xi_{h}^{\top}\phi(s_{h}^1, a_{h}^1)\phi(s_{h}^1, a_{h}^1)^{\top}\xi_{h}\Big]\sum_{h'=1}^{H-h} (M_{\pi})^{h'-1}w_r\Sigma^{-1}(\nu_h^{\pi})^{\top}\\
     &+ 2\sum_{h_1<h_2}(\nu_{h_1}^{\top})^{\top}\Sigma^{-1}w_r^{\top}\sum_{h'=1}^{H-h_1} (M_{\pi})^{h'-1} \mathbb E\Big[\frac{1}{H}\sum_{h=1}^H\xi_{h}^{\top}\phi(s_{h}^1, a_{h}^1)\phi(s_{h}^1, a_{h}^1)^{\top}\xi_{h}\Big]\sum_{h'=1}^{H-h_2} (M_{\pi})^{h'-1}w_r\Sigma^{-1}(\nu_{h_2}^{\top})^{\top}\\
     =& \sum_{h=1}^H(\nu_h^{\pi})^{\top}\Sigma^{-1}\mathbb E\Big[\frac{1}{H}\sum_{h=1}^Hw_r^{\top}\sum_{h'=1}^{H-h} (M_{\pi})^{h'-1} \xi_{h}^{\top}\phi(s_{h}^1, a_{h}^1)\phi(s_{h}^1, a_{h}^1)^{\top}\xi_{h}\sum_{h'=1}^{H-h} (M_{\pi})^{h'-1}w_r\Big]\Sigma^{-1}(\nu_h^{\pi})^{\top}\\
     &+ 2\sum_{h_1<h_2}(\nu_{h_1}^{\top})^{\top}\Sigma^{-1} \mathbb E\Big[\frac{1}{H}\sum_{h=1}^Hw_r^{\top}\sum_{h'=1}^{H-h_1} (M_{\pi})^{h'-1}\xi_{h}^{\top}\phi(s_{h}^1, a_{h}^1)\phi(s_{h}^1, a_{h}^1)^{\top}\xi_{h}\sum_{h'=1}^{H-h_2} (M_{\pi})^{h'-1}w_r\Big]\Sigma^{-1}(\nu_{h_2}^{\top})^{\top},
    \end{split}
\end{equation*}
where  $\xi_h^1 = \phi(s_h^1, a_h^1)^{\top}M_{\pi}-\phi^{\pi}(s_{h+1}^1)^{\top}$. Recall that we define
\begin{equation*}
\begin{split}
    \varepsilon_{h,h'}^{1} &=Q_{h}^{\pi}(s_{h'}^{1}, a_{h'}^{1}) - (r_{h'}^{1}+V_{h+1}^{\pi}(s_{h+    1}^{1}))\\
    &=\sum_{h=1}^{H-h_1}\Big(\phi(s_{h}^1, a_{h}^1)^{\top}M_{\pi}-\phi^{\pi}(s_{h+1}^1)^{\top}\Big)(M_{\pi})^{h'-1}w_r = \sum_{h=1}^{H-h}\xi_h^1(M_{\pi})^{h-1}w_r,
\end{split}
\end{equation*}
where the second equation is from Eq.~\eqref{eqn:Q_bellman}. This implies 
\begin{equation*}
    \begin{split}
    \nabla^{\top} &g(\vec(M_{\pi}))(I_d\otimes\Sigma^{-1})\Delta(I_d\otimes\Sigma^{-1})\nabla g(\vec(M_{\pi}))\\
 = &\sum_{h=1}^H(\nu_h^{\pi})^{\top}\Sigma^{-1}\mathbb E\Big[\frac{1}{H}\sum_{h'=1}^H\phi(s_{h'}^1, a_{h'}^1)\phi(s_{h'}^1, a_{h'}^1)^{\top}( \varepsilon_{h,h'}^{1})^2\Big]\Sigma^{-1}(\nu_h^{\pi})^{\top}\\
        &+ 2\sum_{h_1<h_2}(\nu_{h_1}^{\top})^{\top}\Sigma^{-1} \mathbb E\Big[\frac{1}{H}\sum_{h'=1}^H\phi(s_{h'}^1, a_{h'}^1)\phi(s_{h'}^1, a_{h'}^1)^{\top}\varepsilon_{h_1,h'}^{1}\varepsilon_{h_2,h'}^{1}\Big]\Sigma^{-1}(\nu_{h_2}^{\top})^{\top}= \sigma^2.
    \end{split}
\end{equation*}
Therefore, we have proven that
\begin{equation*}
    \sqrt{N}\Big(g(\hat{M}_{\pi}^*)- g(\hat{M}_{\pi})\Big) \overset{d}{\to} \cN\big(0, \sigma^2\big).
\end{equation*}

On the other hand, 
\begin{equation*}
    \begin{split}
        \hat{R}^* = (\hat{\Sigma}^*)^{-1}\sum_{k=1}^K\sum_{h=1}^Hr_{hk}^*\phi_{hk}^* = KH(\hat{\Sigma}^*)^{-1}\frac{1}{K}\sum_{k=1}^K\frac{1}{H}\sum_{h=1}^Hr_{hk}^*\phi_{hk}^*.
    \end{split}
\end{equation*}
Using Lemma \ref{lemma:BF8_6}, we have 
\begin{equation*}
    \Lambda_1\Big(\frac{1}{K}\sum_{k=1}^K\frac{1}{H}\sum_{h=1}^Hr_{hk}^*\phi_{hk}^*, \frac{1}{K}\sum_{k=1}^K\frac{1}{H}\sum_{h=1}^Hr_{hk}\phi_{hk}\Big)\leq  \Lambda_1\Big(\frac{1}{H}\sum_{h=1}^Hr_{hk}^*\phi_{hk}^*, \frac{1}{H}\sum_{h=1}^Hr_{hk}\phi_{hk}\Big).
\end{equation*}
The right hand side of the display goes to 0 as $K\to\infty$. From the law of large number,
\begin{equation*}
    \frac{1}{K}\sum_{k=1}^K\frac{1}{H}\sum_{h=1}^Hr_{hk}\phi_{hk} \overset{p}{\to} \mathbb E\Big[\frac{1}{H}\sum_{h=1}^H\phi(s_h^1,a_h^1)\phi(s_h^1,a_h^1)^{\top}\Big]w_r
\end{equation*}
Combining with the fact that the conditional laws of $\hat{\Sigma}^*$ concentrates around $\Sigma$, this ends the proof.
\hfill $\blacksquare$\\

\subsection{Proofs of Corollary \ref{cor:CI_consistency} and Corollary \ref{cor:moment}}\label{sec:proof_CI_consistency}

We prove the consistency of bootstrap confidence interval by using Lemma 23.3 in \citet{van2000asymptotic}. Suppose $ \Psi(t) = \mathbb P(\cN(0,\sigma^2)\leq t)$. Combining Theorem \ref{thm:asy_normality_OPE} and Theorem \ref{thm:bootstrap_consistency}, we have 
\begin{equation*}
    \begin{split}
       & \mathbb P_{\cD}\Big(\sqrt{N}(\hat{v}_{\pi}-v_{\pi})\leq t\Big)\to \Psi(t), \ \mathbb P_{W^*|\cD}\Big(\sqrt{N}(\hat{v}_{\pi}^*-\hat{v}_{\pi})\leq t\Big)\to\Psi(t).
    \end{split}
\end{equation*}
Using the quantile convergence theorem (Lemma 21.1 in \citet{van2000asymptotic}), it implies $q^{\pi}_{\delta}\to \Psi^{-1}(t)$ almost surely. Therefore,
\begin{equation*}
    \begin{split}
        \mathbb P_{\cD W^*}\Big(v_{\pi}\leq \hat{v}_{\pi}-q_{\delta/2}^{\pi}\Big) &= \mathbb P_{\cD W^*}\Big(\sqrt{N}(\hat{v}_{\pi}-v_{\pi})\geq q_{\delta/2}^{\pi}\Big)\\
        &\to \mathbb P_{\cD W^*}\Big(\cN(0,\sigma^2)\geq \Psi^{-1}(\delta/2)\Big) = 1-\delta/2.
    \end{split}
\end{equation*}
This finishes the proof of Corollary \ref{cor:CI_consistency}.

It is well known that the convergence in distribution implies the convergence in moment under the
uniform integrability condition.
The proof of the consistency of bootstrap moment estimation is straightforward since the condition  $\limsup_{N\to\infty} \mathbb E_{W^*|\cD}[(\sqrt{N}(\hat{v}_{\pi}^*-\hat{v}_{\pi}))^q]<\infty$ for some $q>2$ ensures a similar uniform integrability condition. Together with the distributional consistency in Theorem \ref{thm:bootstrap_consistency}, we apply Lemma 2.1 in \citet{kato2011note} then we reach the conclusion.
\hfill $\blacksquare$\\

\section{Supporting Results}

We present a series of useful lemmas about Mallows metric.
\begin{lemma}[Lemma 8.4 in \citet{bickel1981some}]\label{lemma:BF_8_4}
    Let $\{X_i\}_{i=1}^n$ be independent random variables with common distribution $\mu$. Let $\mu_n$ be the empirical distribution of $X_1, \ldots, X_n$. Then $\Lambda_l(\mu_n, \mu)\to 0$ a.e..
\end{lemma}

\begin{lemma}[Lemma 8.5 in \citet{bickel1981some}]\label{lemma:BF_8_5}
    Suppose $X_n, X$ are random variables and $\Lambda_l(X_n, X)\to 0$. Let $f$ be a continuous function. Then $\Lambda_l(f(X_n), f(X))\to 0$.
\end{lemma}

\begin{lemma}[Lemma 8.6 of \citet{bickel1981some}]\label{lemma:BF8_6}
    Let $\{U_i\}_{i=1}^n, \{V_i\}_{i=1}^n$ be independent random vectors. Then we have \begin{equation*}
        \Lambda_1\Big(\sum_{i=1}^nU_i, \sum_{i=1}^nV_i\Big)\leq \sum_{i=1}^n \Lambda_1\Big(U_i, V_i\Big).
    \end{equation*}
\end{lemma}

\begin{lemma}[Lemma 8.7 of \citet{bickel1981some}]\label{lemma:BF8_7}
     Let $\{U_i\}_{i=1}^n, \{V_i\}_{i=1}^n$ be independent random vectors and $\mathbb E[U_j] = \mathbb E[V_j]$. Then we have 
     \begin{equation*}
        \Lambda_2\Big(\sum_{i=1}^nU_i, \sum_{i=1}^nV_i\Big)^2\leq \sum_{i=1}^n \Lambda_2\Big(U_i, V_i\Big)^2.
    \end{equation*}
\end{lemma}

Let $\mu_{K}$ and $\mu$ be probabilities on $\mathbb R^{2Hd}$. A data point in $\mathbb R^{2Hd}$ can be written as $(x_1,\ldots, x_H, y_1,\ldots, y_H)$ where $x_h\in\mathbb R^d$ and $y_{h}\in\mathbb R^d$.
Denote
\begin{equation*}
\begin{split}
    &\Sigma(\mu) = \int \frac{1}{H}\sum_{h=1}^Hx_hx_h^{\top}\mu(dx_1,\ldots, dx_H,dy_1,\ldots, dy_H),\\
    &M(\mu) = \Sigma(\mu)^{-1}\int \sum_{h=1}^Hx_hy_h^{\top}\mu(dx_1,\ldots, dx_H,dy_1,\ldots, dy_H),\\
    &\varepsilon(\mu,x_1,\ldots, x_H,y_1,\ldots, y_H) = \sum_{h=1}^H(y_h-x_h^{\top}M(\mu)).
\end{split}
\end{equation*}
\begin{lemma}[Lemma 7 in \citet{eck2018bootstrapping}]\label{lemma:converge}
If $\Lambda_4(\mu_K,\mu)\to 0$ as $K\to\infty$, then we have the $\mu_K$-law of $\text{vec}(\sum_{h=1}^H\varepsilon(\mu_K,x_1,\ldots, x_H,y_1,\ldots, y_H)x_h^{\top} )$ converges to the $\mu$-law of $\text{vec}(\sum_{h=1}^H\varepsilon(\mu,x_1,\ldots, x_H,y_1,\ldots, y_H)x_h^{\top} )$ in $\Lambda_2$.
\end{lemma}

\begin{theorem}[Multivariate delta theorem]\label{thm:delta}
Suppose $\{T_n\}$ is a sequence of $k$-dimensional random vectors such that $\sqrt{n}(T_n-\theta)\overset{d}{\to} N(0, \Sigma(\theta))$. Let $g:\mathbb R^k\to\mathbb R$ be once differentiable at $\theta$ with the gradient matrix $\nabla g(\theta)$. Then 
\begin{equation*}
    \sqrt{n}(g(T_n)-g(\theta))\overset{d}{\to} N(0, \nabla^{\top}g(\theta)\Sigma(\theta)\nabla g(\theta)).
\end{equation*}
\end{theorem}

We restate  Lemma B.5 in \citet{duan2020minimax} in the following that is proven using matrix Bernstein
inequality.
\begin{lemma} \label{lemma:Term1}
	Under the assumption $\phi(s,a)^{\top} \Sigma^{-1} \phi(s,a) \leq C_1 d$ for all $(s,a) \in \mathcal{X}$, with probability at least $1 - \delta$,
	\begin{equation} \label{Term1} \Bigg\| \Sigma^{-1/2} \bigg( \frac{1}{N} \sum_{n=1}^N \phi(s_n,a_n) \phi(s_n,a_n)^{\top} \bigg) \Sigma^{-1/2} - I \Bigg\|_2 \leq \sqrt{\frac{2\ln(2d/\delta)C_1dH}{N}} + \frac{2\ln(2d/\delta)C_1 d H}{3N}. \end{equation}
\end{lemma}

\section{Supplement for Experiments}\label{sec:experi_appendix}

\subsection{Experiment details}
The original CliffWalking environment from OpenAI gym has deterministic state transitions.  That is, for any state-action pair $(s,a)$, there exists a corresponding $s'\in \cS$ such that $\mP(\cdot \mid s, a) = \delta_{s'}(\cdot)$. We modify the environment in order to make it stochastic. Specifically, we introduce randomness in state transitions such that given a state-action pair $(s,a)$, the transition takes place in the same way as in the deterministic environment with probability $1 - \epsilon$ and takes place as if the action were a random action $a'$, instead of the intended $a$, with probability $\epsilon$. This is an episodic tabular MDP and the agent stops when falling from the cliff or reaching the terminal point. We also reduce the penalty of falling off the cliff from $-100$ to $-50$. 

The original MountainCar environment from OpenAI gym has deterministic state transitions. We modify the environment in order to make it stochastic. Specifically, we introduce randomness in state transitions by adding a Gaussian random force, namely, $\mathcal{N}(0,\frac{1}{10})$ multiplied by the constant-magnitude force from the original environment. We also increase the gravity parameter from $0.0025$ to $0.008$, the force parameter from $0.001$ to $0.008$ and the maximum allowed speed from $0.07$ to $0.2$.

\paragraph{Empirical coverage probability.} The preceding discussion leads to the simulation method for estimating the coverage probability of a confidence interval. The simulation method has three steps: 
\begin{enumerate}
    \item Simulate many fresh dataset of episode size $K$ following the behavior policy.
    \item Compute the confidence interval for each dataset.
    \item Compute the proportion of dataset for which the true value of target policy is contained in the confidence interval. That proportion is an estimate for the empirical coverage probability for the confidence interval.
\end{enumerate}
The true value of target policy is computed through Monte Carlo rollouts with sufficient number of samples (10000 in our experiments).

With linear function approximation, we use the confidence interval proposed in Section 6 in \citet{duan2020minimax} as a baseline since it is only available confidence interval based on FQE. In particular, it shows that with probability at least $1-\delta$, 
\begin{equation*}
\begin{split}
     |\hat{v}^{\pi}-v^{\pi}|\leq & \sum_{h=1}^H(H-h+1)\sqrt{(\hat{\nu}^{\pi}_h)^{\top}\hat{\Sigma}^{-1}\hat{\nu}_h^{\pi}}\Big(\sqrt{2\lambda}+2\sqrt{2d\log\Big(1+\frac{N}{\lambda d}\Big)\log\Big(\frac{3N^2H}{\delta}\Big)}+\frac{4}{3}\log\Big(\frac{3N^2H}{\delta}\Big)\Big),
\end{split}
\end{equation*}
where $(\hat{\nu}_h^{\pi})^{\top}=(\nu_1^{\pi})^{\top}(\hat{M}^{\pi})^h$ and $\hat{M}_{\pi}$ is defined in Eq.~\eqref{eqn:mean_embedding}.

\subsection{Additional experiments}

\begin{figure}[!t]
 \centering
   \includegraphics[width=0.3\linewidth]{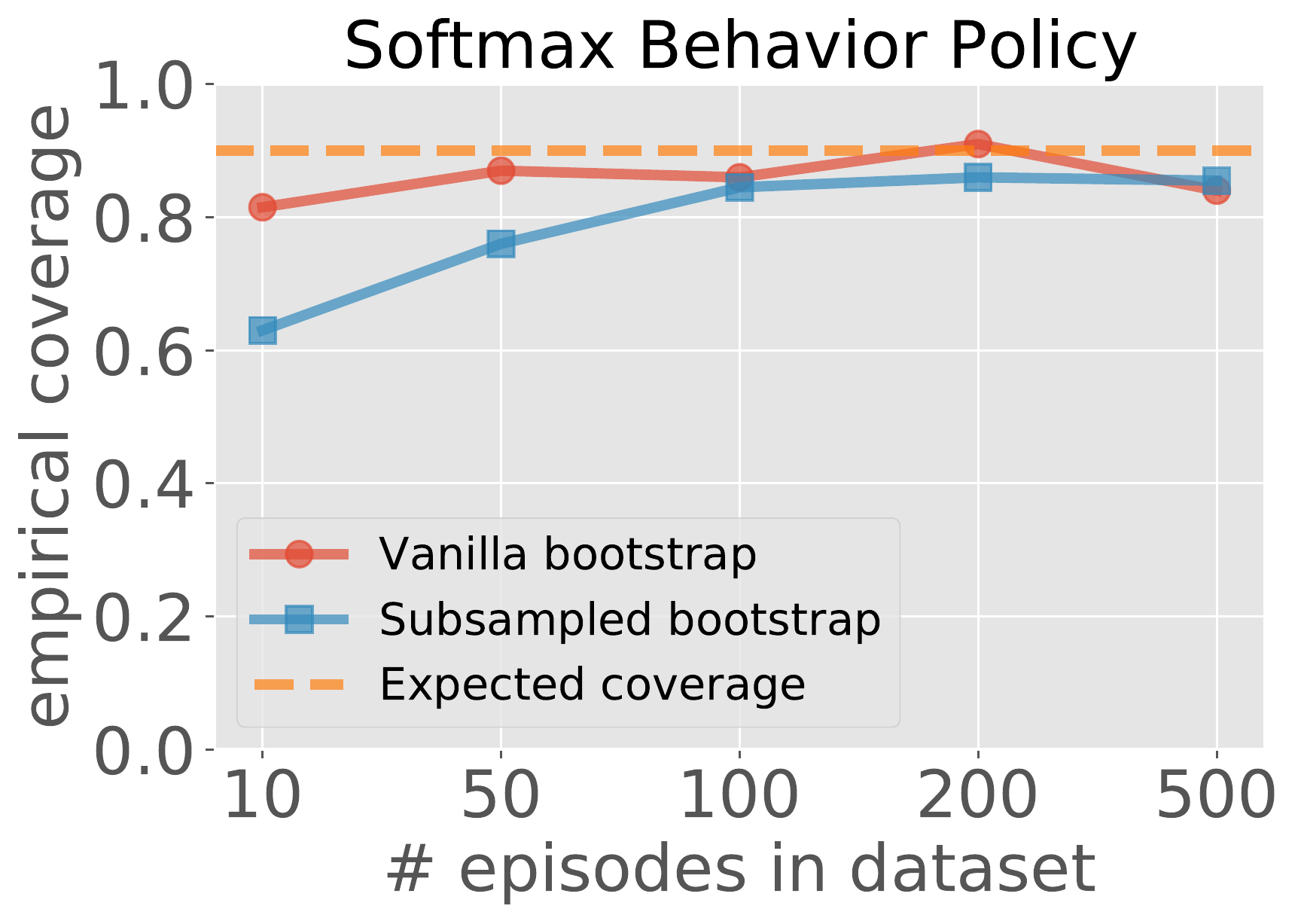}
  \includegraphics[width=0.3\linewidth]{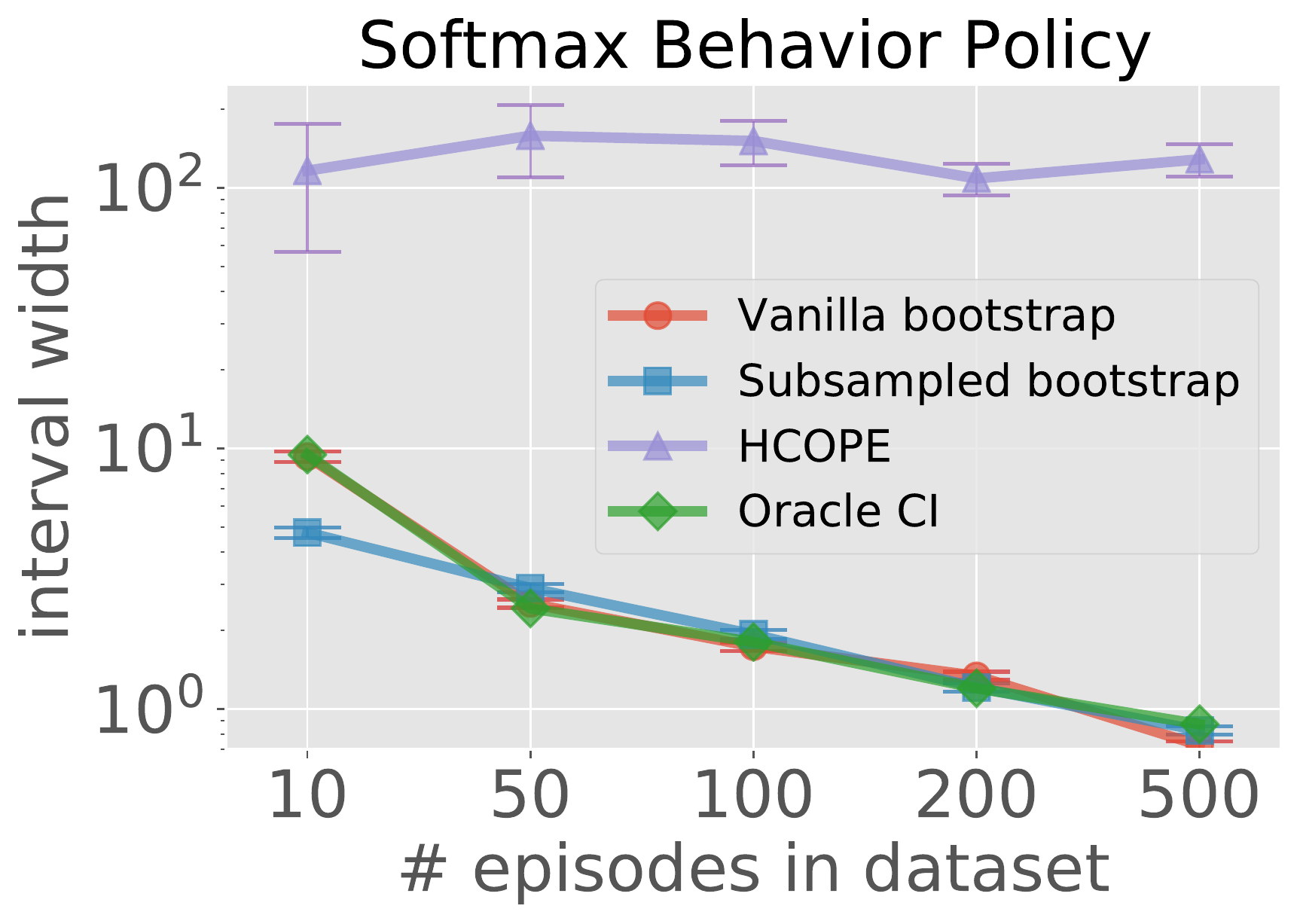}
\caption{Left: Empirical coverage probability of CI; Right: CI width under different behavior policies.}
\label{fig:tabular_CI_soft_max}
\end{figure}

\begin{figure}[!t]
 \centering
  \includegraphics[width=0.3\linewidth]{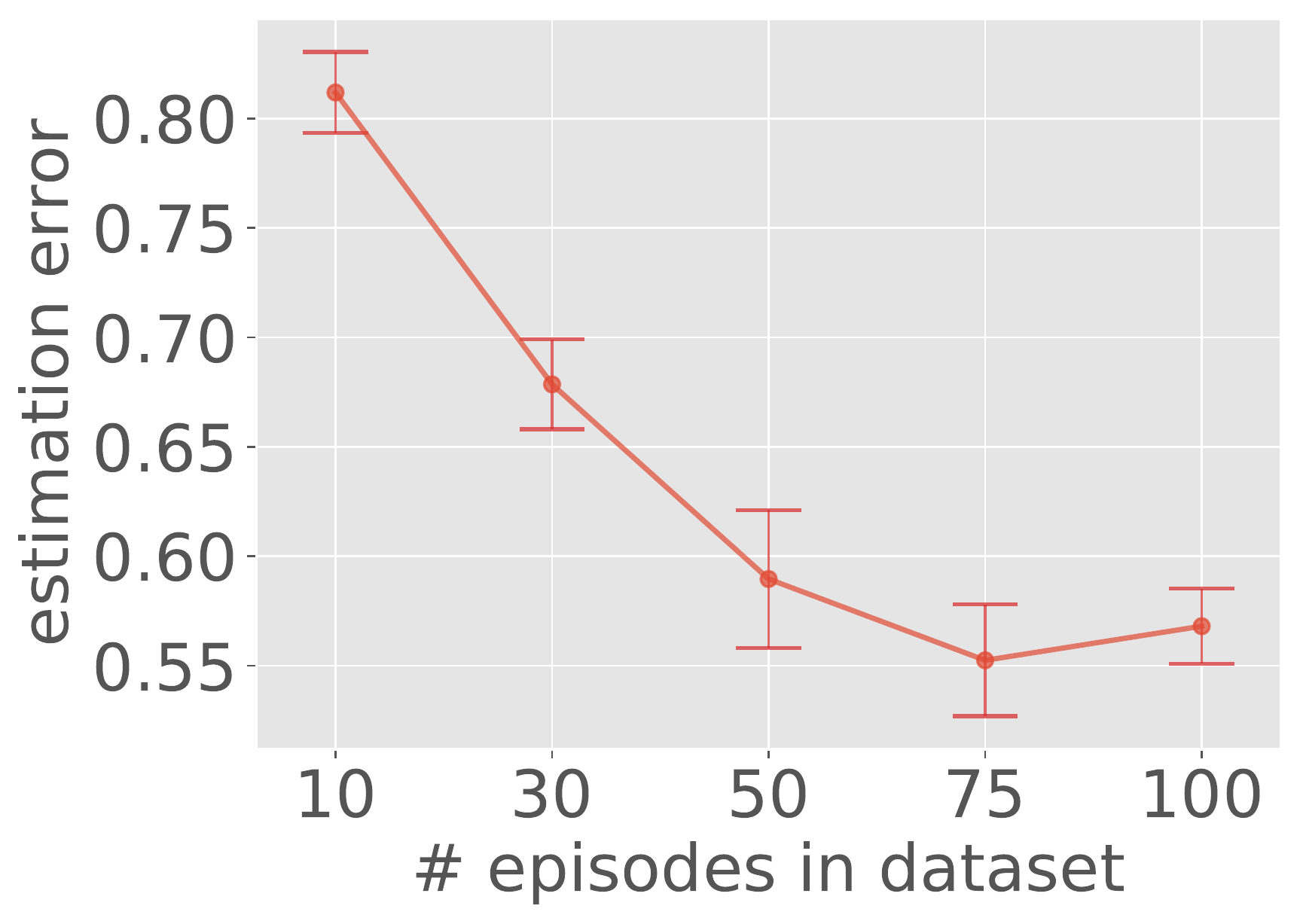}
\caption{Error of correlation estimates, as data size increases.}

\label{fig:correlation_est}
\end{figure}

In Figure \ref{fig:tabular_CI_soft_max}, we include the result for soft-max behavior policy in the Cliff Walking environment.
In Figure \ref{fig:correlation_est}, we include the result for correlation estimation in Cliff Walking environment. The behavior policy is 0.1 $\epsilon$-greedy policy while two target policies are optimal policy and 0.1 $\epsilon$-greedy policy.

In order to better understand the tradeoff between computational efficiency and accuracy with finite samples, we conduct some empirical demonstrations based on Cliffwalking. We set $s = K^{\gamma}$ and the true coverage probability is 0.9. It is relatively safe to set $\gamma > 0.5$. Note that $\gamma =1$ corresponds to the vanilla bootstrap that has the highest accuracy but heaviest computation.
 \begin{figure}[h!]\label{fig:subsample}
\includegraphics[width=0.3\linewidth]{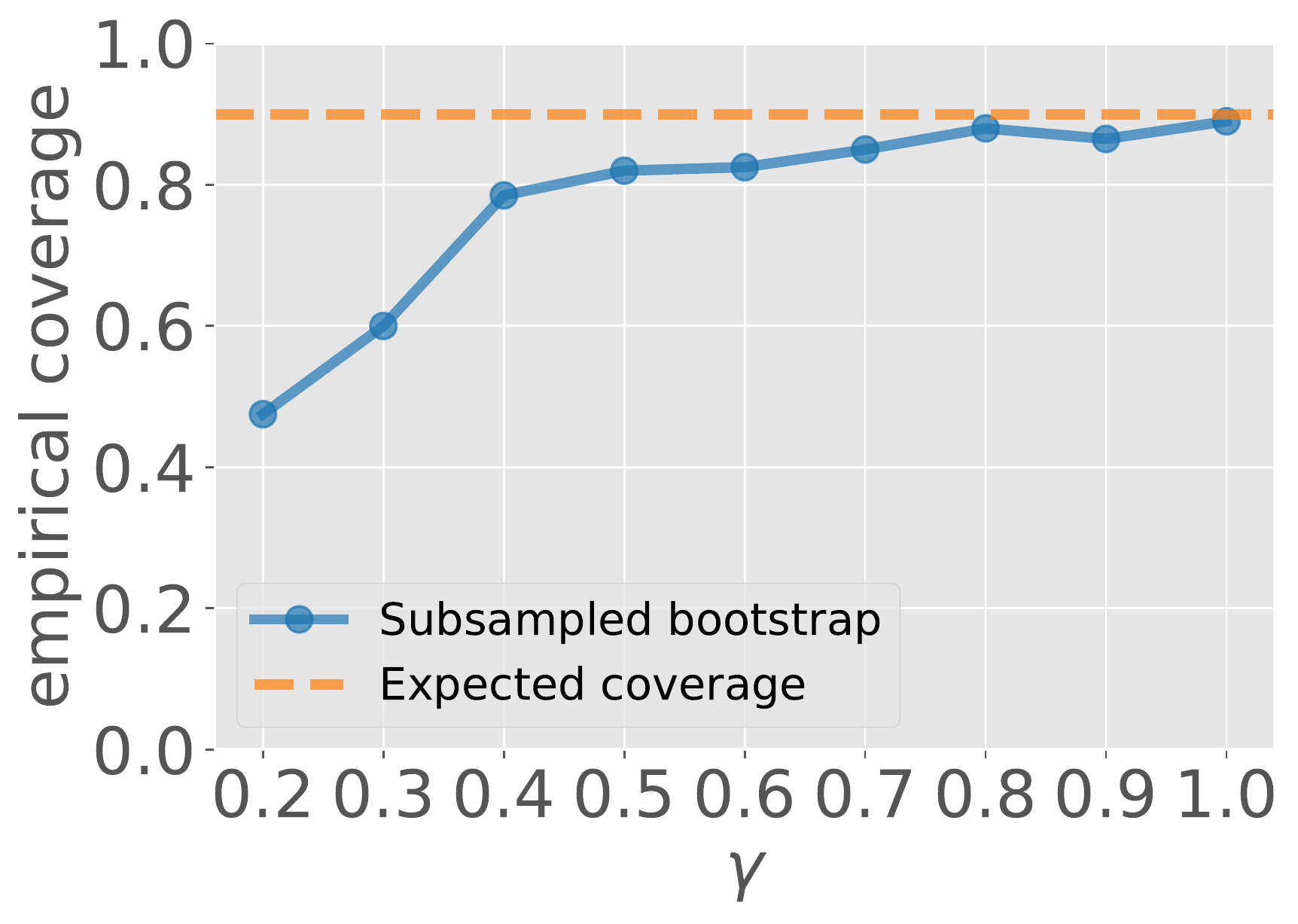}
\includegraphics[width=0.3\linewidth]{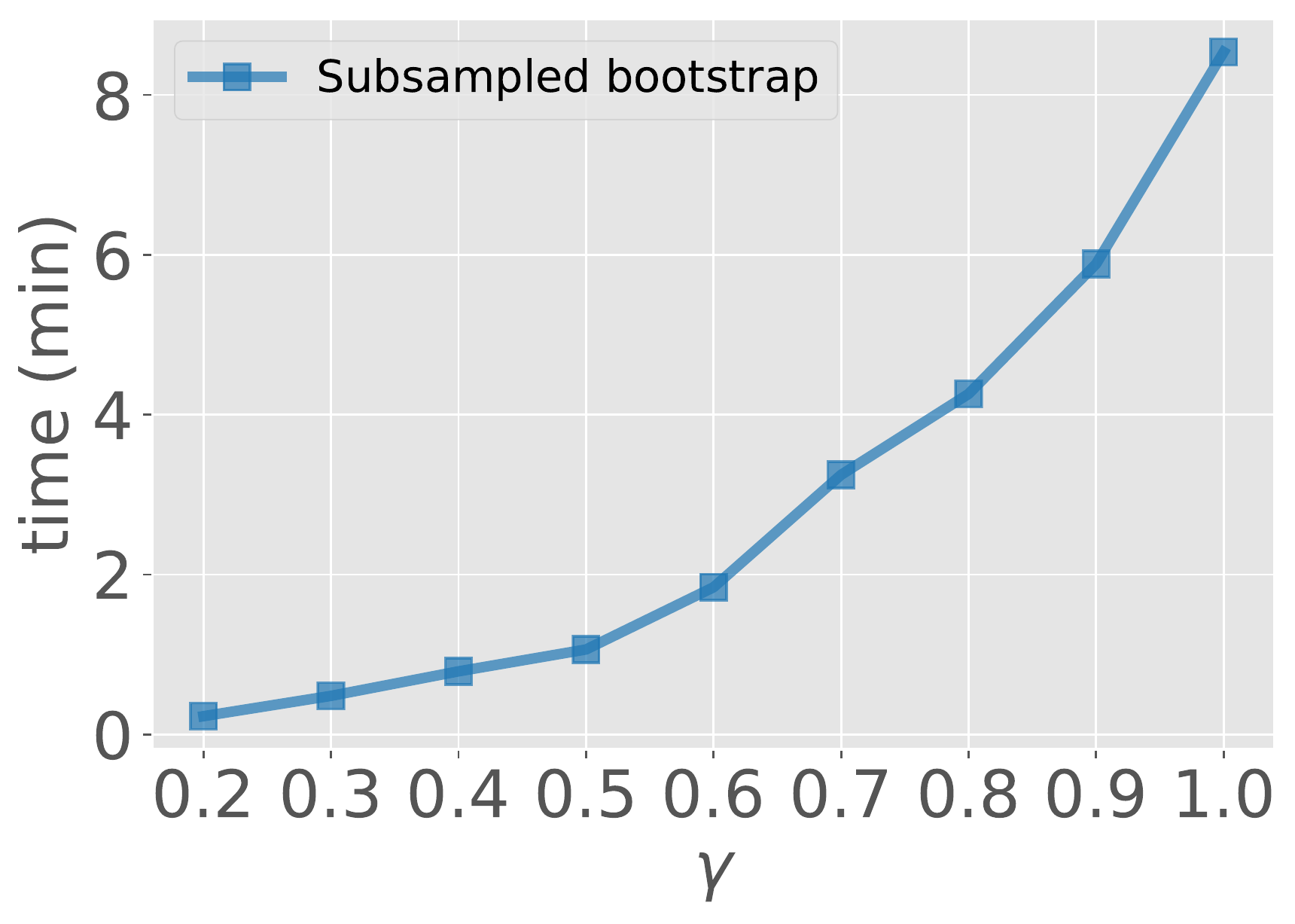}
\caption{Subsampled bootstrap with $s=K^{\gamma}$}

\end{figure}

We argue that bootstrapping sample transitions (which are dependent) would lead to inconsistent estimations of the error distribution and thus output wrong confidence interval and variance estimation. We further run one additional test using the taxi environment and further compute the CI and variance estimations based on different bootstrap distributions in Cliffwalking (CW) and taxi environments. This is already in an asymptotic regime since both the number of episodes and the number of bootstrap samples are $~$ 10e+6. It is clear that bootstrapping by sample transition gives an incorrect distribution, thus it is inconsistent.
\begin{table}[h!]
\scalebox{0.9}{
\begin{tabular}{ |l|c|c|c| } 
 \hline
& True distribution& By episodes & By sample transition\\ 
 \hline
 CI (CW) & (-20.44, -19.74) & (-20.40, -19.74) &  \red{(-20.52, -19.58)} \\ 
 \hline
 Variance (CW) & 0.45 & 0.44&  \red{0.082} \\ 
 \hline
 CI (Taxi) & (2.49, 3.82) & (2.45, 3.73) & \red{(2.01, 4.09)} \\ 
 \hline
 Variance (Taxi) & 0.16 & 0.16 & \red{0.39} \\ 
 \hline
\end{tabular}}
\end{table}
 \begin{figure}
 \includegraphics[width=0.25\linewidth]{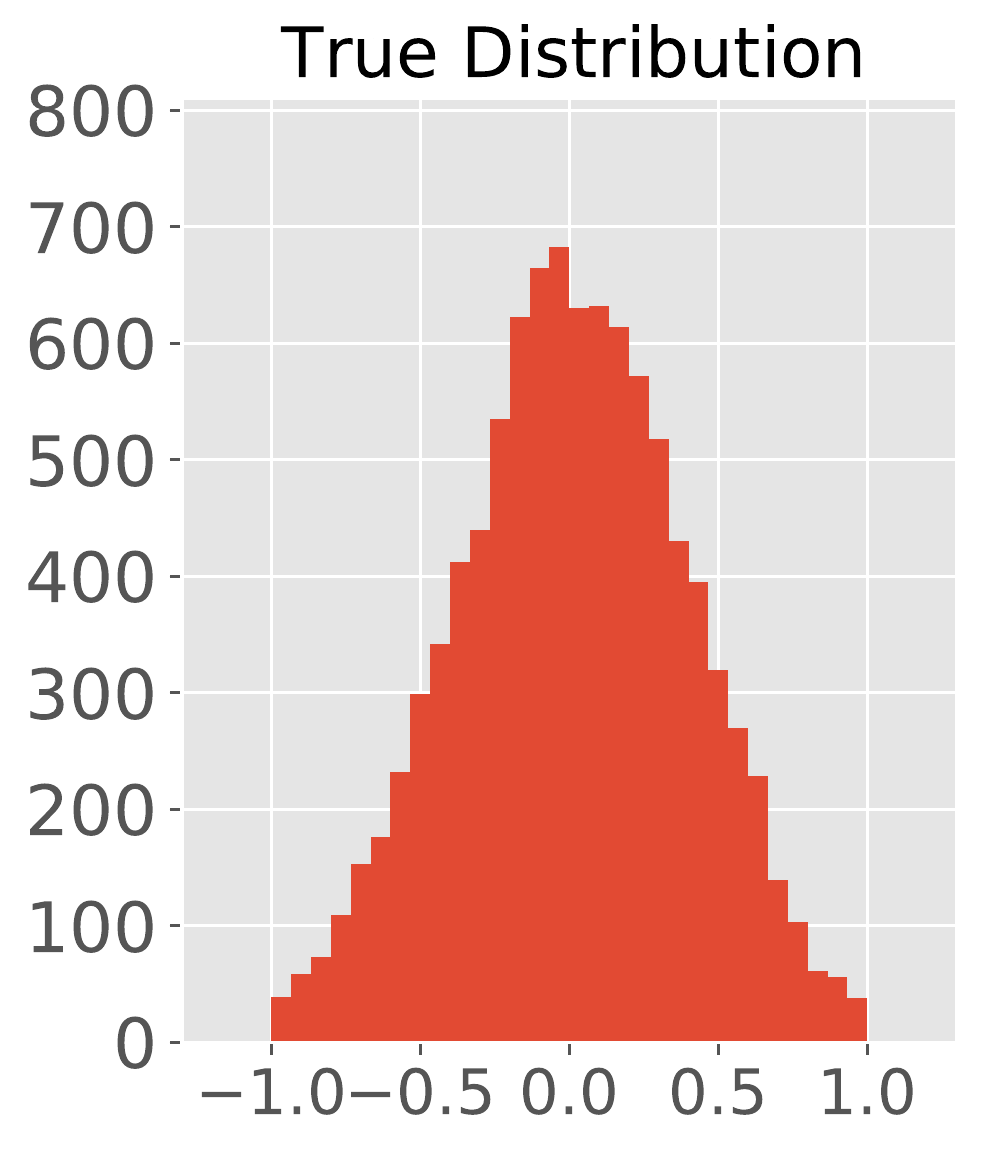}
\includegraphics[width=0.25\linewidth]{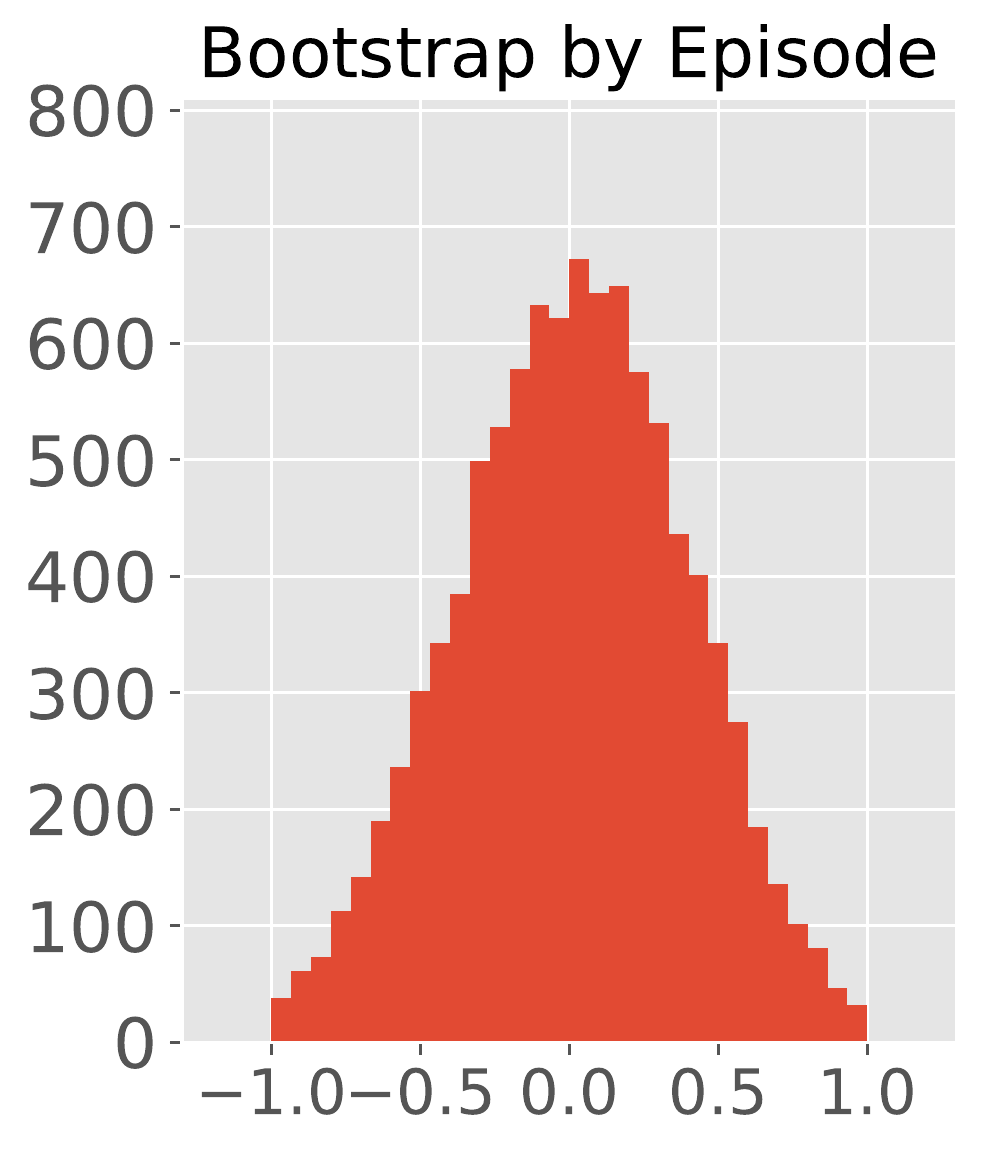}
\includegraphics[width=0.25\linewidth]{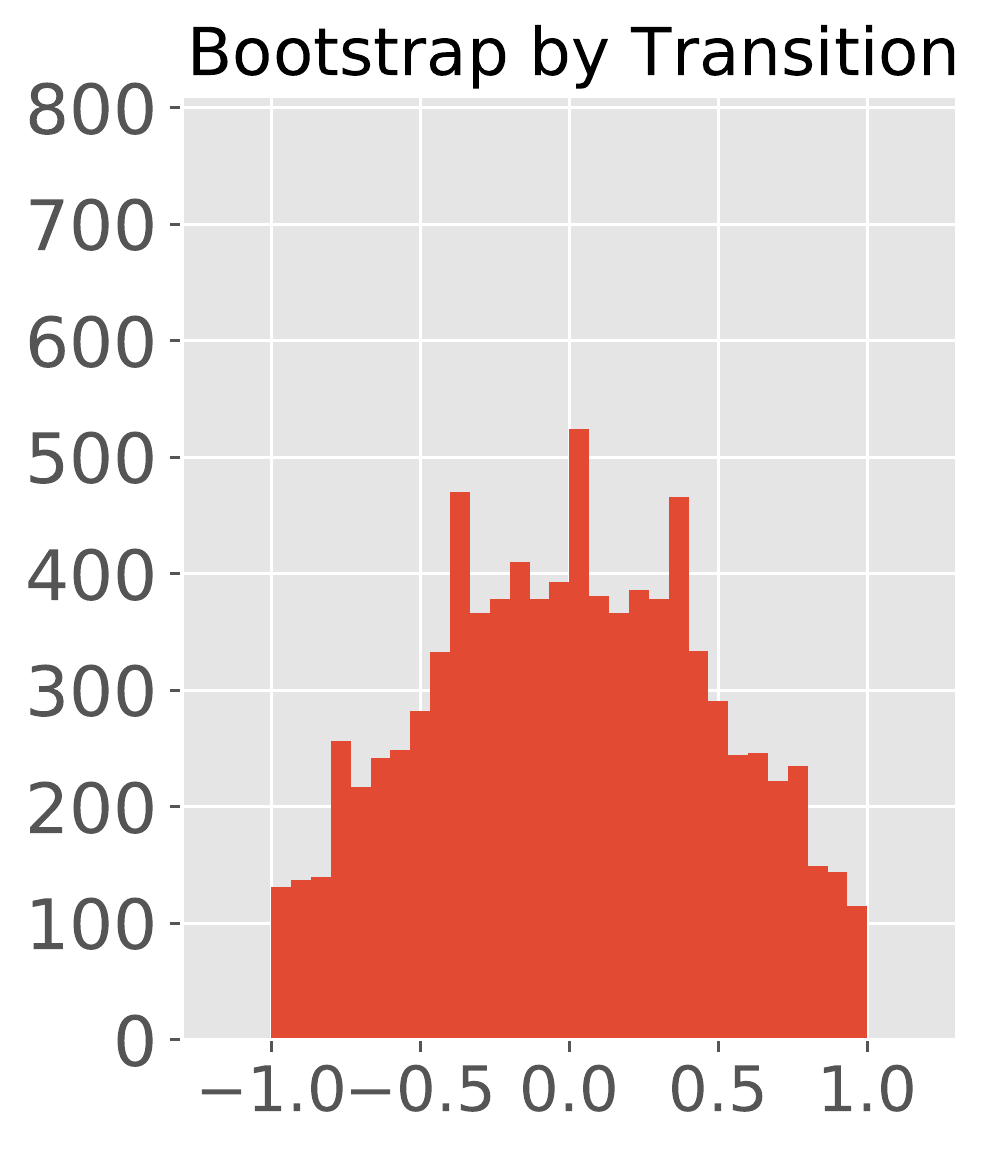}
\caption{Taxi environment (Dietterich 2000)}
\end{figure}

%% file: lb.tex
\paragraph{Influence function.}
	Recall that our dataset $\mathcal{D}$ consists of $K$ {\it i.i.d.} trajectories, each of which has length $H$. Denote
	\[ \boldsymbol{\tau} :\,= \big(s_1, a_1, r_1, s_2, a_2, r_2, \ldots, s_H, a_H, r_H, s_{H+1} \big). \]
	For simplicity, we assume that the reward $r_h$ is deterministic given $(s_h,a_h)$, {\it i.e.} $r_h = r(s_h,a_h)$ for some reward function $r$.
	The distribution of $\boldsymbol{\tau}$ is given by
	\begin{align*} \mathcal{P}({\rm d}\boldsymbol{\tau}) = & \bar{\xi}_1 ({\rm d}s_1, {\rm d}a_1) \mathcal{P}({\rm d}s_2 \, \big| \, s_1, a_1) \bar{\pi}({\rm d}a_2 \mid s_2) \cP({\rm d}s_3 \mid s_2, a_2) \\ & \ldots \mathcal{P}({\rm d}s_H \mid s_{H-1}, a_{H-1}) \bar{\pi}({\rm d}a_H \mid s_H) \mathcal{P}({\rm d}s_{H+1} \mid s_H, a_H). \end{align*}
	
	Define
	$\mathcal{P}_{\eta} :\,= \mathcal{P} + \eta \Delta \mathcal{P}$ where $\Delta \mathcal{P}$ satisfies 
	\[ (\Delta \mathcal{P}) \mathcal{F} \subseteq \mathcal{F} \]
	under condition~\ref{assum:completeness}.
	Denote score functions \[ g(\boldsymbol{\tau}) :\,= \frac{\partial}{\partial \eta} \log \mathcal{P}_{\eta}({\rm d} \boldsymbol{\tau}) \qquad \text{and} \qquad g(s' \mid s,a) :\,= \frac{\partial}{\partial \eta} \log \mathcal{P}_{\eta}({\rm d}s' \mid s,a). \]
	Note that
	\[ g(\boldsymbol{\tau}) = \sum_{h=1}^H g(s_{h+1} \mid s_h, a_h). \]
	
	We consider the pointwise estimation. The objective function
	$\psi_{\xi_1}$ is defined as \[ \psi_{\xi_1}(\mathcal{P}_{\eta}) :\,= \mathbb{E}\Bigg[ \sum_{h=1}^H r_{\eta}(s_h,a_h) \Biggm| (s_1,a_1) \sim \xi_1, \mathcal{P}_{\eta}, \pi \Bigg]. \]
	
	We calculate the derivative $\frac{\partial}{\partial \eta} \psi_{\xi_1}(\mathcal{P}_{\eta})$ and have
	\begin{align*} \frac{\partial}{\partial \eta} \psi_{\xi_1}(\mathcal{P}_{\eta}) = & \frac{\partial}{\partial \eta} \Bigg[ \sum_{h=1}^H \int_{(\mathcal{S}\times\mathcal{A})^h} r(s_h,a_h) \xi_1({\rm d}s_1, {\rm d}a_1) \prod_{j=1}^{h-1} \mathcal{P}_{\eta}({\rm d}s_{j+1} \mid s_j, a_j) \pi({\rm d}a_{j+1} \mid s_{j+1}) \Bigg] \\ = & \sum_{h=1}^H \int_{(\mathcal{S} \times \mathcal{A})^h} r(s_h,a_h) \bigg( \sum_{j=1}^{h-1} g(s_{j+1} \, | \, s_j,a_j) \bigg) \xi_1({\rm d}s_1, {\rm d}a_1) \prod_{j=1}^{h-1} \mathcal{P}_{\eta}({\rm d}s_{j+1} \, | \, s_j,a_j) \pi({\rm d}a_{j+1} \mid s_{j+1}). \end{align*}
	By using Q-functions $Q_{\eta,j}^{\pi}(s_j,a_j) := \mathbb{E} \big[ \sum_{h=j}^H r_{\eta}(s_h,a_h) \bigm| (s_j, a_j), \mathcal{P}_{\eta}, \pi \big]$ for $j = 1,2,\ldots,H$, $Q_{\eta,H+1} := 0$, we find that
	\begin{align*}
	\frac{\partial}{\partial \eta} \psi_{\xi_1}(\mathcal{P}_{\eta}) = & \int_{(\mathcal{S} \times \mathcal{A})^H} \sum_{j=1}^{H-1} g(s_{j+1} \, | \, s_j,a_j) \sum_{h=j+1}^H r_{\eta}(s_h,a_h) \xi_1({\rm d}s_1, {\rm d}a_1) \prod_{i=1}^{H-1} \mathcal{P}_{\eta}({\rm d}s_{i+1} \, | \, s_i,a_i) \pi({\rm d}a_{i+1} \mid s_{i+1}) \\
	= & \int_{(\mathcal{S} \times \mathcal{A})^H} \sum_{j=1}^{H-1} g(s_{j+1} \, | \, s_j,a_j) \xi_1({\rm d}s_1, {\rm d}a_1) \prod_{i=1}^j \mathcal{P}_{\eta}({\rm d}s_{i+1} \, | \, s_i,a_i) \pi({\rm d}a_{i+1} \mid s_{i+1}) \\ & \qquad \qquad \cdot \Bigg( \sum_{h=j+1}^H r_{\eta}(s_h,a_h) \prod_{i=j+1}^{H-1} \mathcal{P}_{\eta}({\rm d}s_{i+1} \, | \, s_i,a_i) \pi({\rm d}a_{i+1} \mid s_{i+1}) \Bigg) \\
	= & \sum_{j=1}^{H-1} \int_{(\mathcal{S} \times \mathcal{A})^{j+1}} g(s_{j+1} \, | \, s_j,a_j) Q_{\eta,j+1}^{\pi}(s_{j+1},a_{j+1}) \xi_1({\rm d}s_1, {\rm d}a_1) \prod_{i=1}^j \mathcal{P}_{\eta}({\rm d}s_{i+1} \, | \, s_i,a_i) \pi({\rm d}a_{i+1} \mid s_{i+1}) \\
	= & \sum_{h=1}^{H} \mathbb{E}\big[ g(s_{h+1} \, | \, s_h,a_h) V_{\eta,h+1}^{\pi}(s_{h+1}) \, \big| \, (s_1, a_1) \sim \xi_1, \mathcal{P}_{\eta}, \pi \big]. \end{align*}
	It follows that
	\begin{align*}
		\frac{\partial}{\partial \eta} \psi_{\xi_1}(\mathcal{P}_{\eta}) \bigg|_{\eta = 0} = & \mathbb{E}\Bigg[ \sum_{h=1}^H g(s_{h+1} \mid s_h,a_h) V_{h+1}^\pi(s_{h+1}) \Biggm| (s_1,a_1) \sim \xi_1, \mathcal{P}, \pi \Bigg].
	\end{align*}
	
	Define $w_h(s,a) :\,= \phi(s,a)^{\top} \Sigma^{-1} \nu_h^{\pi} = \phi(s,a)^{\top} \Sigma^{-1} \mathbb{E} \big[ \phi(s_h,a_h) \bigm| (s_1, a_1) \sim \xi_1, \mathcal{P}, \pi \big]$ for $h=1,2,\ldots,H$. For any $f \in \mathds{H}$ with $f(s,a) = \phi(s,a)^{\top} w_f$, we have
	\begin{align*} \mathbb{E}\big[ f(s_h,a_h) \bigm| (s_1, a_1) \sim \xi_1, \mathcal{P}, \pi \big] = & \mathbb{E}\big[ \phi(s_h,a_h)^{\top} w_f \bigm| (s_1, a_1) \sim \xi_1, \mathcal{P}, \pi \big] \\ = & \mathbb{E}\big[ \phi(s_h,a_h)^{\top} \Sigma^{-1} \mathbb{E}_{(s,a) \sim \bar{\mu}} [\phi(s,a) \phi(s,a)^{\top}] w_f \bigm| (s_1, a_1) \sim \xi_1, \mathcal{P}, \pi \big] \\ = & \mathbb{E}_{(s,a) \sim \bar{\mu}} \Big[ \mathbb{E}\big[ \phi(s_h,a_h) \bigm| (s_1, a_1) \sim \xi_1, \mathcal{P}, \pi \big]^{\top} \Sigma^{-1} \phi(s,a) \phi(s,a)^{\top} w_f \Big] \\ = & \mathbb{E}_{(s,a) \sim \bar{\mu}} \big[ w_h(s,a) f(s,a) \big], \end{align*}
	where $\bar{\mu}$ is the distribution of dataset $\mathcal{D}$.
	Since the mapping $(s,a) \mapsto \mathbb{E}\big[ g(s' \, | \, s,a) V_h^{\pi}(s') \bigm| s,a \big]$ belongs to $\mathds{H}$, therefore, 
	\begin{align*}
		\frac{\partial}{\partial \eta} \psi_{\xi_1}(\mathcal{P}_{\eta}) \bigg|_{\eta = 0} = & \mathbb{E}_{(s,a) \sim \bar{\mu}} \Bigg[ \sum_{h=1}^H w_h(s,a) g(s' \mid s,a) V_{h+1}^\pi(s') \Bigg].
	\end{align*}
		Note that $\mathbb{E}\big[ g(s' \mid s,a) \bigm| s,a \big] = 0$, therefore,
	\begin{align*} 
		\frac{\partial}{\partial \eta} \psi_{\xi_1}(\mathcal{P}_{\eta}) \bigg|_{\eta = 0} = \mathbb{E}_{(s,a) \sim \bar{\mu}} \Bigg[ \sum_{h=1}^H w_h(s,a) g(s' \mid s,a) \Big( V_{h+1}^\pi(s') - \mathbb{E}\big[V_{h+1}^\pi(s') \bigm| s, a \big] \Big) \Bigg]. 
	\end{align*}
	By definition of $\mu$, we have
	\begin{align*}
		& \frac{\partial}{\partial \eta} \psi_{\xi_1}(\mathcal{P}_{\eta}) \bigg|_{\eta = 0} \\ = & \frac{1}{H} \sum_{j=1}^H \mathbb{E} \Bigg[ \sum_{h=1}^H w_h(s_j,a_j) g(s_{j+1} \mid s_j,a_j) \Big(V_{h+1}^\pi(s_{j+1}) - \mathbb{E}\big[V_{h+1}^\pi(s_{j+1}) \bigm| s_j, a_j \big] \Big) \Biggm| (s_1, a_1) \sim \bar{\xi}_1, \mathcal{P}, \bar{\pi} \Bigg].
	\end{align*}
	We use the property $\mathbb{E}\big[ g(s' \mid s,a) \bigm| s,a \big] = 0$ again and derive that
	\begin{align*}
		& \frac{\partial}{\partial \eta} \psi_{\xi_1}(\mathcal{P}_{\eta}) \bigg|_{\eta = 0} \\ = & \frac{1}{H} \sum_{j=1}^H \mathbb{E} \Bigg[ \sum_{h=1}^H w_h(s_j,a_j) \bigg( \sum_{l=1}^H g(s_{l+1} \mid s_l,a_l) \bigg) \Big(V_{h+1}^\pi(s_{j+1}) - \mathbb{E}\big[V_{h+1}^\pi(s_{j+1}) \bigm| s_j, a_j \big] \Big) \Biggm| (s_1, a_1) \sim \bar{\xi}_1, \mathcal{P}, \bar{\pi} \Bigg] \\ = & \frac{1}{H} \mathbb{E} \Bigg[ g(\boldsymbol{\tau}) \sum_{h=1}^H \sum_{j=1}^H  w_h(s_j,a_j) \Big(V_{h+1}^\pi(s_{j+1}) - \mathbb{E}\big[V_{h+1}^\pi(s_{j+1}) \bigm| s_j, a_j \big] \Big) \Biggm| (s_1, a_1) \sim \bar{\xi}_1, \mathcal{P}, \bar{\pi} \Bigg].
	\end{align*}
	We can conclude that
	\[ \dot{\psi}_{\mathcal{P}}(\boldsymbol{\tau}) :\,= \frac{1}{H} \sum_{h=1}^H \sum_{h'=1}^H w_{h'}(s_h,a_h) \Big(V_{h'+1}^\pi(s_{h+1}) - \mathbb{E}\big[V_{h'+1}^\pi(s_{h+1}) \bigm| s_{h}, a_{h} \big] \Big), \]
	is an influence function.
	
	\paragraph{Efficiency bound.}
		For notational convenience, we take shorthands
		\[ q(s,a,s') :\,= \sum_{h=1}^H w_h(s,a)\Big( V_{h+1}^{\pi}(s') - \mathbb{E}\big[ V_{h+1}^{\pi}(s') \bigm| s,a \big] \Big), \]
		and rewrite
		\[ \dot{\psi}_{\mathcal{P}}(\boldsymbol{\tau}) = \frac{1}{H} \sum_{h=1}^H q(s_h, a_h, s_{h+1}). \]
		Since $\mathbb{E}\big[ q(s,a,s') \bigm| s,a \big] = 0$, we find that
		\begin{align*}
			\mathbb{E}\big[ \dot{\psi}_{\mathcal{P}}^2(\boldsymbol{\tau}) \big] = &  \frac{1}{H^2} \mathbb{E}\Bigg[ \bigg( \sum_{h=1}^H q(s_h,a_h,s_{h+1}) \bigg)^2 \Biggm| \bar{\xi}_1, \mathcal{P}, \bar{\pi} \Bigg] = \frac{1}{H^2} \sum_{h=1}^H \mathbb{E}\Big[ q^2 (s_h,a_h,s_{h+1}) \Bigm| \bar{\xi}_1, \mathcal{P}, \bar{\pi} \Big].
		\end{align*}
		It follows that
		\begin{align*} \mathbb{E}\big[ \dot{\psi}_{\mathcal{P}}^2(\boldsymbol{\tau}) \big] = & \frac{1}{H} \mathbb{E}_{(s,a) \sim \bar{\mu}}\Big[ \mathbb{E}\big[ q^2 (s,a,s') \bigm| s,a \big] \Big] = \frac{1}{H} \mathbb{E}_{(s,a) \sim \bar{\mu}}\Big[ \mathbb{E}\big[ q^2 (s,a,s') \bigm| s,a \big] \Big] \\ = & \frac{1}{H} \mathbb{E}_{(s,a) \sim \bar{\mu}} \Bigg[ \bigg( \phi(s,a)^{\top} \Sigma^{-1} \sum_{h=1}^H \Big( V_{h+1}^{\pi}(s') - \mathbb{E}\big[ V_{h+1}^{\pi}(s') \bigm| s,a \big] \Big) \nu_h^{\pi} \bigg)^2 \Bigg], \end{align*}
	which coincides with the asymptotic variance of OPE estimator defined in \eqref{def:asy_variance}.
	
	\hfill $\blacksquare$\\

%% file: icml_submission.bbl
\begin{thebibliography}{47}
\providecommand{\natexlab}[1]{#1}
\providecommand{\url}[1]{\texttt{#1}}
\expandafter\ifx\csname urlstyle\endcsname\relax
  \providecommand{\doi}[1]{doi: #1}\else
  \providecommand{\doi}{doi: \begingroup \urlstyle{rm}\Url}\fi

\bibitem[Bickel \& Freedman(1981)Bickel and Freedman]{bickel1981some}
Bickel, P.~J. and Freedman, D.~A.
\newblock Some asymptotic theory for the bootstrap.
\newblock \emph{The annals of statistics}, pp.\  1196--1217, 1981.

\bibitem[Dai et~al.(2020)Dai, Nachum, Chow, Li, Szepesv{\'a}ri, and
  Schuurmans]{dai2020coindice}
Dai, B., Nachum, O., Chow, Y., Li, L., Szepesv{\'a}ri, C., and Schuurmans, D.
\newblock Coindice: Off-policy confidence interval estimation.
\newblock \emph{arXiv preprint arXiv:2010.11652}, 2020.

\bibitem[Duan \& Wang(2020)Duan and Wang]{duan2020minimax}
Duan, Y. and Wang, M.
\newblock Minimax-optimal off-policy evaluation with linear function
  approximation.
\newblock \emph{Internation Conference on Machine Learning}, 2020.

\bibitem[Eck(2018)]{eck2018bootstrapping}
Eck, D.~J.
\newblock Bootstrapping for multivariate linear regression models.
\newblock \emph{Statistics \& Probability Letters}, 134:\penalty0 141--149,
  2018.

\bibitem[Efron(1982)]{efron1982jackknife}
Efron, B.
\newblock \emph{The jackknife, the bootstrap and other resampling plans}.
\newblock SIAM, 1982.

\bibitem[Ernst et~al.(2005)Ernst, Geurts, and Wehenkel]{ernst2005tree}
Ernst, D., Geurts, P., and Wehenkel, L.
\newblock Tree-based batch mode reinforcement learning.
\newblock \emph{Journal of Machine Learning Research}, 6\penalty0
  (Apr):\penalty0 503--556, 2005.

\bibitem[Feng et~al.(2020)Feng, Ren, Tang, and Liu]{feng2020accountable}
Feng, Y., Ren, T., Tang, Z., and Liu, Q.
\newblock Accountable off-policy evaluation with kernel bellman statistics.
\newblock \emph{Proceedings of the International Conference on Machine
  Learning}, 2020.

\bibitem[Feng et~al.(2021)Feng, Tang, Zhang, and Liu]{feng2021nonasymptotic}
Feng, Y., Tang, Z., Zhang, N., and Liu, Q.
\newblock Non-asymptotic confidence intervals of off-policy evaluation: Primal
  and dual bounds.
\newblock In \emph{International Conference on Learning Representations}, 2021.
\newblock URL \url{https://openreview.net/forum?id=dKg5D1Z1Lm}.

\bibitem[Fonteneau et~al.(2013)Fonteneau, Murphy, Wehenkel, and
  Ernst]{fonteneau2013batch}
Fonteneau, R., Murphy, S.~A., Wehenkel, L., and Ernst, D.
\newblock Batch mode reinforcement learning based on the synthesis of
  artificial trajectories.
\newblock \emph{Annals of operations research}, 208\penalty0 (1):\penalty0
  383--416, 2013.

\bibitem[Freedman et~al.(1981)]{freedman1981bootstrapping}
Freedman, D.~A. et~al.
\newblock Bootstrapping regression models.
\newblock \emph{The Annals of Statistics}, 9\penalty0 (6):\penalty0 1218--1228,
  1981.

\bibitem[Hallak \& Mannor(2017)Hallak and Mannor]{hallak2017consistent}
Hallak, A. and Mannor, S.
\newblock Consistent on-line off-policy evaluation.
\newblock In \emph{Proceedings of the 34th International Conference on Machine
  Learning-Volume 70}, pp.\  1372--1383. JMLR. org, 2017.

\bibitem[Hanna et~al.(2017)Hanna, Stone, and Niekum]{hanna2017bootstrapping}
Hanna, J.~P., Stone, P., and Niekum, S.
\newblock Bootstrapping with models: Confidence intervals for off-policy
  evaluation.
\newblock In \emph{Thirty-First AAAI Conference on Artificial Intelligence},
  2017.

\bibitem[Hao et~al.(2020{\natexlab{a}})Hao, Abbasi-Yadkori, Wen, and
  Cheng]{hao2019bootstrapping}
Hao, B., Abbasi-Yadkori, Y., Wen, Z., and Cheng, G.
\newblock Bootstrapping upper confidence bound.
\newblock \emph{Thirty-fourth Annual Conference on Neural Information
  Processing Systems}, 2020{\natexlab{a}}.

\bibitem[Hao et~al.(2020{\natexlab{b}})Hao, Duan, Lattimore, Szepesv{\'a}ri,
  and Wang]{hao2020sparse}
Hao, B., Duan, Y., Lattimore, T., Szepesv{\'a}ri, C., and Wang, M.
\newblock Sparse feature selection makes batch reinforcement learning more
  sample efficient.
\newblock \emph{arXiv preprint arXiv:2011.04019}, 2020{\natexlab{b}}.

\bibitem[Jiang \& Huang(2020)Jiang and Huang]{jiang2020minimax}
Jiang, N. and Huang, J.
\newblock Minimax value interval for off-policy evaluation and policy
  optimization.
\newblock \emph{Advances in Neural Information Processing Systems}, 33, 2020.

\bibitem[Jiang \& Li(2016)Jiang and Li]{jiang2016doubly}
Jiang, N. and Li, L.
\newblock Doubly robust off-policy value evaluation for reinforcement learning.
\newblock In \emph{International Conference on Machine Learning}, pp.\
  652--661, 2016.

\bibitem[Kallus \& Uehara(2020)Kallus and Uehara]{kallus2020double}
Kallus, N. and Uehara, M.
\newblock Double reinforcement learning for efficient off-policy evaluation in
  markov decision processes.
\newblock \emph{Journal of Machine Learning Research}, 21\penalty0
  (167):\penalty0 1--63, 2020.

\bibitem[Kato(2011)]{kato2011note}
Kato, K.
\newblock A note on moment convergence of bootstrap m-estimators.
\newblock \emph{Statistics \& Risk Modeling}, 28\penalty0 (1):\penalty0 51--61,
  2011.

\bibitem[Kleiner et~al.(2014)Kleiner, Talwalkar, Sarkar, and
  Jordan]{kleiner2014scalable}
Kleiner, A., Talwalkar, A., Sarkar, P., and Jordan, M.~I.
\newblock A scalable bootstrap for massive data.
\newblock \emph{Journal of the Royal Statistical Society: Series B: Statistical
  Methodology}, pp.\  795--816, 2014.

\bibitem[Kostrikov \& Nachum(2020)Kostrikov and
  Nachum]{kostrikov2020statistical}
Kostrikov, I. and Nachum, O.
\newblock Statistical bootstrapping for uncertainty estimation in off-policy
  evaluation.
\newblock \emph{arXiv preprint arXiv:2007.13609}, 2020.

\bibitem[Kuzborskij et~al.(2020)Kuzborskij, Vernade, Gy{\"o}rgy, and
  Szepesv{\'a}ri]{kuzborskij2020confident}
Kuzborskij, I., Vernade, C., Gy{\"o}rgy, A., and Szepesv{\'a}ri, C.
\newblock Confident off-policy evaluation and selection through self-normalized
  importance weighting.
\newblock \emph{arXiv preprint arXiv:2006.10460}, 2020.

\bibitem[Lagoudakis \& Parr(2003)Lagoudakis and Parr]{lagoudakis2003least}
Lagoudakis, M.~G. and Parr, R.
\newblock Least-squares policy iteration.
\newblock \emph{Journal of machine learning research}, 4\penalty0
  (Dec):\penalty0 1107--1149, 2003.

\bibitem[Le et~al.(2019)Le, Voloshin, and Yue]{le2019batch}
Le, H.~M., Voloshin, C., and Yue, Y.
\newblock Batch policy learning under constraints.
\newblock \emph{Proceedings of Machine Learning Research}, 97:\penalty0
  3703--3712, 2019.

\bibitem[Liao et~al.(2019)Liao, Klasnja, and Murphy]{liao2019off}
Liao, P., Klasnja, P., and Murphy, S.
\newblock Off-policy estimation of long-term average outcomes with applications
  to mobile health.
\newblock \emph{arXiv preprint arXiv:1912.13088}, 2019.

\bibitem[Liu et~al.(2018)Liu, Li, Tang, and Zhou]{liu2018breaking}
Liu, Q., Li, L., Tang, Z., and Zhou, D.
\newblock Breaking the curse of horizon: Infinite-horizon off-policy
  estimation.
\newblock In \emph{Advances in Neural Information Processing Systems}, pp.\
  5356--5366, 2018.

\bibitem[McLeish et~al.(1974)]{mcleish1974dependent}
McLeish, D.~L. et~al.
\newblock Dependent central limit theorems and invariance principles.
\newblock \emph{the Annals of Probability}, 2\penalty0 (4):\penalty0 620--628,
  1974.

\bibitem[Moore(1990)]{moore1990efficient}
Moore, A.~W.
\newblock Efficient memory-based learning for robot control.
\newblock 1990.

\bibitem[Munos \& Szepesv{\'a}ri(2008)Munos and
  Szepesv{\'a}ri]{munos2008finite}
Munos, R. and Szepesv{\'a}ri, C.
\newblock Finite-time bounds for fitted value iteration.
\newblock \emph{Journal of Machine Learning Research}, 9\penalty0 (5), 2008.

\bibitem[Nachum et~al.(2019)Nachum, Chow, Dai, and Li]{nachum2019dualdice}
Nachum, O., Chow, Y., Dai, B., and Li, L.
\newblock {DualDICE}: Behavior-agnostic estimation of discounted stationary
  distribution corrections.
\newblock In \emph{Advances in Neural Information Processing Systems}, pp.\
  2315--2325, 2019.

\bibitem[Oberst \& Sontag(2019)Oberst and Sontag]{oberst2019counterfactual}
Oberst, M. and Sontag, D.
\newblock Counterfactual off-policy evaluation with gumbel-max structural
  causal models.
\newblock In \emph{International Conference on Machine Learning}, pp.\
  4881--4890, 2019.

\bibitem[Paine et~al.(2020)Paine, Paduraru, Michi, Gulcehre, Zolna, Novikov,
  Wang, and de~Freitas]{paine2020hyperparameter}
Paine, T.~L., Paduraru, C., Michi, A., Gulcehre, C., Zolna, K., Novikov, A.,
  Wang, Z., and de~Freitas, N.
\newblock Hyperparameter selection for offline reinforcement learning.
\newblock \emph{arXiv preprint arXiv:2007.09055}, 2020.

\bibitem[Petersen \& Pedersen(2008)Petersen and Pedersen]{petersen2008matrix}
Petersen, K. and Pedersen, M.
\newblock The matrix cookbook. technical university of denmark.
\newblock \emph{Technical Manual}, 2008.

\bibitem[Precup et~al.(2000)Precup, Sutton, and Singh]{precup2000eligibility}
Precup, D., Sutton, R.~S., and Singh, S.
\newblock Eligibility traces for off-policy policy evaluation.
\newblock In \emph{ICML'00 Proceedings of the Seventeenth International
  Conference on Machine Learning}, 2000.

\bibitem[Sengupta et~al.(2016)Sengupta, Volgushev, and
  Shao]{sengupta2016subsampled}
Sengupta, S., Volgushev, S., and Shao, X.
\newblock A subsampled double bootstrap for massive data.
\newblock \emph{Journal of the American Statistical Association}, 111\penalty0
  (515):\penalty0 1222--1232, 2016.

\bibitem[Shi et~al.(2020)Shi, Zhang, Lu, and Song]{shi2020statistical}
Shi, C., Zhang, S., Lu, W., and Song, R.
\newblock Statistical inference of the value function for reinforcement
  learning in infinite horizon settings.
\newblock \emph{arXiv preprint arXiv:2001.04515}, 2020.

\bibitem[Singh(1981)]{singh1981asymptotic}
Singh, K.
\newblock On the asymptotic accuracy of efron's bootstrap.
\newblock \emph{The Annals of Statistics}, pp.\  1187--1195, 1981.

\bibitem[Sutton \& Barto(2018)Sutton and Barto]{sutton2018reinforcement}
Sutton, R.~S. and Barto, A.~G.
\newblock \emph{Reinforcement learning: An introduction}.
\newblock MIT press, 2018.

\bibitem[Thomas \& Brunskill(2016)Thomas and Brunskill]{thomas2016data}
Thomas, P. and Brunskill, E.
\newblock Data-efficient off-policy policy evaluation for reinforcement
  learning.
\newblock In \emph{International Conference on Machine Learning}, pp.\
  2139--2148, 2016.

\bibitem[Thomas et~al.(2015)Thomas, Theocharous, and
  Ghavamzadeh]{thomas2015high}
Thomas, P., Theocharous, G., and Ghavamzadeh, M.
\newblock High confidence policy improvement.
\newblock In \emph{International Conference on Machine Learning}, pp.\
  2380--2388. PMLR, 2015.

\bibitem[Uehara \& Jiang(2019)Uehara and Jiang]{uehara2019minimax}
Uehara, M. and Jiang, N.
\newblock Minimax weight and {Q}-function learning for off-policy evaluation.
\newblock \emph{arXiv preprint arXiv:1910.12809}, 2019.

\bibitem[Van~der Vaart(2000)]{van2000asymptotic}
Van~der Vaart, A.~W.
\newblock \emph{Asymptotic statistics}, volume~3.
\newblock Cambridge university press, 2000.

\bibitem[Voloshin et~al.(2019)Voloshin, Le, Jiang, and
  Yue]{voloshin2019empirical}
Voloshin, C., Le, H.~M., Jiang, N., and Yue, Y.
\newblock Empirical study of off-policy policy evaluation for reinforcement
  learning.
\newblock \emph{arXiv preprint arXiv:1911.06854}, 2019.

\bibitem[Wang et~al.(2020)Wang, Foster, and Kakade]{wang2020statistical}
Wang, R., Foster, D.~P., and Kakade, S.~M.
\newblock What are the statistical limits of offline rl with linear function
  approximation?
\newblock \emph{arXiv preprint arXiv:2010.11895}, 2020.

\bibitem[Xie et~al.(2019)Xie, Ma, and Wang]{xie2019towards}
Xie, T., Ma, Y., and Wang, Y.-X.
\newblock Towards optimal off-policy evaluation for reinforcement learning with
  marginalized importance sampling.
\newblock In \emph{Advances in Neural Information Processing Systems}, pp.\
  9665--9675, 2019.

\bibitem[Yin \& Wang(2020)Yin and Wang]{yin2020asymptotically}
Yin, M. and Wang, Y.-X.
\newblock Asymptotically efficient off-policy evaluation for tabular
  reinforcement learning.
\newblock \emph{International Conference on Artificial Intelligence and
  Statistics}, 2020.

\bibitem[Zhang et~al.(2020{\natexlab{a}})Zhang, Dai, Li, and
  Schuurmans]{zhang2020gendice}
Zhang, R., Dai, B., Li, L., and Schuurmans, D.
\newblock {GenDICE}: Generalized offline estimation of stationary values.
\newblock \emph{arXiv preprint arXiv:2002.09072}, 2020{\natexlab{a}}.

\bibitem[Zhang et~al.(2020{\natexlab{b}})Zhang, Liu, and
  Whiteson]{zhang2020gradientdice}
Zhang, S., Liu, B., and Whiteson, S.
\newblock {GradientDICE}: Rethinking generalized offline estimation of
  stationary values.
\newblock \emph{arXiv preprint arXiv:2001.11113}, 2020{\natexlab{b}}.

\end{thebibliography}
